\PassOptionsToPackage{table}{xcolor}
\PassOptionsToPackage{pagebackref,breaklinks=true,colorlinks,citecolor=blue,urlcolor=blue,linkcolor=blue,bookmarks=false}{hyperref}
\PassOptionsToPackage{noabbrev,nameinlink,capitalize}{cleveref}

\documentclass[onecolumn]{april_aigc}

\usepackage[utf8]{inputenc} 
\usepackage{url}            
\usepackage{amsfonts}       
\usepackage{amssymb}        
\usepackage{amsmath}        
\usepackage{nicefrac}       
\usepackage{microtype}      
\usepackage{xspace}         
\usepackage{fix-cm}
\usepackage[T1]{fontenc}

\usepackage{booktabs}       
\usepackage{wrapfig}        
\usepackage{multirow}       
\usepackage{makecell}       
\usepackage{tabularx}       
\usepackage{adjustbox}      
\usepackage{colortbl}       
\usepackage{pifont}         
\usepackage{enumitem}
\usepackage[normalem]{ulem}

\usepackage{xcolor}
\definecolor{linkcolor}{named}{aprilblue}
\definecolor{urlcolor}{RGB}{255,105,180}
\definecolor{citecolor}{RGB}{66,168,235}
\definecolor{lightgray}{rgb}{0.8, 0.8, 0.8}
\definecolor{darkgreen}{rgb}{0.00, 0.81, 0.78}

\definecolor{gray_tab}{RGB}{220, 220, 220}
\definecolor{blue_tab}{RGB}{227, 240, 251}
\definecolor{oran_tab}{RGB}{252, 242, 237}
\definecolor{whit_tab}{RGB}{255, 255, 255}
\definecolor{green_code}{RGB}{55, 126, 34}

\usepackage{algorithm}
\usepackage{algorithmic}
\usepackage{listings}
\usepackage{etoolbox}

\makeatletter
\AfterEndEnvironment{algorithm}{\let\@algcomment\relax}
\AtEndEnvironment{algorithm}{\kern2pt\hrule\relax\vskip3pt\@algcomment}
\let\@algcomment\relax
\newcommand\algcomment[1]{\def\@algcomment{\footnotesize#1}}
\renewcommand\fs@ruled{\def\@fs@cfont{\bfseries}\let\@fs@capt\floatc@ruled
  \def\@fs@pre{\hrule height.8pt depth0pt \kern2pt}%
  \def\@fs@post{}%
  \def\@fs@mid{\kern2pt\hrule\kern2pt}%
  \let\@fs@iftopcapt\iftrue}
\makeatother

\usepackage[pagebackref,breaklinks=true,colorlinks,citecolor=blue,urlcolor=blue,linkcolor=blue,bookmarks=false]{hyperref}
\AtEndPreamble{
    \usepackage[capitalize]{cleveref}
    \crefname{section}{Sec.}{Secs.}
    \Crefname{section}{Section}{Sections}
    \crefname{table}{Tab.}{Tabs.}
    \Crefname{table}{Table}{Tables}
    \crefname{equation}{Eq.}{Eqs.}
    \Crefname{equation}{Equation}{Equations}
    \crefname{figure}{Fig.}{Figs.}
    \Crefname{figure}{Figure}{Figures}
}
\hypersetup{colorlinks=true,linkcolor=linkcolor,urlcolor=urlcolor,citecolor=citecolor}

\usepackage{caption}
\DeclareCaptionFormat{custom}{{\color{aprilblue}\sffamily\textbf{#1 #2}} #3}
\titleformat*{\section}{\color{aprilblue}\Large\sffamily\bfseries}
\titleformat*{\subsection}{\color{aprilblue}\large\sffamily\bfseries}
\titleformat*{\subsubsection}{\color{aprilblue}\normalsize\sffamily\bfseries}

\usepackage{fancyhdr}
\newif\ifshowlogo
\showlogotrue   
\newcommand{\insertlogo}{%
  \ifshowlogo
    \IfFileExists{assets/april_logo1.png}%
    {\includegraphics[height=0.68cm]{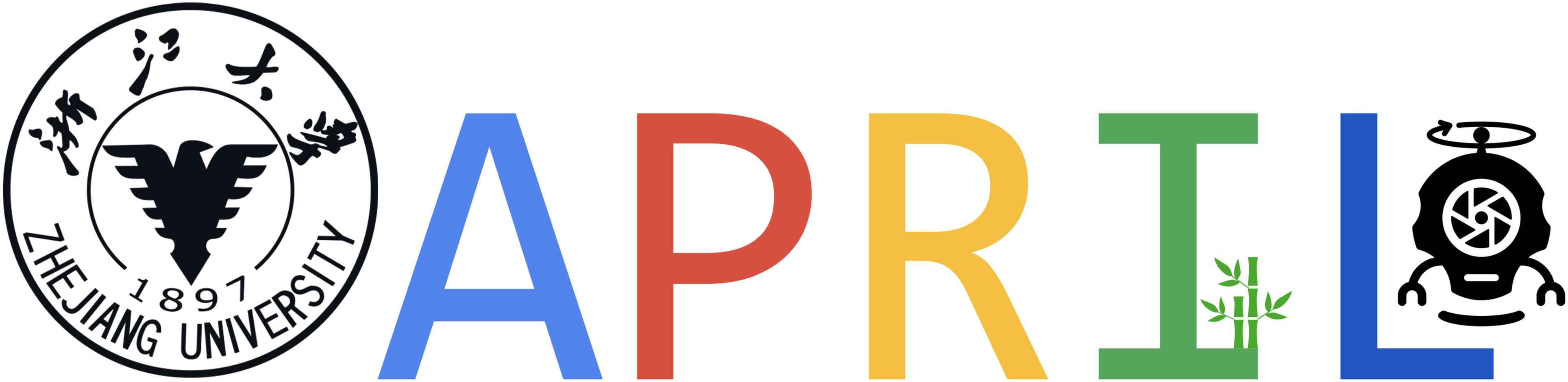}}
    {}%
  \fi
}
\newif\ifshowtoc
\showtoctrue   

\renewcommand{\title}[1]{\def\titlelist{{\fontsize{20pt}{28pt}\selectfont\sffamily\bfseries #1}}}
\title{Evolution of Optimization Methods: Algorithms, Scenarios, and Evaluations}

\author[1]{Tong Zhang}
\author[1\dagger\raisebox{-0.2em}{\includegraphics[height=0.85em]{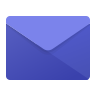}}]{Jiangning Zhang}
\author[1]{Zhucun Xue}
\author[1]{Juntao Jiang}
\author[1]{Yicheng Xu}
\author[2]{Chengming Xu}
\author[3]{Teng Hu}
\author[4]{Xingyu Xie}
\author[5\dagger]{Xiaobin Hu}
\author[1]{Yabiao Wang}
\author[1\dagger]{Yong Liu}
\author[4]{Shuicheng Yan}

\affiliation[1]{Zhejiang University, APRIL Lab}
\affiliation[2]{Fudan University}
\affiliation[3]{Shanghai Jiaotong University}
\affiliation[4]{National University of Singapore}

\abstract{
Balancing convergence speed, generalization capability, and computational efficiency remains a core challenge in deep learning optimization. First-order gradient descent methods, epitomized by stochastic gradient descent (SGD) and Adam, serve as the cornerstone of modern training pipelines. However, large-scale model training, stringent differential privacy requirements, and distributed learning paradigms expose critical limitations in these conventional approaches regarding privacy protection and memory efficiency. To mitigate these bottlenecks, researchers explore second-order optimization techniques to surpass first-order performance ceilings, while zeroth-order methods reemerge to alleviate memory constraints inherent to large-scale training. Despite this proliferation of methodologies, the field lacks a cohesive framework that unifies underlying principles and delineates application scenarios for these disparate approaches. In this work, we retrospectively analyze the evolutionary trajectory of deep learning optimization algorithms and present a comprehensive empirical evaluation of mainstream optimizers across diverse model architectures and training scenarios. We distill key emerging trends and fundamental design trade-offs, pinpointing promising directions for future research. By synthesizing theoretical insights with extensive empirical evidence, we provide actionable guidance for designing next-generation highly efficient, robust, and trustworthy optimization methods.
}

\coverdate{\today}
\coverproject{https://github.com/APRIL-AIGC/Awesome-Optimizer}

\begin{document}
\pagenumbering{gobble} 
\maketitle
\thispagestyle{empty} 

\ifshowtoc
    \setcounter{tocdepth}{3} 
    
    \makeatletter
    \makeatother
    \tableofcontents
    \vspace{1.0cm} 


    \clearpage
\fi
\twocolumn 
\pagenumbering{arabic}

\begin{figure*}[t!] 
    \centering
    \adjincludegraphics[width=1.0\linewidth]{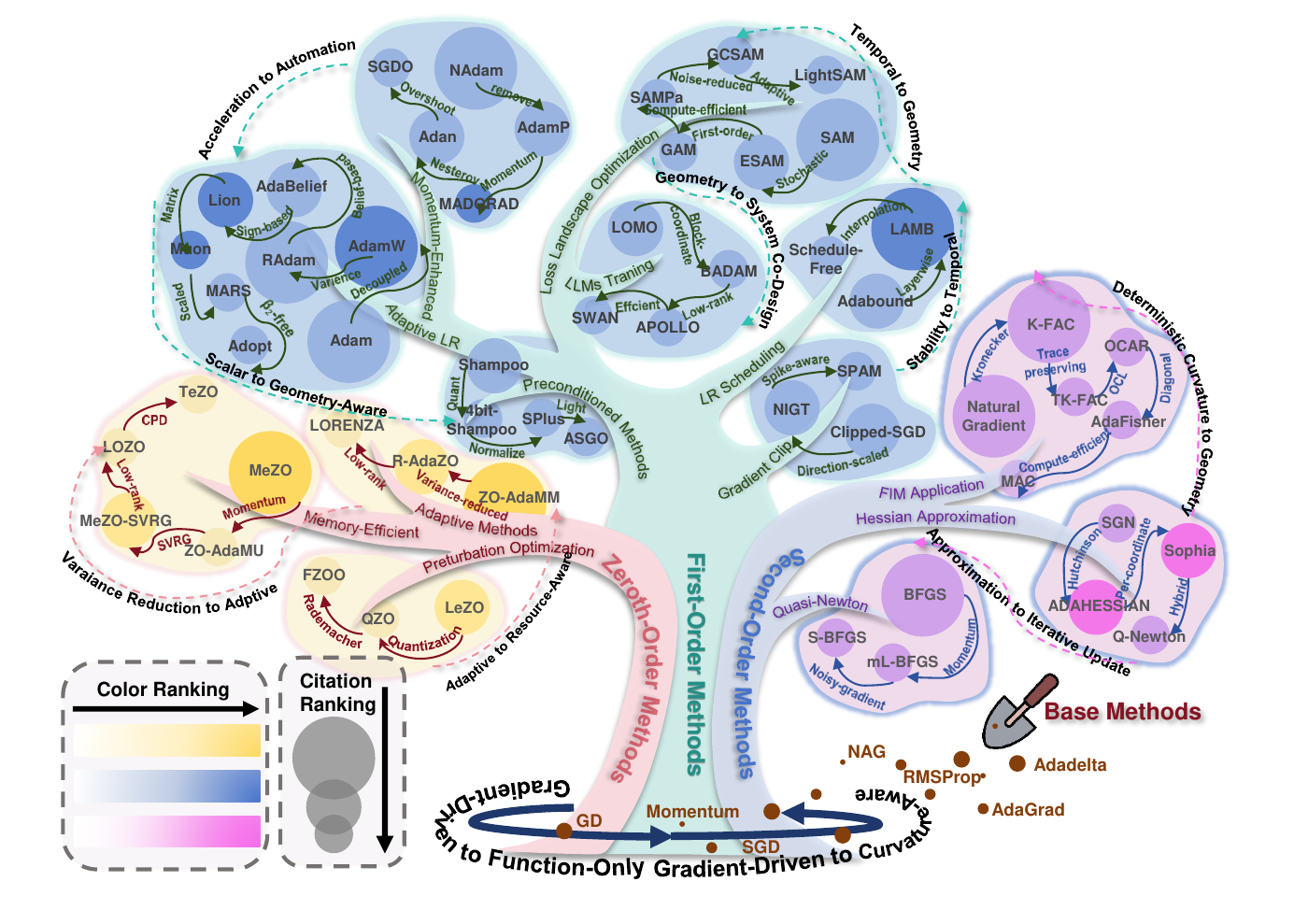}
   
    \caption{\textbf{The evolutionary tree of optimization methods.} Rooted in classic base methods, the development trajectory branches into first-order, second-order, and zeroth-order paradigms. Node sizes reflect citation impact, while distinct clusters illustrate the progression from foundational gradient updates towards advanced, scenario-tailored frameworks.}
    
    \label{fig:optim_main}
    \vspace{-1.0em} 
\end{figure*}
\section{Introduction} \label{section:intro}
Deep learning has historically been driven by foundational optimization algorithms, with early methods like stochastic gradient descent (SGD)~\cite{307} establishing the basic training framework, and the Adam~\cite{287} family later becoming the standard for various tasks. However, the unprecedented scale of modern foundation models has fundamentally shifted this optimization paradigm. During the training of large language models (LLMs), these traditional workhorses have exposed severe limitations. The primary bottleneck is no longer merely the theoretical convergence rate, but the physical and systemic feasibility of training: the memory wall imposed by backpropagation, the communication wall in decentralized networks, and the privacy wall for sensitive data. This survey argues that the recent diversification of optimization algorithms represents a unified, albeit fragmented, response to this feasibility crisis. We demonstrate that traditional mathematical primitives, including first-order (FO) workhorses, second-order (SO) curvature-aware methods, and the recently revived memory-efficient zeroth-order (ZO) techniques, are being fundamentally re-architected. By evolving into scenario-oriented paradigms, including distributed systems and rigorously differentially private frameworks, modern optimizers are moving beyond pure algorithmic design to become highly constrained, systems-aware engineering solutions. Optimization methods can generally be divided into four mainstream research fields: \textbf{\textit{1)}} FO~\cite{307,287} methods rely on first-order gradients (and their derived statistics) to achieve superior convergence with low computational overhead, strictly avoiding the explicit approximation of true second-order curvature. \textbf{\textit{2)}} SO methods~\cite{280,282,283} explicitly construct and incorporate true curvature information (e.g., the Hessian or true FIM) to precondition the update direction, aiming to overcome the fundamental performance limits of FO algorithms. \textbf{\textit{3)}} ZO methods~\cite{284} aim to approximate gradient directions via function evaluations to avoid the massive memory burden of backpropagation. And one overarching application framework: \textbf{\textit{4)}} Scenario-oriented paradigms, which focus on re-engineering these base primitives for distributed communication, differential privacy, and other specialized use cases. The development of foundational techniques has laid the groundwork for these specialized paradigms, but the field now requires a more unified perspective to advance. 

Also, we introduce a standardized evaluation framework to evaluate 23 optimizers across continuous vision tasks (ResNet~\cite{297}, ViT~\cite{296}) and discrete causal language modeling (Llama~\cite{330}). Our extensive benchmarking yields several critical insights: First, foundational FO algorithms like SGD~\cite{307} suffer catastrophic training collapse on the highly anisotropic loss landscapes of LLMs (\cref{fig:figure1_2}), dictating the absolute necessity of adaptive scaling. Second, while standard adaptive methods (e.g., Adam~\cite{287}) exhibit severe sensitivity to large learning rates, emerging methods utilizing matrix orthogonalization (e.g., Muon~\cite{173}) or gradient correction (e.g., MARS~\cite{181}) demonstrate superior hyperparameter robustness and cross-architecture generalization (\cref{generalization}). 

Existing surveys~\cite{310,311} suffer from three critical limitations that hinder the development of the field. First, most reviews are limited in scope, concentrating nearly entirely on conventional FO and SO approaches while neglecting ZO optimization and scenario-oriented frameworks. These two fast-developing paradigms are now essential for memory-efficient LLM training, distributed learning, and privacy-preserving applications. Second, existing works lack a unified, mathematically rigorous taxonomy, resulting in fragmented, inconsistent categorization of algorithms that obscures the intrinsic connections and evolutionary logic between different methods. Third, none of the existing surveys provide a standardized, large-scale empirical benchmark for fair cross-architecture evaluation of modern optimizers, leading to conflicting, non-reproducible conclusions in the literature and leaving researchers and engineers without reliable guidance for optimizer selection in practical scenarios. To address these critical gaps, this survey provides a systematic, panoramic review of deep learning optimization methods, with a unified theoretical framework, standardized empirical benchmark, and actionable insights for both theoretical research and engineering practice.

\textbf{Contributions.} In this survey, we make three key contributions to the field of deep learning optimization: 
\textbf{\textit{1) Unified Taxonomy:}} We unify disparate conceptual definitions and establish a rigorous mathematical taxonomy (\cref{section:background}), tracing the evolution of fundamental optimization primitives from FO to SO and ZO methods. This taxonomy clarifies the intrinsic connections and evolutionary logic between different optimization approaches, providing a coherent framework for understanding the field's development. 
\textbf{\textit{2) Scenario-Oriented Analysis:}} We demonstrate how these foundational algorithms are being fundamentally re-architected into scenario-oriented paradigms to address severe physical bottlenecks, such as distributed communication barriers and strict differential privacy constraints (\cref{section:method}). This analysis highlights the shift from pure algorithmic design to systems-aware engineering solutions that balance theoretical guarantees with practical constraints. 
\textbf{\textit{3) Standardized Evaluation:}} We introduce a rigorously controlled evaluation framework that separates algorithmic performance from large-scale engineering optimizations. We develop a standardized testbed to evaluate 23 distinct optimizers across diverse architectural proxies, including CNN~\cite{337} and Transformer-based~\cite{334} models. This extensive evaluation systematically isolates and examines learning rate sensitivity, long term training scalability, and cross-architecture generalization, providing quantitative and qualitative insights that guide the design of next generation's optimizers.

\textbf{Survey scope.} 
This survey primarily focuses on four core optimization paradigms: FO, SO, and ZO optimization, along with scenario-oriented optimization frameworks. Given the vast volume of literature, including both published articles and preprints, we primarily include representative and impactful works.

\textbf{Survey pipeline.}~\cref{fig:optim_main} illustrates the evolutionary trajectory of optimization methods, systematically classified by their respective gradient orders. We first cover the background knowledge, including the concept of fundamental optimization methods (\cref{section:background}). Next, we conduct a comprehensive review of various optimization methods categorized by their gradient information usage and application scenarios (\cref{section:method}). We then provide comprehensive experimental evaluations of representative optimizers (\cref{section:exp}). In \cref{section:fut}, we critically examine remaining challenges and highlights potential directions for future research. Finally, \cref{section:con} provides a concise summary of the survey. We closely follow the latest developments in this \href{https://github.com/APRIL-AIGC/awesome-optimizer}{project}.

\section{Background} \label{section:background}
This section establishes the core conceptual foundations, critical challenges, and standardized evaluation metrics underpinning modern deep learning optimization. We first formalize the canonical optimization problem setup for empirical risk minimization, then analyze the unique non-convex loss landscape of deep neural networks, and finally survey key evaluation metrics and hardware-aware scaling strategies tailored to large-scale model training.
\subsection{Optimization Problem Setup}\label{subsection:optimization_setup}
\textbf{Objective function.} The fundamental goal of optimization is to find a set of parameters $\theta \in \mathbb{R}^d$ such that the model $f(\cdot; \theta)$ performs optimally on unseen data. Assuming data samples $z = (x, y)$ are drawn from an unknown true distribution $\mathcal{P}$, our objective is to minimize the expected risk, denoted as $R(\theta)$:
\begin{equation}
    R(\theta) = \mathbb{E}_{z \sim \mathcal{P}} [\ell(f(x; \theta), y)],
\end{equation}
where $\ell(\cdot, \cdot)$ represents the loss function. Since the distribution $\mathcal{P}$ is unknown, $R(\theta)$ cannot be computed directly. Following the principle of Empirical Risk Minimization~\cite{312}, we use a training set $\mathcal{S} = \{z_1, \dots, z_N\}$ consisting of $N$ independent and identically distributed (i.i.d.) samples to construct a surrogate objective function, the empirical risk $ J(\theta)$:
\begin{equation}
    \min_{\theta \in \mathbb{R}^d} \ J(\theta) = \frac{1}{N} \sum_{i=1}^N \ell(f(x_i; \theta), y_i).
\end{equation}
While $J(\theta)$ converges to $R(\theta)$ as $N \to \infty$, in finite-sample regimes, excessive optimization of $J(\theta)$ may lead to overfitting. Therefore, the design of an optimizer must consider not only the minimization of training loss but also the generalization gap, often addressing this through regularization or implicit bias.\\
\textbf{Stochastic gradients.} Computing the full gradient $\nabla J(\theta) = \frac{1}{N} \sum_{i=1}^N \nabla \ell(z_i; \theta)$ is computationally prohibitive for large datasets. Consequently, modern optimizers universally adopt stochastic gradients. At iteration $t$, a mini-batch $\mathcal{B}_t$ is sampled to compute the gradient estimate $g_t(\theta)$:
\begin{equation}
    g_t(\theta) = \frac{1}{|\mathcal{B}_t|} \sum_{z \in \mathcal{B}_t} \nabla \ell(z; \theta).
\end{equation}
For theoretical analysis, it is standard to assume that $g_t$ is an unbiased estimator of the true gradient, i.e., $\mathbb{E}[g_t] = \nabla J(\theta)$, with bounded variance $\mathbb{E}[\|g_t - \nabla J\|^2] \le \sigma^2$.\\
\textbf{Convexity.} To analyze the convergence properties of optimization algorithms, we formally define convexity and strong convexity. Let $f: \mathbb{R}^d \to \mathbb{R}$ be a differentiable function representing the objective loss.

A differentiable function $f$ is convex if, for all $x, y \in \mathbb{R}^d$, its first-order Taylor approximation provides a global underestimator:
\begin{equation}
    f(y) \geq f(x) + \langle \nabla f(x), y - x \rangle.
\end{equation}
This property implies that every local minimum is necessarily a global minimum. If $f$ is twice differentiable, convexity is equivalent to the condition that the Hessian matrix is positive semi-definite, denoted as $\nabla^2 f(x) \succeq 0$, for all $x \in \mathbb{R}^d$.

A function $f$ is $\mu$-strongly convex with parameter $\mu > 0$ if, for all $x, y \in \mathbb{R}^d$, the following inequality holds:
\begin{equation}
    f(y) \geq f(x) + \langle \nabla f(x), y - x \rangle + \frac{\mu}{2} \|y - x\|^2.
\end{equation}
This property guarantees the existence of a unique global optimal solution. Intuitively, strong convexity implies that the function grows at least quadratically. For a twice-differentiable function, this is equivalent to requiring the eigenvalues of the Hessian to be bounded below by $\mu$:
\begin{equation}
    \nabla^2 f(x) \succeq \mu I,
\end{equation}
where $I$ denotes the identity matrix.

\textbf{Convergence.} Under classical convex optimization theory~\cite{313,314}, gradient descent (GD)~\cite{315} typically exhibits sublinear convergence with a rate of $O(1/T)$ for convex functions. Due to gradient noise variance, SGD~\cite{307} has a slower sublinear rate of $O(1/\sqrt{T})$. For strongly convex functions, GD~\cite{315} achieves linear convergence with a rate of $O((1 - \mu\eta)^T)$, where $\eta$ is the learning rate. SGD~\cite{307} exhibits linear convergence to a fixed error radius, which depends on the gradient noise variance and batch size; however, to attain exact convergence, the rate must reduce to sublinear.
\subsection{Deep Learning Optimization}
While convex theory provides intuition, the loss landscape of deep neural networks (DNNs)~\cite{336} is highly non-convex, presenting unique challenges that violate standard assumptions.\\
\textbf{Non-convexity.} The loss function of deep neural networks contains numerous local minima. Deep learning optimization leverages inherent gradient noise and high-dimensional geometry to escape saddle points, typically converging to flat minima that yield strong generalization performance.\\
\textbf{Saddle point.} In high-dimensional spaces, local minima are relatively scarce, and saddle points, which are local minima in some directions and a local maximum in others, are far more prevalent. For first-order methods like GD~\cite{315}, saddle points cause stagnation and slow convergence; SGD~\cite{307} can often escape saddle points due to stochastic gradient noise, making it more effective for nonconvex optimization.\\
\textbf{Over-parameterization regimes.} Modern deep networks are typically over-parameterized, meaning the number of parameters far exceeds the number of training samples. Under this regime, the optimization process exhibits distinctive properties. The neural tangent kernel (NTK)~\cite{316} theory shows that, in the infinite-width limit with small initialization and gradient flow, the training dynamics of neural networks can be approximated by kernel regression. This provides a theoretical framework for understanding why over-parameterized networks converge to global optima and generalize well.
\begin{figure}[t!] 
    \centering
    \adjincludegraphics[width=1.0\linewidth]{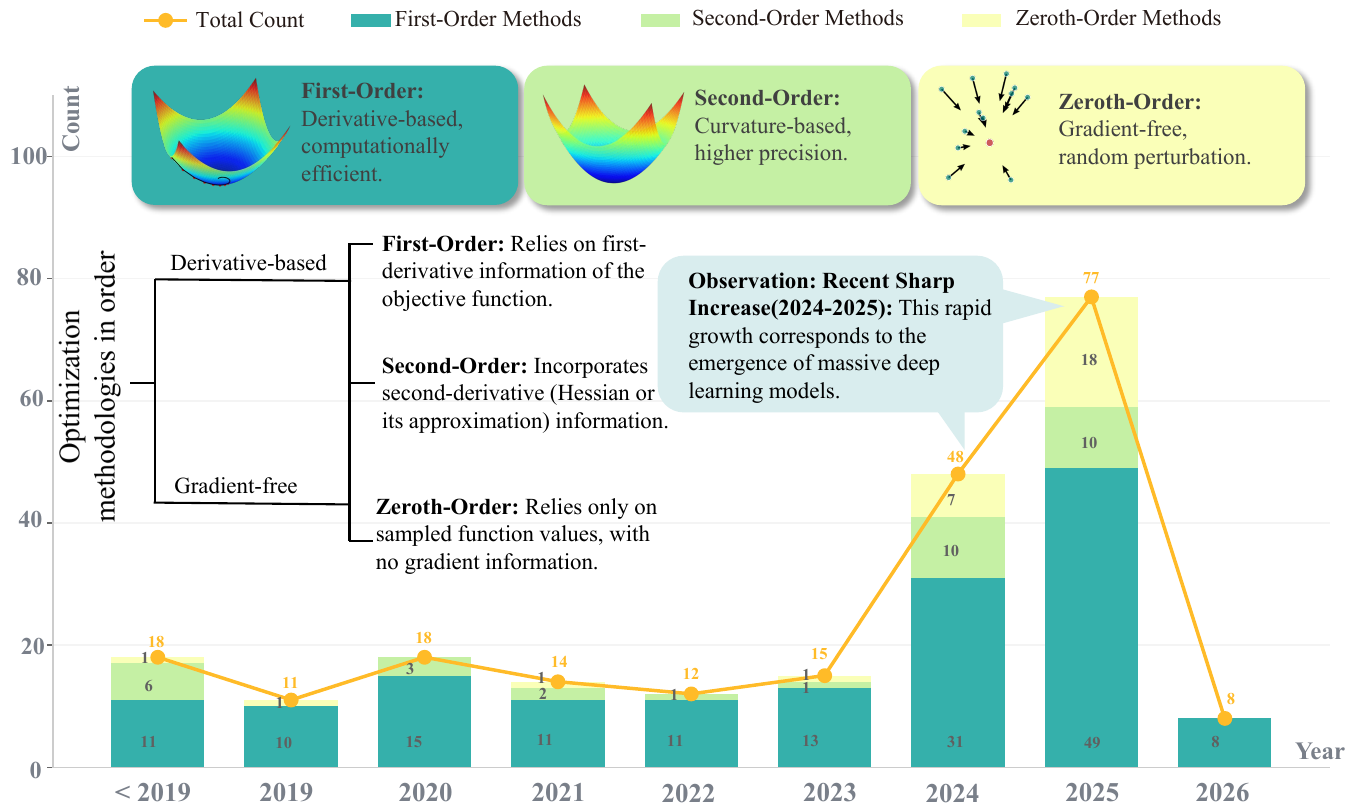}
   
    \caption{\textbf{Quantitative evolution of optimization methods over time.} The stacked bars categorize algorithms into FO, SO, and ZO algorithms, with the line graph tracking the total annual count. A significant surge is observed starting in 2024, driven by the rapid development and optimization demands of massive deep learning models. Please note that the statistics for 2026 are incomplete, covering data only up to April. Based on the prevailing upward trend, the total count for 2026 is expected to continue growing; thus, the current partial figure should not be misinterpreted as a decline in research activity.}
    
    \label{fig:optim_count}
    \vspace{-1.0em} 
\end{figure}
\subsection{Evaluation Metrics}
Evaluation metrics are as critical as model accuracy in large-scale training. Computational complexity is typically measured in floating-point operations (FLOPs), a metric that directly correlates with training duration and hardware costs. Training memory consumption primarily stems from two sources: activations and optimizer states. Activations are intermediate results generated during forward propagation that must be retained for gradient calculation during backward propagation, though their memory footprint can be reduced via gradient checkpointing~\cite{317}. Optimizer states introduce substantial overhead because methods like Adam~\cite{287} require storing first-order and second-order moment estimates. Furthermore, privacy and security considerations are necessary when processing sensitive data. Techniques such as differentially private SGD (DP-SGD)~\cite{318} provide privacy guarantees through gradient clipping and the addition of Gaussian noise, but this often reduces convergence speed and final model accuracy.\\
\subsection{Optimization for Large Models}
As model parameter counts scale toward the billion or even trillion level, optimization strategies must account for hardware constraints.\\
\textbf{Mixed precision training.} Using FP16 or BF16 instead of FP32 reduces memory footprint and enhances computational efficiency. Notably, FP16 requires loss scaling~\cite{319} to prevent numerical underflow of gradients, whereas BF16 typically does not need additional loss scaling because its dynamic range is consistent with that of FP32.\\
\textbf{Parallelism strategies.} It is necessary to combine data parallelism, tensor/model parallelism~\cite{320}, and pipeline parallelism~\cite{321} to overcome single-GPU memory limitations.\\
\textbf{Batch size and learning rate.} Increasing batch size enhances parallel efficiency; however, to maintain convergence, the learning rate must typically be scaled according to specific rules (e.g., the linear scaling rule $\eta \propto B$~\cite{322}).\\
\textbf{Gradient accumulation.} When memory constraints prevent the use of large batch sizes, gradients are accumulated over multiple micro-steps before performing a single parameter update, thereby simulating the effect of large-batch training.
\begin{figure}[t!] 
    \centering
    \adjincludegraphics[width=1.0\linewidth]{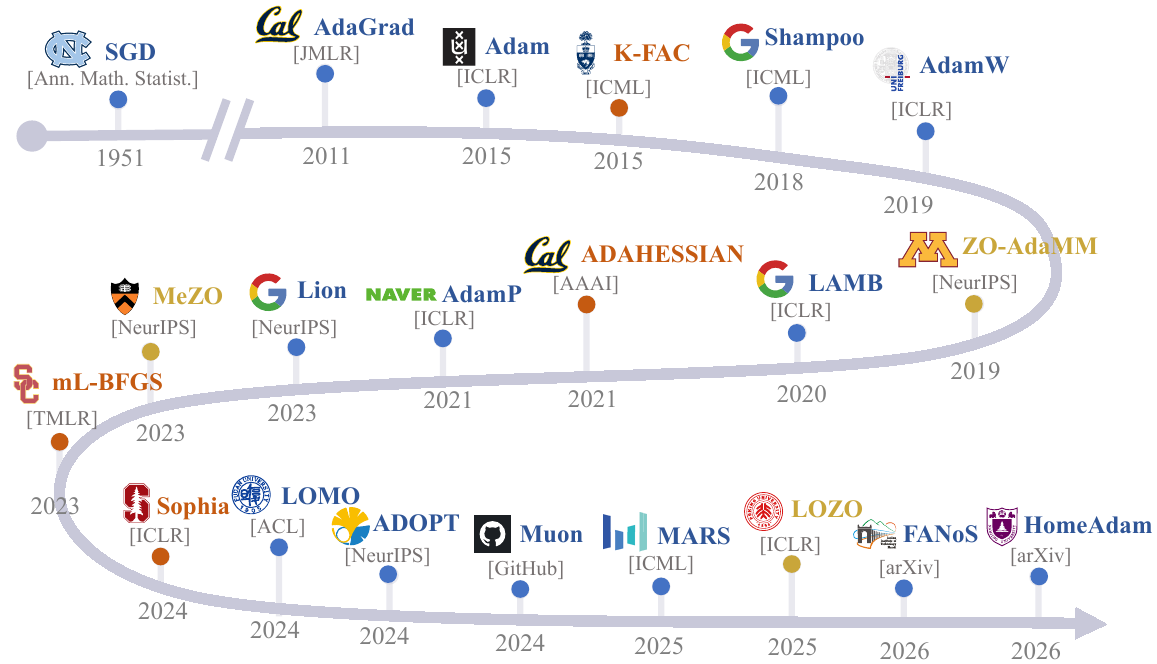}
    \caption{\textbf{Timeline of prominent optimization algorithms.} The evolution highlights key algorithmic milestones, associated research institutions, and publication venues over time.}
    \label{fig:optim_timeline}
    \vspace{-1.0em} 
\end{figure}
\subsection{History and Roadmap} 
Before detailing modern optimizers, we summarize their growth trends (\cref{fig:optim_count}) and evolution (\cref{fig:optim_timeline}). SGD~\cite{307} established the foundation of first-order training, while second-order methods like K-FAC~\cite{255} sought curvature-aware efficiency despite high computational costs. The rise of large language models has shifted the paradigm toward resource efficiency. To bypass backpropagation's memory bottlenecks, zeroth-order methods and memory-efficient designs have emerged. Recent developments, including automated rules like Lion~\cite{167}, matrix-wise approaches like Muon~\cite{173}, and low-rank adaptations like LOZO~\cite{24}, demonstrate a clear trend toward hardware-aware optimization. Detailed taxonomies of these methods are provided in \cref{section:method}.

\begin{figure*}[t!] 
    \centering
    \adjincludegraphics[width=1.0\linewidth]{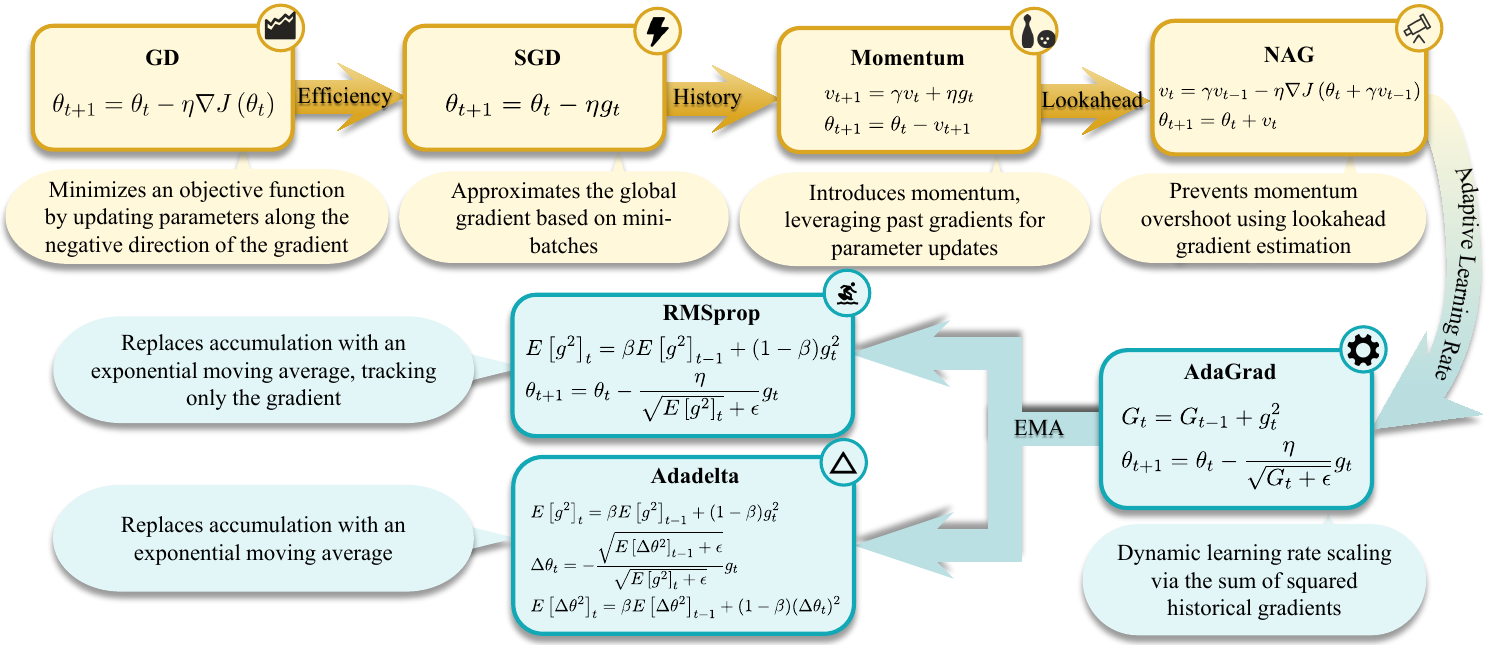}
   
    \caption{\textbf{Evolution and formulations of base methods.} We details the developmental trajectory from GD~\cite{315}, SGD~\cite{307}, Momentum~\cite{323}, NAG~\cite{324} to AdaGrad~\cite{169}, RMSprop~\cite{152}, Adadelta~\cite{325}, and highlights the core mathematical update rules and key transition mechanisms at each stage.}
    
    \label{fig:evolution_base}
    \vspace{-1.0em} 
\end{figure*}
\section{Optimization Algorithms}\label{section:method}
\begin{figure*}[t!] 
    \centering
    \adjincludegraphics[width=1.0\linewidth]{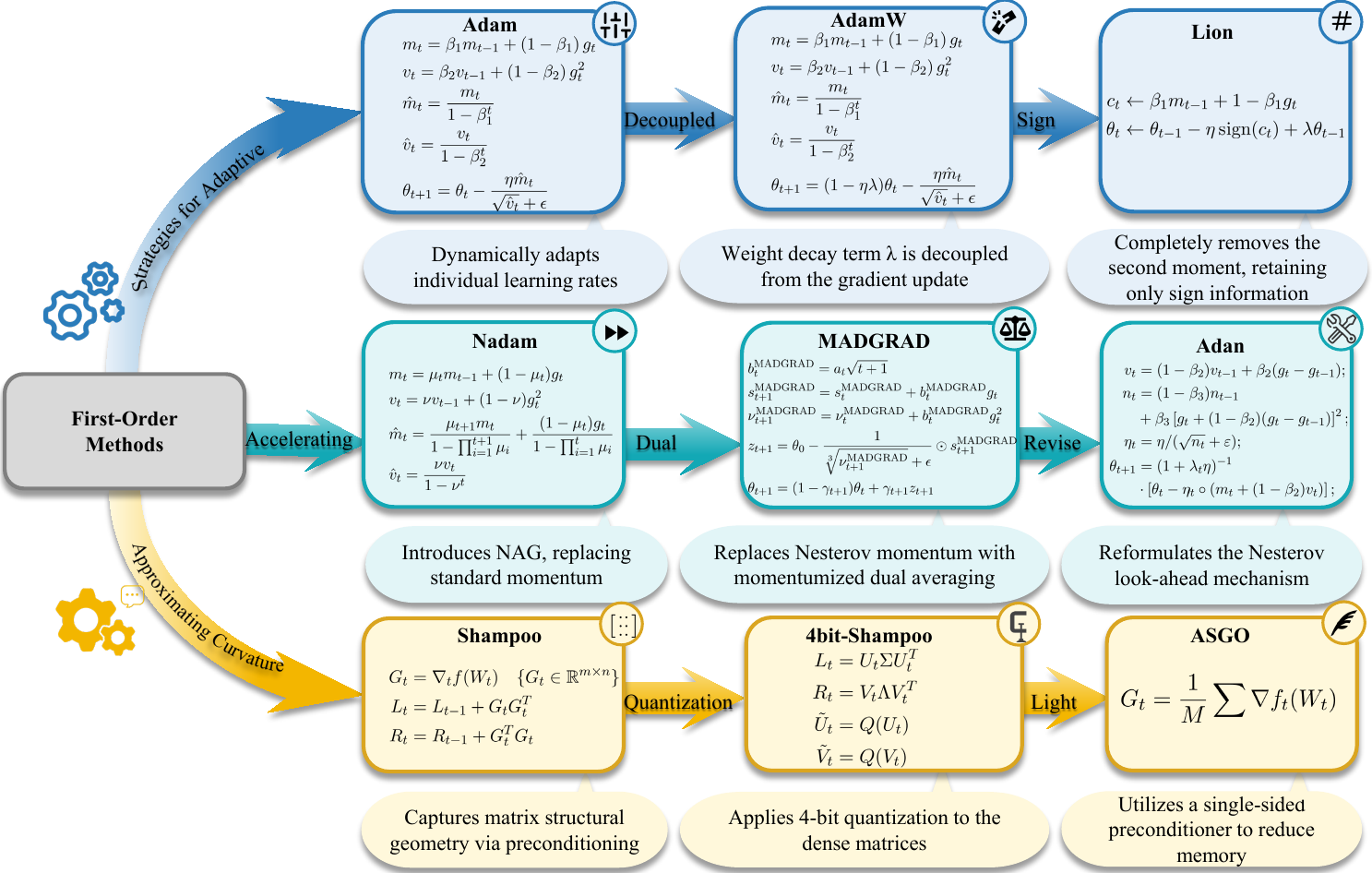}
    \caption{\textbf{Evolution and formulations of typical first-order methods.} We show the prominent algorithms from ~\cref{fo_methods}, categorizing these algorithms by adaptive strategies (~\cite{287,148,167}) acceleration (~\cite{155,171,168}), and curvature approximation (~\cite{170,172,176}), and analyze their key transitional mechanisms through the lens of mathematical formulations.}
    \label{fig:evolution_fo}
    \vspace{-1.0em} 
\end{figure*}

In this section, we conduct a comprehensive review of various optimization algorithms categorized by their gradient information usage and application scenarios. We systematically delineate the fundamental mathematical principles of neural network optimization across different gradient orders, FO, SO, and ZO, along with their evolutionary development and prevailing challenges (\cref{fo_methods,so_methods,zo_methods}). However, deploying these theoretical paradigms in the engineering practice of modern deep learning scenarios encounters severe bottlenecks. We therefore shift our research perspective from pure algorithmic design to scenario-driven optimization paradigms, systematically investigating how the aforementioned FO, SO, and ZO optimization primitives can be re-engineered to breach the memory wall, mitigate communication bottlenecks, and enforce rigorous privacy guarantees (\cref{distr_methods,privacy_methods,me_methods,to_methods}). As illustrated in~\cref{fig:optim_main}, to provide a comprehensive overview of the development of optimization algorithms, we visualize the existing landscape as an "evolutionary tree". Rooted in classic base methods such as GD~\cite{315}, SGD~\cite{307}, and momentum~\cite{323}, this tree branches into three distinct categories based on technical trajectories: first-order, second-order, and zeroth-order algorithms. Furthermore, the latter part of this section provides a supplementary classification based on application scenarios, such as distributed optimization and differential privacy. To facilitate a structured analysis of algorithmic evolution (\cref{fig:evolution_base}, \cref{fig:evolution_fo}, \cref{fig:evolution_so}, \cref{fig:evolution_zo}), we have unified the mathematical notation system (\cref{tab:notation}). This framework combines universal vanilla symbols with superscripted algorithm-specific notations to highlight the core differences between various methods. Furthermore, it explicitly delineates the specific improvements made by each algorithm relative to its predecessors.\\
Specifically,~\cref{fig:evolution_base} dissects the evolutionary logic of base methods from a formulation perspective: SGD~\cite{307} improves upon GD~\cite{315} by computing gradients based on mini-batches, significantly enhancing computational efficiency; momentum~\cite{323} introduces a variable $v_t$ to accumulate historical gradient information, ensuring parameter updates rely on more than just the current gradient; NAG~\cite{324} further refines this by computing the gradient at a "lookahead" position $(\theta_t + \gamma v_{t-1})$, correcting potential overshooting caused by momentum~\cite{323}; AdaGrad~\cite{169} introduces a cumulative term $G_t = G_{t-1} + g_t^2$ to achieve parameter-wise adaptive learning rates; RMSprop~\cite{325} replaces AdaGrad's direct accumulation with an exponential moving average $E[g^2]_t$, effectively preventing premature learning rate decay caused by an infinitely growing denominator; Adadelta~\cite{152} eliminates the reliance on a manually set global learning rate $\eta$, instead utilizing the root mean square of parameter updates $\sqrt{E[\Delta \theta^2]_{t-1}}$ for adaptive adjustment. We trace the evolutionary paths of first-order (\cref{fig:evolution_fo}), second-order (\cref{fig:evolution_so}), and zeroth-order (\cref{fig:evolution_zo}) algorithms through representative methods, providing detailed theoretical discussions for each category in~\cref{fo_methods},~\cref{so_methods}, and~\cref{zo_methods}.\\
\textbf{Motivation of survey organization.} Current research on emerging optimizers lacks a systematic classification framework and comparison of core logic, for several reasons:\\
\textbf{\textit{1) Coarse classification granularity:}} Existing surveys often lump diverse algorithms under generic labels such as "adaptive methods" without dissecting underlying mechanisms, such as specific momentum improvement strategies (e.g., scheduled reset vs. lookahead) or different types of gradient clipping.\\
\textbf{\textit{2) Disconnect from application scenarios:}} Current literature lacks practical guidance on which optimization logic applies best to specific engineering hurdles, such as high memory consumption in large models or noise sensitivity in differential privacy.\\
\textbf{\textit{3) Expanded optimization scope:}} The scope of optimization has expanded beyond narrow gradient updates to include broad algorithmic strategies like zeroth-order estimation for memory efficiency and communication-efficient compression for distributed systems. This makes it difficult for researchers to quickly identify suitable application scenarios and for engineers to select optimal algorithms.\\
By establishing a unified classification taxonomy and systematically analyzing algorithmic behaviors across diverse scenarios, this survey aims to inspire the structural design and theoretical advancement of next-generation optimization algorithms.

\subsection{Unified Mathematical Perspective}\label{unified_perspective}
To provide a unified basis for comparing modern optimizers, we depart from conventional categorizations and formulate a generalized constrained optimization framework. Most methods discussed in this survey can be cast as specific instantiations of the following discrete-time dynamical system:
\begin{equation}\label{eq:unified_master}
\begin{aligned}
    \tilde{g}_t &= \mathcal{E}(f, \theta_t, \xi_t), \\
    \hat{g}_t &= \mathcal{T}_{\text{scenario}} (\tilde{g}_t), \\
    m_t &= \phi(m_{t-1}, \hat{g}_t), \\
    \theta_{t+1} &= \mathcal{P}_{\Theta} \Big( \theta_t - \eta_t \, M_t^{-1} \, m_t - \eta_t \lambda \theta_t \Big),
\end{aligned}
\end{equation}
where $\xi_t$ denotes the stochastic data batch, $\eta_t$ is the step size at iteration $t$, and $\lambda$ represents weight decay. This unified formulation disentangles the optimization process into four distinct dimensions, allowing us to systematically categorize the evolutionary trajectories of existing methods:

\textbf{Gradient estimator $\mathcal{E}(\cdot)$.} This operator dictates how objective landscape information is acquired.

\textit{\textbf{1)} First-order:} $\mathcal{E}$ yields the standard stochastic gradient $g_t$.
\textit{\textbf{2}) Zeroth-order:} Bypassing backpropagation, $\mathcal{E}$ constructs gradients via finite-difference forward queries. For instance, using random directions $u_i \sim \mathcal{N}(0, I)$ and a smoothing radius $\mu$, it is typically formulated as $\tilde{g}_t^{\text{ZO}} = \frac{1}{q}\sum_{i=1}^{q}\frac{f(\theta_t+\mu u_i)-f(\theta_t-\mu u_i)}{2\mu}u_i$.

\textbf{Preconditioner $M_t$.} This matrix warps the gradient vector to accelerate convergence in ill-conditioned spaces.

\begin{table}[tp]\label{defination}
    \caption{
    \textbf{Nomenclature of key variables and hyperparameters in gradient-based optimization.} This table summarizes the unified mathematical notations consistently used throughout the paper.
    }
    \renewcommand{\arraystretch}{1.0}
    \setlength\tabcolsep{3.0pt}
    \resizebox{1.0\linewidth}{!}{
        \begin{tabular}{cl}
        \toprule[1.5pt]
        Symbol & Description \\
        \midrule
        \makecell[c]{$\theta_t$} & \makecell[l]{Model parameters (weights and biases) evaluated at iteration $t$}. \\
        \midrule
        \makecell[c]{$\eta$} & \makecell[l]{Learning rate, determining the step size of parameter updates}. \\
        \midrule
        \makecell[c]{$\nabla J(\theta_t)$} & \makecell[l]{Exact gradient of the objective function with respect to \\parameters at step $t$}. \\
        \midrule
        \makecell[c]{$g_t$} & \makecell[l]{Stochastic gradient (mini-batch gradient) of the objective \\function at step $t$}. \\
        \midrule
        \makecell[c]{$m_t$} & \makecell[l]{Biased first moment estimate (exponential moving average of \\past gradients)}. \\
        \midrule
        \makecell[c]{$v_t$} & \makecell[l]{Biased second raw moment estimate (exponential moving \\average of past squared gradients)}. \\
        \midrule
        \makecell[c]{$\beta_1, \beta_2$} & \makecell[l]{Decay rates for the first and second moment estimates}. \\
        \midrule
        \makecell[c]{$\epsilon$} & \makecell[l]{Small smoothing constant added to prevent division by zero}. \\
        \bottomrule[1.5pt]
        \end{tabular}
    }
    \label{tab:notation}
\end{table}

\textit{\textbf{1)} First-order adaptation.} $M_t$ is an operator or transformation matrix derived solely from first-order information, such as historical gradient statistics or the inherent geometric structure of the gradient matrix itself. Crucially, approaches utilizing empirical Fisher Information Matrix (EFIM) (~\cite{170}) are classified under this first-order category, as they inherently capture gradient variance rather than explicitly approximating true second-order curvature~\cite{345}. This spans from strict diagonal matrices (e.g., $M_t = \text{diag}(\sqrt{v_t} + \epsilon)$ in Adam~\cite{287}) to matrix-level structural normalizations under specific operator norms (e.g., spectral orthogonalization in Muon~\cite{173}), all operating without computing the true Hessian or FIM. 
\textit{\textbf{2)} Second-order algorithms.} $M_t$ explicitly incorporates high-order geometry, typically taking the form $M_t = \widehat{H}_t + \gamma I$, where $\widehat{H}_t \in \{\widehat{\nabla^2 f(\theta_t)}, \widehat{F}_t\}$ is an approximation of the Hessian or the FIM, and $\gamma$ is a damping factor.

\textbf{Scenario-aware transformation $\mathcal{T}_{\text{scenario}}(\cdot)$.} This non-linear operator introduces environmental constraints directly into the gradient flow, serving as the core mechanism for modern deployment challenges:

\textit{\textbf{1)} Privacy-preserving.} $\mathcal{T}(\tilde{g}_t) = \text{Clip}(\tilde{g}_t, C) + \mathcal{N}(0, \sigma^2 C^2 I)$, ensuring differential privacy via norm bounding and Gaussian noise injection.
\textit{\textbf{2)} Distributed optimization.} $\mathcal{T}(\tilde{g}_t) = \mathcal{C}(\tilde{g}_t)$, where $\mathcal{C}(\cdot)$ denotes a lossy compression operator (e.g., quantization, sparsification) to mitigate communication bottlenecks.

\textbf{Structural projection $\mathcal{P}_{\Theta}(\cdot)$.} In highly constrained scenarios, updates must be projected onto specific structural manifolds. For memory-efficient optimizers (e.g., low-rank training), $\mathcal{P}_{\Theta}$ restricts updates to a low-dimensional subspace to circumvent the storage overhead of dense optimizer states.

As a theoretical framework, ~\cref{eq:unified_master} is not intended as a universal convergence proof framework for all existing algorithms, but rather as a modular decoupling perspective. By abstracting complex optimization trajectories into four distinct structural operators ($\mathcal{E}, \mathcal{T}, \phi, \mathcal{P}$), this theoretical framework reveals structural equivalences across distinct optimization paradigms. Identifying these equivalences simultaneously exposes theoretical gaps within the design space. Researchers can use these insights to intuitively recombine mathematical primitives, guiding the algorithmic design of next-generation optimizers.\\
\textbf{Theoretical insights.} The unified formulation reveals non-trivial structural equivalences across distinct optimization paradigms. Examining the structural projection $\mathcal{P}_{\Theta}$ and the preconditioner $M_t^{-1}$ highlights a connection between memory-efficient ZO methods (such as LOZO~\cite{24}, which restricts $\mathcal{P}_{\Theta}$ to a low-rank subspace) and FO single-sided preconditioning methods (such as ASGO~\cite{176}, where $M_t$ uses low-rank gradient decomposition). In this framework, both approaches execute the same mathematical operation: they apply a dimensionality-reduced mapping on the $\mathbb{R}^d$ space to offset computational constraints on either the gradient estimator $\mathcal{E}$ or the scenario transformation $\mathcal{T}_{scenario}$. Rather than treating memory-efficient optimizers as isolated heuristics, this perspective systematically shows they operate as joint compromises between $M_t^{-1}$ and $\mathcal{P}_{\Theta}$.\\
\textbf{Systematic design space.} Furthermore, ~\cref{eq:unified_master} outlines a comprehensive design space that guides the development of new optimizers. Abstracting existing algorithms into these four operators reveals fundamental theoretical gaps. For example, although Kronecker-factored methods like Shampoo~\cite{170} are computationally efficient, their reliance on the empirical Fisher information restricts their preconditioner $M_t$ to capturing gradient variance rather than true geometric curvature. The proposed formulation suggests that this limitation can be addressed by redesigning the interaction between the gradient estimator and the preconditioner. A practical direction involves integrating memory-efficient gradient estimators for $\mathcal{E}$ (such as those utilizing model-distribution sampling to capture true Fisher information) with structured or localized preconditioners for $M_t$. Investigating such hybrid configurations provides a clear pathway to move beyond the variance-adaptation limits of empirical methods, enabling genuine curvature-aware optimization under strict memory constraints for large language models.

By viewing modern optimizers through the lens of~\cref{eq:unified_master}, we shift the narrative from isolated algorithmic tricks to a holistic engineering co-design. The subsequent sections will unfold how researchers systematically engineer $\mathcal{E}$, $M_t$, $\mathcal{T}$, and $\mathcal{P}$ to navigate the complex trade-offs among theoretical convergence, hardware efficiency, and systemic constraints.
\subsection{First-Order Algorithms}\label{fo_methods}
We define FO methods as parameter update schemes that construct an adaptive transformation or preconditioning operator $M_t$ by exclusively utilizing first-order gradient information, specifically including Empirical FIM, which capture historical gradient variance rather than explicitly approximating true second-order curvature. Despite their widespread adoption, vanilla stochastic gradient descent suffers from several fundamental limitations, including sensitivity to the choice of learning rate and instability under noisy gradients. To address these challenges, the research community has developed a rich ecosystem of fo variants that enhance the basic update rule across multiple complementary dimensions. This section systematically reviews these advancements, organizing them into eight core themes. Each theme represents a distinct evolutionary trajectory, yet together they form a coherent narrative of how fo algorithms have evolved to meet the escalating demands of contemporary deep learning architectures. We also trace the evolutionary trajectory of representative algorithms (\cref{fig:evolution_fo}) and systematically evaluate their strengths and limitations (\cref{tab:optim_method}).
\subsubsection{Accelerating Convergence Rate}
While momentum techniques significantly accelerate optimizer convergence speed, traditional momentum frameworks exhibit inherent limitations in complex optimization landscapes. The general mathematical framework of momentum methods typically formulates the update rule as $m_t = \beta m_{t-1} + \gamma g_t$ and $\theta_{t} = \theta_{t-1} - \eta m_t$. To transcend the boundaries of this standard formulation, recent studies explore several core improvement dimensions. These dimensions encompass the adjustment of momentum resetting and weighting, the enhancement of computational look-ahead and direction alignment, the introduction of dual momentum mechanisms, and the application of novel analytical perspectives from the frequency domain and dynamical systems.\\
\textbf{Scheduled momentum reset.} Several methodologies address issues such as update bias or overfitting to embedding layers in vanilla momentum methods under non-stationary objectives. These issues typically emerge from inappropriate accumulation of historical information. The unified formal approach involves active resetting or selective application of momentum states, conceptually modifying the update to $m_t = \gamma_t m_{t-1} + g_t$, where $\gamma_t$ acts as a binary or continuous decay trigger. For instance, SRSGD~\cite{10} resets the momentum buffer via linear or exponential scheduling to prevent the accumulation of the error of Nesterov accelerated gradient under stochastic conditions, thereby achieving simplicity and efficiency. Adam-Rel~\cite{201} targets non-stationarity in reinforcement learning by resetting time steps to balance momentum estimation, which avoids large updates caused by abrupt gradient changes. Furthermore, the LazyOptimizer~\cite{239} selectively updates active parameters through zero-gradient checks. The advantage of these methods lies in their capacity to enhance the adaptability of optimizers in dynamic tasks through adaptive momentum resetting. However, the inherent limitation of this category is the reliance on heuristic reset triggers, which may interrupt the continuous accumulation of effective historical gradients and necessitate meticulous hyperparameter tuning.\\
\textbf{Accelerated momentum.} Certain approaches surpass traditional momentum strategies by introducing complex momentum computations or look-ahead mechanisms to achieve faster convergence rates. From a unified derivation perspective, these methods estimate gradients at extrapolated future positions, denoted as $g_t(\theta_{t-1} - \eta \beta m_{t-1})$, or apply explicit bias correction to the standard formulation. Specifically, adaNAPG~\cite{40} and SGDO~\cite{242} focus on obtaining superior gradient estimates by looking ahead to future parameter configurations. The method adaNAPG~\cite{40} combines the accelerated gradient of Nesterov~\cite{324} with adaptive sampling to handle stochastic composite optimization with optimal complexity. SGDO~\cite{242} directly improves the update direction by computing gradients at future weights, achieving zero additional memory overhead. Conversely, RSGDM~\cite{252} and SQuARM-SGD~\cite{87} introduce correction mechanisms to rectify inherent biases in vanilla momentum. The momentum implementation of MADGRAD~\cite{171} eschews the exponential moving average of gradients, making it highly suitable for sparse models. RSGDM~\cite{252} applies exponential moving averages on both the gradient and the differential of the gradient to proactively correct bias and lag within SGDM~\cite{286}. SQuARM-SGD~\cite{87} integrates the momentum of Nesterov~\cite{324} with local stochastic gradient descent in distributed learning, where event-triggered communication acts as a conditional correction of communicated gradients. The primary advantage of this category is the attainment of higher-order convergence performance through look-ahead estimation. The corresponding limitation is the increased computational complexity per step and potential instability when future gradient approximations contain high variance.\\ 
\textbf{Double momentum mechanism.} To address the limitations of single momentum configurations in complex optimization scenarios, recent techniques employ multiple momentum terms or recursive estimators to capture gradient information comprehensively. The unified formal derivation entails the construction of a secondary momentum buffer or recursive estimator, formulated conceptually as $m^{(2)}_t = f(m^{(1)}_t, g_t)$, to reduce variance and bias systematically. The algorithm $\mu^2$-SGD~\cite{14} integrates anytime averaging with corrected momentum, achieving optimal convergence rates and enabling a fixed learning rate. AdEMAMix~\cite{253} utilizes a mixture of fast and slow exponential moving averages, adaptively balancing gradient relevance through schedulers to mitigate the forgetting of models during training. YOGI~\cite{153} employs an additive adaptive update for the second-moment accumulator with signed and bounded increments, which prevents abrupt increases in the effective learning rate. MARS~\cite{181} unifies variance reduction with adaptive methods via scaled stochastic recursive momentum. The advantage of double momentum mechanisms is the substantial improvement in optimization robustness and gradient estimation precision. The inherent limitation is the introduction of additional memory overhead to store multiple momentum states and the challenge of balancing the interaction between fast and slow moving averages.\\
\textbf{Momentum-gradient alignment.} Addressing the persistent zig-zag effect and suboptimal convergence paths in vanilla momentum methods, this category of research aligns momentum directions with current gradients. The formal mechanism operates by modulating the step size or update vector based on the cosine similarity or angular metric between the historical momentum $m_{t-1}$ and the current gradient $g_t$. AngularGrad~\cite{216} utilizes consecutive gradient angles to control the step size, thereby reducing zig-zag effects and ensuring smooth convergence paths. The Cautious Optimizer~\cite{235} modifies momentum methods by strictly aligning update directions with gradients, which preserves convergence guarantees and is implementable in a single line of code. The advantage of these alignment methods is the effective improvement of convergence smoothness and speed by preventing historical updates from deviating from current optimization trends. The limitation is that enforcing strict alignment may discard beneficial acceleration provided by historical momentum in narrow ravines, potentially slowing down the escape from shallow local minima.\\
\textbf{Dynamic momentum weight.} To overcome the constraints of fixed momentum coefficients, several studies dynamically adapt weights based on gradient variations. The unified derivation replaces the static coefficient with a dynamic function, computing the state as $m_t = \beta_t m_{t-1} + (1-\beta_t) g_t$, where $\beta_t$ adapts strictly to current gradient variations. For example, DEAM~\cite{186} computes adaptive weights using the discriminative angle between historical momentum and the current gradient, employs a backtrack term to restrict redundant updates, and reduces hyperparameters compared to the mechanism of Adam~\cite{287}. Furthermore, SGDF~\cite{129} leverages the principles of the Wiener filter to dynamically adjust gradient estimation, minimizing mean square error while balancing noise reduction and signal preservation. The advantage of dynamic weighting is the enhancement of convergence and generalization through tailored momentum adjustments. The primary limitation is the computational overhead required to evaluate the dynamic weight function at each iteration, along with the risk of vanishing momentum if the coefficient decays excessively.\\
\textbf{Frequency domain momentum analysis.} This dimension provides a novel perspective to interpret and optimize momentum methods via frequency domain insights. By conceptualizing the momentum operation as a time-varying filter $H(\omega, t)$ in the frequency domain, the framework presented in~\cite{251} demonstrates that high-frequency components become detrimental during the late stages of optimization, whereas low-frequency signals require amplification. FSGDM~\cite{251} dynamically adjusts momentum filtering characteristics based on these specific insights. The distinct advantage of this analytical paradigm is the ability to systematically isolate and suppress harmful noise frequencies. However, the limitation resides in the theoretical complexity of mapping time-domain gradients to frequency-domain filters in highly non-linear deep neural networks, which complicates practical hardware implementations.\\
\textbf{Momentum damping mechanism.} To resolve convergence instability and overshoot issues in large-scale machine learning tasks, researchers model the optimization process as a controlled physical or dynamical system. The unified formal approach involves introducing a damping factor or feedback controller, conceptually modifying the standard discrete update into a robust differential equation representation, such as $\ddot{\theta} + \gamma \dot{\theta} + \nabla f(\theta) = 0$. The PIDAO~\cite{301} framework and PID-based approaches~\cite{224} treat traditional SGDM~\cite{286} as a controlled heavy ball system, incorporating a PID controller driven by gradient history to suppress overshoot. This idea of dynamically regulating system states via feedback mechanisms finds a deeper physical extension in FANoS~\cite{308}, which formulates the momentum update as a discretized second-order dynamical system and introduces a thermostat as an integral controller. Furthermore, VRAdam~\cite{203} dampens excessive velocity via a dynamic learning rate, while AdamP~\cite{229} geometrically removes the radial momentum component responsible for unwarranted weight norm growth. SNGM~\cite{228} eliminates the dependence on batch size by normalizing gradients prior to momentum accumulation. The advantage of damping mechanisms is the critical stabilization of parameter evolution. The inherent limitation is the introduction of complex control dynamics that may require precise tuning of damping coefficients to prevent premature convergence.\\
Regarding the theoretical convergence analysis, the unified mathematical framework across these diverse dimensions typically guarantees optimal convergence complexities, such as $\mathcal{O}(1/\sqrt{T})$ for non-convex objectives, by strictly bounding the variance of momentum estimators. However, the theoretical boundaries and limitations of these methods often assume bounded gradients or specific smoothness conditions. These mathematical assumptions are frequently violated in the highly non-convex loss landscapes of modern deep learning architectures, which restricts the theoretical guarantees from perfectly translating into empirical success.

In terms of core design paradigms and evolution logic, the field has witnessed a milestone paradigm shift from static, heuristic momentum accumulation to dynamic, feedback-driven momentum modulation. This evolutionary trajectory transitions from basic scalar adjustments to complex geometric, physical, and frequency-domain interventions. Despite these comprehensive advancements, the core gap in existing research remains the lack of a universal, hyperparameter-free momentum framework. Current methodologies still struggle to automatically identify and adapt to the spectral properties of the loss landscape without incurring prohibitive memory overhead or computational costs. The future trajectory necessitates bridging the divide between rigorous continuous-time dynamical system models and the discrete, highly stochastic nature of empirical optimization.
\begin{table*}[t!]
\centering
\scriptsize
\setlength{\tabcolsep}{3pt} 
\setlength{\aboverulesep}{0pt}
\setlength{\belowrulesep}{0pt}
\renewcommand{\tabularxcolumn}[1]{m{#1}} 
\renewcommand{\arraystretch}{1.0} 
\caption{\textbf{Taxonomy and Comparison of Representative Optimization Methods.} This table presents a systematic evaluation of various optimizers categorized into FO, SO, and ZO paradigms, shows their limitations and highlights.}
\label{tab:optim_method}
\begin{tabularx}{\textwidth}{>{\centering\arraybackslash}m{2.5cm} >{\raggedright\arraybackslash}m{1.5cm} >{\raggedright\arraybackslash\hsize=0.6\hsize}X >{\raggedright\arraybackslash\hsize=0.9\hsize}X >{\raggedright\arraybackslash\hsize=1.5\hsize}X}
\toprule
\rowcolor{gray!30} \textbf{\textrm{Optimizer}} & \textbf{\textrm{Venue}} & \textbf{\textrm{Type}} & \textbf{\textrm{Limitation}} & \textbf{\textrm{Highlight}} \\
\midrule
\rowcolor{gray!20} \multicolumn{5}{c}{\textbf{\textrm{First-Order Methods}}  (~\cref{fo_methods})} \\
\midrule
SGDM \newline ~\cite{286} & Nature'1986 & Accelerating Convergence Rate & Complex hyperparameter tuning. & First proposed to introduce a momentum mechanism to accelerate convergence. \\
\rowcolor{gray!10} Adam \newline ~\cite{287} & ICLR'15 & Strategies for Adaptive Step-Size Estimation & High memory usage, worse generalization. & First proposed a method combining adaptive step-size estimation with momentum. \\
AdaBelief \newline ~\cite{149} & NeurIPS'20 & Strategies for Adaptive Step-Size Estimation & $\epsilon$ requires task-specific tuning. & Compared to adaptive methods like Adam, it achieves better generalization. \\
\rowcolor{gray!10} Shampoo \newline ~\cite{170} & ICML'18 & Approximating Curvature Information & Computationally expensive. & First proposed to exploit tensor structures to maintain preconditioners \\
4-bit Shampoo \newline ~\cite{172} & NeurIPS'24 & Approximating Curvature Information & Computationally expensive. & Compared to standard Shampoo, it significantly reduces memory usage by quantizing optimizer states to 4-bit. \\
\rowcolor{gray!10} AdaNorm \newline ~\cite{194} & WACV'23 & Enhancing training stability & Add computational overhead. & Compared to Adam, it proposes to solve the slow convergence issue caused by gradient anomalies through adaptively scaling updates. \\
AdaBound \newline ~\cite{226} & ICLR'19 & Learning Rate Scheduling & Convergence may slow in later stages. & Compared to Adam, it achieves SGD-level generalization capability through dynamic bound transitions. \\
\rowcolor{gray!10} SAM \newline ~\cite{131} & ICLR'21 & Enhancing Generalization Ability & Additional gradient computation steps. & First proposed to enhance model generalization by simultaneously minimizing loss value and sharpness. \\
GAM \newline ~\cite{134} & CVPR'23 & Enhancing Generalization Ability & Computationally expensive. & Compared to SAM, it proposes first-order flatness as a stronger measure to further enhance generalization. \\
\rowcolor{gray!10} AdaSGD \newline ~\cite{6} & arXiv'20 & Hybrid Methods & Inferior convergence speed compared to Adam. & First proposed to combine Adam's convergence speed with SGD's generalization via an adaptive global learning rate. \\
LOMO \newline ~\cite{17} & ACL'24 & Towards LLMs Training & Slow training throughput. & First proposed to reduce gradient memory usage in large model training by fusing gradient computation and parameter updates. \\

\midrule
\rowcolor{gray!20} \multicolumn{5}{c}{\textbf{\textrm{Second-Order Methods}}(~\cref{so_methods})} \\
\midrule
AdaHessian \newline ~\cite{258} & AAAI'21 & Hessian Approximation\&Estimation & Higher per-step computational cost and memory footprint. & First proposed to use Hutchinson's method to estimate the Hessian diagonal distribution, achieving superior convergence accuracy and speed. \\
\rowcolor{gray!10} K-FAC \newline ~\cite{255} & ICML'15 & Fisher Information Matrix Application & Higher computational cost. & First proposed factoring the FIM into Kronecker products, thereby achieving faster convergence. \\
MAC \newline ~\cite{277} & arXiv'25 & Fisher Information Matrix Application & Inferior accuracy relative to other second-order methods. & Compared to other second-order methods, it greatly simplifies curvature estimation, achieving memory usage and training speed comparable to SGD. \\
\rowcolor{gray!10} mL-BFGS \newline ~\cite{262} & TMLR'23 & Quasi-Newton Methods & Introduce additional memory and computational complexity. & Compared to standard L-BFGS, it effectively overcomes its instability by introducing a momentum mechanism. \\
Sophia \newline ~\cite{264} & ICLR'24 & Curvature-Guided Preconditioning & Lack the maturity. & Compared to traditional methods, it achieves faster convergence through curvature-guided preconditioning. \\
\rowcolor{gray!10} SGDHess \newline ~\cite{261} & NeurIPS'22 & Second-Order Moment Fusion & Higher per-step computational overhead and complexity. & First proposed to utilize Hessian-vector products to correct momentum bias in SGD, achieving a faster convergence speed. \\

\midrule
\rowcolor{gray!20} \multicolumn{5}{c}{\textbf{\textrm{Zeroth-Order Methods}}(~\cref{zo_methods})} \\
\midrule
ZO-AdaMM \newline ~\cite{285} & NeurIPS'19 & Adaptive Methods & Slow convergence speed. & First proposed to generalize the adaptive momentum method to zeroth-order optimization. \\
\rowcolor{gray!10} MeZO \newline ~\cite{16} & NeurIPS'23 & Perturbation Optimization & Require more iterations than first-order methods. & Compared to other methods, it adapts zeroth-order SGD to operate in-place, drastically reducing memory footprint. \\
Addax \newline ~\cite{52} & ICLR'25 & Zeroth-First Order Hybrid & Lower convergence speed compared to first-order methods. & First proposed to dynamically select between first-order and zeroth-order gradient computations, achieving faster convergence speed compared to pure zeroth-order methods. \\
\rowcolor{gray!10} LOZO \newline ~\cite{24} & ICLR'25 & Memory-efficient Methods & Exhibits a performance gap compared to first-order methods. & First proposed to leverage the intrinsic low-rank structure of gradients to achieve faster zeroth-order convergence. \\
MeZO-SVRG \newline ~\cite{19} & ICLR'24 & Variance Reduction & Exhibits a performance gap compared to first-order methods. & First proposed to incorporate SVRG into zeroth-order optimization, improving convergence stability and downstream accuracy. \\
\rowcolor{gray!10} ZoPro \newline ~\cite{93} & CDC'24 & Distributed Zero-Order Optimization & Theoretically only converges to a neighborhood of the optimum rather than the exact solution. & Compared to other zeroth-order methods, it incorporates gradient approximations into a distributed framework, achieving faster convergence. \\

\bottomrule
\end{tabularx}
\end{table*}
\subsubsection{Adaptive Step-Size Control}
Adaptive learning rates are widely utilized to reduce the cost of manual hyperparameter tuning. The generic update framework of these methods can be formulated as $\theta_{t+1}=\theta_t-\eta_t\phi(g_{1:t})/\psi(g_{1:t})$, where $\phi(\cdot)$ and $\psi(\cdot)$ denote the first-order momentum and the second-order preconditioner, respectively. Although the classic optimizer, such as Adam~\cite{287}, establishes the foundation for this paradigm, it exhibits limitations regarding extreme step sizes and non-convergence in complex landscapes. Subsequent advancements systematically evolve along structural dimensions: refining temporal moment estimation, expanding spatial adaptation granularity, decoupling architectural components, and minimizing state overhead. These developments share a unified objective of bounding the gradient variance while ensuring theoretical convergence under non-convex settings. To address the temporal dynamics of gradient sequences, optimizations primarily target the internal formulation of $\phi(\cdot)$ and $\psi(\cdot)$.\\
\textbf{Bias correction rules adaptation.} Initial training instabilities are mitigated by modifying the early-stage estimators. For instance, AdamD~\cite{217} removes the bias correction on the first-order momentum estimate to generate smaller initial updates, which stabilizes the warm-up period. While this strategy enhances early alignment with the gradient geometry, the inherent limitation lies in the marginal impact on late-stage convergence.\\
\textbf{Second-order moment adaptation.} To bound extreme step sizes and resolve the non-convergence of variance estimates, a multitude of strategies reformulate the exponential moving average of $\psi(\cdot)$. NosAdam~\cite{213} increases the weight of historical gradients, whereas ADOPT~\cite{196} eliminates the dependency on the decay rate $\beta_2$. Furthermore, AdamNX~\cite{290} adaptively adjusts the exponential decay rate, and SET-Adam~\cite{197} calibrates moments to prevent infinitesimally small step sizes. HomeAdam~\cite{340} removes the square-root in the second-order momentum to bound unstable step sizes and improve generalization. Additional variants incorporate differential privacy corrections in DP-AdamBC~\cite{62}, higher-order moments in HAdam~\cite{183}, sign-power transformations in AdamPower~\cite{245}, dynamic weighting in MADAM~\cite{188}, direct variance adaptation in M-SVAG~\cite{212}, probabilistic generalization in VSGD~\cite{198}, and memory-efficient factorization in AdaLomo~\cite{18}. The primary advantage of these methods is the rigorous stabilization of convergence trajectories; however, they inherently introduce heightened sensitivity to auxiliary hyperparameters.\\
\textbf{Dynamic epsilon adjustment.} To further refine the preconditioner without auxiliary overhead, algorithms dynamically adjust the stabilization term $\epsilon$. EAdam~\cite{214} applies the additive term directly to the uncorrected accumulators to implicitly reduce the step size for small moments, outperforming the baseline methods~\cite{287,160,24} across diverse domains. Similarly, FAdam~\cite{219} integrates adaptive adjustments through natural gradient corrections based on Riemannian geometry. These approaches effectively prevent division by zero anomalies, yet their theoretical bounds often rely on empirical approximations.\\
\textbf{Prediction deviation adaptation.} By substituting the standard gradient variance with the discrepancy between observed and predicted gradients, AdaBelief~\cite{149} scales the update directions to achieve rapid convergence and strong generalization. Aida~\cite{151} extends this belief-based scaling by utilizing first-momentum projections for second-momentum estimation to strictly suppress step size ranges. This framework optimally couples the learning rate with gradient uncertainty, although it remains susceptible to highly noisy stochastic batches.\\
\textbf{Momentum-based adaptation.} The integration of momentum mechanisms modifies $\phi(\cdot)$ to accelerate convergence through phase-shifted accumulations. Nadam~\cite{155} and Adan~\cite{168} incorporate Nesterov~\cite{324} momentum, while RAdam~\cite{160} rectifies the variance of early updates. Other structural integrations include utilizing extrapolated first moments in Adam+~\cite{161}, quasi-hyperbolic terms in QHAdam~\cite{159}, and adapted rate models in MoMo-Adam~\cite{243}. Conversely, Lion~\cite{167} isolates the sign topology of the momentum to eliminate second-order dependencies. Further augmentations involve second-order injections in AdaInject~\cite{127}, alternating curvature thresholds in EAGLE~\cite{180}, recursive scaling in MARS~\cite{181}, and Newton-Schulz soft-thresholding in ROOT~\cite{291}. INNAprop~\cite{274} additionally merges structural information with RMSprop~\cite{325}. The collective advantage of these methods is the significant acceleration of the optimization trajectory, but the inherent limitation is the accumulation of historical momentum may become dynamically inconsistent with the current gradient field, thereby introducing update bias in complex or non-stationary loss landscapes.

Beyond temporal accumulation, the spatial granularity of adaptation shifts from scalar parameter-level updates to broader structural geometries.\\
\textbf{Layer-wise adaptation.} Uniform learning rates frequently fail to accommodate the structural heterogeneity and anisotropic scaling of deep networks. To facilitate massive batch training, LARS~\cite{223} and LAMB~\cite{227} scale the updates of AlexNet~\cite{335}, ResNet-50~\cite{297}, and BERT~\cite{338} by applying layer-wise normalization with non-convex convergence guarantees. NovoGrad~\cite{157} halves the memory requirements through layer-wise norms while decoupling the weight decay, and AdaL~\cite{215} transforms raw gradients prior to accumulation. Coupled Adam~\cite{48} resolves the anisotropic embeddings of Adam by enforcing a uniform second moment. Transitioning to matrix-wise orthogonalization, Muon~\cite{173} utilizes Newton-Schulz iterations to standardize the spectral norm of updates across layers. CaAdam~\cite{221} further integrates structural depth and connectivity through multi-strategy scaling. These layer-wise normalizations effectively resolve gradient vanishing issues in heterogeneous architectures, yet they fundamentally assume that the gradient distributions within a single layer are uniformly bounded. 
\textbf{Neuron-level adaptation.} At a finer granularity, AdaAct~\cite{237} adjusts the gradient updates inversely with the square root of the activation variance. By utilizing the exponential moving average of activation variances rather than gradient variances, it shares the learning rates across parameters that process identical input features. This approach isolates the variance at the activation level to maximize output stability, though the inherent limitation is the substantial memory footprint required for networks with expansive width.

To synthesize the complementary strengths of isolated paradigms, contemporary methods decouple orthogonal components and hybridize discrete algorithms.\\
\textbf{Decoupled learning rate and adaptability.} Standard optimizers frequently conflate the global step size with local adaptivity. AvaGrad~\cite{150} explicitly decouples these components to optimize performance across diverse modalities, offering superior task-specific tuning capabilities. While this isolation prevents the degradation of the global convergence rate, it inherently expands the hyperparameter search space.\\
\textbf{Hybrid adaptive strategy.} By fusing disparate adaptation criteria, hybrid models establish generalized optimization frameworks. AdaSGD~\cite{6} merges the uniform trajectory of SGD~\cite{307} with adaptive scaling, while EXAdam~\cite{206} introduces debiasing terms for moment interactions. Memory reduction is achieved through structured projections in APOLLO~\cite{43}. Furthermore, theoretical bounds are fortified by p-th order momentums and Nesterov~\cite{324} acceleration in S3~\cite{44}, asynchronous second-moment centering in ACProp~\cite{230}, and gradient difference step sizing in DiffGrad~\cite{184}. Parameter regularization is isolated in AdamW~\cite{148}, and the adaptivity is interpolated with signed methods in LaProp~\cite{156}. Finally, AdaMuon~\cite{178} incorporates variance adaptivity into orthogonal updates, and AdaFamily~\cite{218} establishes a continuous algorithmic spectrum via hyperparameter blending. The primary advantage of hybridization is the simultaneous mitigation of multiple failure modes, but the inherent limitation is the resultant computational complexity and the intractability of theoretical convergence analysis.

As model dimensions scale exponentially, the memory overhead of maintaining historical states necessitates optimal approximation strategies.\\
\textbf{Stateless adaptation.} To eliminate the redundant storage of second-order momentum, AlphaGrad~\cite{204} employs L2 normalization and hyperbolic transformations supported by non-convex proofs. AutoDrop~\cite{234} leverages angular velocity saturation to adjust rates without supplementary hyperparameters. AdamS~\cite{33} replaces the temporal variance with a momentum-current gradient denominator, thereby matching the memory efficiency of SGD while inheriting the configuration of AdamW~\cite{148}. AEGDM~\cite{126} merges momentum with transformed gradient sums to ensure energy stability over baseline methods~\cite{286,287}. These designs approach the memory limits of first-order methods while retaining adaptivity, but they strictly rely on the assumption of locally stationary gradient distributions. \\
\textbf{Kalman filtering based method.} RLEKF~\cite{254} reorganizes network layers to accommodate the covariance tracking of Kalman filtering, approximating the dense error matrix with a sparse diagonal block structure. This efficiently utilizes second-order covariance information; however, the truncation errors of the sparse approximation inevitably accumulate across deep architectures.

The evolution of adaptive step-size control demonstrates a definitive paradigm shift from temporal scalar correction to spatial structural optimization, and ultimately towards memory-efficient geometric approximations. Early methodologies predominantly conceptualized the optimization landscape as an independent sequence of historical gradients, focusing on bounding the variance through first-order and second-order moment refinements. As network architectures increased in depth and heterogeneity, the paradigm transitioned to layer-wise and neuron-level normalizations. Currently, the trajectory converges upon stateless and hybrid adaptations, prioritizing the mathematical decoupling of momentum, variance, and weight decay to maximize the efficiency of large-scale pre-training. Despite these milestones, a critical gap remains in the existing literature: the theoretical convergence guarantees of hybrid and structurally adaptive methods largely depend on strict assumptions of convexity or bounded smoothness, which fail to accurately characterize the highly non-convex and singular optimization landscapes of modern deep neural networks.
\subsubsection{Variance Adaptation}
Classic FO algorithms flatten parameters into vectors, thereby discarding inherent structural information. In contrast, preconditioning methods formulate the optimization update through a unified mathematical framework, typically expressed as $\theta_{t+1} = \theta_t - \eta P_L \nabla J(\theta_t) P_R$, where gradients are multiplied by approximations often derived from the empirical FIM. This formulation can be uniformly incorporated into the aforementioned framework~\cref{eq:unified_master}, where the preconditioner $M_t^{-1}$ is instantiated as the left-right multiplication operator $P_L \cdot P_R$. While traditionally motivated as achieving curvature adaptation by approximating the Hessian, recent perspectives suggest that the EFIM primarily captures the non-central second moment of gradients~\cite{345}. Therefore, these preconditioning techniques are increasingly understood as performing variance adaptation, mitigating gradient noise in stochastic optimization, while preserving the structural information of parameters. The core improvement dimensions of these methods involve efficiently constructing these preconditioning matrices while minimizing computational overhead, ultimately balancing optimization efficiency with potential generalization. The evolutionary logic of this field transitions from complex dimension-wise preconditioning to highly efficient single-metric approximations to mitigate severe resource bottlenecks.\\
\textbf{Preconditioners based on two metrics.} To capture intricate structural information, the initial design paradigm utilized two distinct metrics, typically approximating the row and column covariance matrices. Shampoo~\cite{170} pioneers an online structure-aware algorithm by maintaining separate dimension-wise preconditioners constructed from second-order gradient statistics. This formulation achieves faster convergence compared to traditional optimizers while preserving a comparable per-step runtime. Building upon this framework, 4-bit Shampoo~\cite{172} adopts eigenvector matrix quantization technology to reduce memory costs. PSGD~\cite{306} introduces a noise-robust criterion for preconditioner estimation that equilibrates the perturbations of preconditioned gradients with the perturbations of parameters. Additionally, SPlus~\cite{175} enhances the stability of the framework of Shampoo~\cite{172} through instant-sign normalization and iterate averaging. It incorporates shape-aware scaling for adaptation to network width and combines bounded updates with historical eigenbases to ensure stability even under lower-frequency matrix inversions. Theoretically, these dual-metric preconditioners efficiently approximate natural gradient descent, accelerating convergence in highly non-convex landscapes. Nevertheless, their boundary of applicability is constrained by the substantial memory overhead required to store and invert covariance matrices for large-scale models.\\
\textbf{Preconditioners based on a single metric.} To address the severe resource demands, subsequent research evolved to leverage a unified single metric. Within this unified formulation, algorithms adjust the update step based on a single structural proxy to balance efficiency and performance effectively. For instance, ASGO~\cite{176} employs a single-side preconditioner that preserves the matrix structure, utilizing low-rank gradients to reduce computational overhead while maintaining convergence properties compared to the method of~\cite{170}. Similarly, NYSACT~\cite{269} applies an eigenvalue-shifted Nyström approximation for scalable covariance estimation, bridging the gap between FO and SO algorithms through improved estimation accuracy and reduced resource consumption. Furthermore, the method of Hessian-aware scaling~\cite{247} adjusts gradients based on local curvature to guarantee a unit step size. Theoretical convergence analysis indicates that these methods effectively improve the condition number of the optimization landscape, facilitating faster convergence under convex assumptions with a significantly lower memory footprint. However, the inherent limitation of single-metric approaches resides in their reduced capacity to capture the complex inter-dimensional correlations fully represented in dual-metric designs.

The core design paradigm of curvature approximation has evolved from Kronecker-factored dimension-wise preconditioning back to streamlined single-matrix preconditioning, driven by the critical need for extreme scalability in modern architectures. The advantage of methods based on two metrics is their comprehensive structural awareness, but their inherent limitation is the high memory footprint. Conversely, the advantage of methods based on a single metric is their low computational barrier and memory efficiency, yet they suffer from limited curvature representation. The core gap in existing research remains the lack of fully dynamic and hardware-friendly preconditioners that can achieve the representational capacity of dual-metric methods while maintaining the algorithmic efficiency of single-metric methods, without requiring complex hyperparameter tuning or suffering from precision degradation during aggressive quantization.
\subsubsection{Enhancing Training Stability}
To address gradient-related optimization challenges such as gradient explosion and variance imbalance, contemporary stabilization methods can be unified under a general framework that modulates the gradient update step via functional transformations. The core improvement dimensions of these transformations are primarily categorized by processing granularity and statistical adaptation, evolving from static constraints to dynamic, distribution-aware calibrations.\\
\textbf{Basic fixed gradient clipping.} Bounding gradient magnitudes provides a fundamental mechanism to stabilize convergence under heavy-tailed noise and satisfy constraints such as privacy preservation. Within this paradigm, Clipped-SGD~\cite{7} is an accelerated stochastic method utilizing fixed gradient clipping to suppress heavy-tailed noise. Furthermore, DP-SGD~\cite{59} integrates gradient clipping with Gaussian noise injection for differentially private non-convex optimization. A unified characteristic of these approaches is the application of a predefined scalar threshold to restrict the gradient norm. The primary advantage of this category is the provision of robust noise resilience and strict privacy bounds. However, the inherent limitation lies in the reliance on manually tuned static thresholds, which often induce gradient bias and hinder optimal convergence.\\
\textbf{DP-enhanced gradient clipping.} To address the biased optimization trajectory caused by rigid truncation, recent methods augment traditional clipping with error correction frameworks. DiceSGD~\cite{63} introduces an error feedback mechanism to eliminate clipping bias in differentially private stochastic gradient descent~\cite{307}. This formulation allows for flexible clipping thresholds independent of problem-specific parameters while maintaining rigorous utility and privacy guarantees through updated convergence analysis. The distinct advantage of this method is the theoretical decoupling of privacy thresholds from optimization utility. The inherent limitation is the increased computational overhead required to persistently track and integrate error feedback states.\\
\textbf{Dynamic gradient clipping.} Transitioning from static to adaptive thresholds, dynamic clipping adjusts constraints based on historical gradient statistics to accommodate varying parameter sensitivities and sudden gradient spikes. Stable-SPAM~\cite{28} augments Adam~\cite{287} with AdaClip to apply entry-wise clipping derived from exponential moving averages. Similarly, AdaGC~\cite{30} employs per-parameter moving average thresholds, maintaining broad compatibility with existing optimizers~\cite{287}~\cite{149}. Both methods share a unified derivation relying on historical momentum to dictate local clipping boundaries. The advantage of dynamic adaptation is the universal applicability across diverse architectures and modalities. Conversely, the inherent limitation is the potential lag in threshold adjustment during rapid topological shifts in the loss landscape.\\
\textbf{Spike-aware gradient clipping.} For highly specific non-convex scenarios where standard dynamic adjustment fails to react to extreme anomalies, specialized clipping targets abrupt variations. SPAM~\cite{46} integrates spike-aware clipping mechanisms directly into the Adam~\cite{287} optimizer to mitigate the severe impact of gradient spikes. The advantage of this approach is the targeted enhancement of large-scale training stability against sparse anomalies. The inherent limitation is its narrow applicability, as the mechanism is predominantly effective only in regimes exhibiting extreme gradient variance.\\
\textbf{Element-wise gradient scaling.} Moving beyond magnitude truncation, fine-grained scaling resolves training instability by individually adjusting each gradient element, thereby addressing the limitations of uniform scaling. SPAM~\cite{46} integrates momentum reset, spike-aware clipping, and sparse momentum into Adam to mitigate spikes and optimize memory consumption. AdaNorm~\cite{194} utilizes exponential moving averages of past norms to correct individual gradients, ensuring optimization consistency. Additionally, pbSGD~\cite{3} applies powerball functions for flexible scaling, providing theoretical convergence guarantees for non-convex objectives. The advantage of element-wise scaling is the maximum flexibility afforded to individual parameter updates. The inherent limitation is the substantial memory footprint required to maintain distinct statistical states for every parameter.\\
\textbf{Layer-wise gradient normalization.} To balance optimization stability with memory efficiency, layer-wise methods normalize gradients at macroscopic group levels. Stable-SPAM~\cite{28} incorporates AdaGN to compute moving averages of per-layer gradient norms, rescaling current gradients by the ratio of the historical mean to the root second moment to suppress norm explosions. MultiAdam~\cite{193} categorizes loss terms into separate groups and maintains second-order momentum individually to balance diverse objectives, improving convergence in systems such as physics-informed neural networks. AuON~\cite{294} automatically suppresses unstable updates by normalizing the entire momentum matrix. The advantage is the effective improvement of convergence in specialized multi-objective scenarios with reduced memory overhead. The inherent limitation is the assumption of uniform gradient behavior within a layer, which may inadvertently suppress heterogeneous feature learning.\\
\textbf{Mean-removal normalization.} Targeting gradient noise and non-smooth optimization trajectories, statistical centering stabilizes updates by shifting gradients or corresponding moments. GCSAM~\cite{142} integrates gradient centralization into the ascent step of sharpness-aware minimization~\cite{131} to reduce noise and computational overhead while enhancing generalization. AdamMC~\cite{190} normalizes the first-order moments of adaptive optimizers to yield shorter and smoother optimization paths. These methods share a unified core of mean-removal but operate on different analytical components. The advantage of centralization is the improved smoothness of the training path and enhanced generalization. The inherent limitation is the assumption of symmetrically distributed gradient noise, which can degrade performance under heavily skewed stochastic distributions.\\
\textbf{Noise-robust normalization.} To ensure convergence consistency across varied modalities under atypical gradient conditions, advanced normalization techniques integrate comprehensive historical statistics. AdaGC~\cite{30} utilizes exponential moving averages to adaptively bound parameters while maintaining standard non-convex convergence rates comparable to Adam~\cite{148}~\cite{167}. AdaNorm~\cite{194} corrects gradients through historical norm averages, creating variants that counteract atypical updates. Furthermore, NIGT~\cite{130} modifies the momentum formulation to accelerate convergence on second-order smooth objectives, effectively eliminating the reliance on large batch sizes. The advantage of noise-robust methods is the mitigation of batch size constraints and universal stabilization across tasks. The inherent limitation is the complexity of theoretical convergence analysis, which often requires strict assumptions regarding the underlying landscape smoothness.

The evolutionary logic of these stability enhancement techniques reveals a clear milestone paradigm shift: progressing from rigid constraints to granular rescaling, and ultimately culminating in adaptive statistical calibrations. While existing literature provides a robust empirical foundation for gradient manipulation, a fundamental gap persists in bridging local gradient corrections with global topological guarantees. Current methodologies operate predominantly as reactive mechanisms based on historical observations. The critical missing link is a proactive, unified mathematical framework that inherently couples stability constraints with the intrinsic geometry of the loss landscape, thereby eliminating the reliance on auxiliary tuning parameters and heuristic statistical tracking.\\
\subsubsection{Learning Rate Scheduling}
Learning rate scheduling constitutes a fundamental mechanism to dynamically adjust the step size of the optimizer, balancing the speed of convergence and the stability of model training across distinct optimization phases. The general mathematical framework of these methods can be formulated as $\theta_{t+1} = \theta_t - \eta_t \Phi(g_{1:t}, \mathcal{H}_t)$, where $\Phi$ denotes the transformation function conditioned on historical gradients and structural priors. This formulation aligns with the unified perspective in~\cref{eq:unified_master} by absorbing the momentum and preconditioning operations into $\Phi$, and specializing to the case of identity projection and no explicit weight decay. The core improvement dimensions of recent studies focus on the scheduling basis, the granularity of adaptation, the reduction of manual dependency, and the enhancement of theoretical stability. Through targeted strategies, these dimensions collectively improve training efficiency and generalization.\\
\textbf{Lightweight scaling and batch-aware scheduling.} Addressing computational efficiency and batch dynamics, this paradigm optimizes scaling properties and layer-wise learning rates. SGD-SaI~\cite{13} leverages a scaling mechanism for the initial learning rate based on the signal-to-noise ratio of gradients. This approach eliminates the necessity for complex adaptive methods by maintaining constant adjustments to the signal-to-noise ratio, which reduces memory consumption by half compared to previous approaches while preserving competitive performance. Furthermore, addressing the critical challenge of scaling batch sizes for faster deep learning training, LAMB~\cite{227} utilizes layer-wise adaptive learning rates. This batch-aware scheduling overcomes a key barrier to efficient large-scale deployment by preventing the degradation of model performance in tasks requiring both rapid convergence and high accuracy. The advantage of this category lies in its exceptional memory efficiency and computational simplicity, while the inherent limitation is the reliance on specific empirical initial values that may require careful tuning.\\
\textbf{Element adaptation and gradient angle scheduling.} Progressing towards finer granularity and directional awareness, the second paradigm dynamically calibrates step sizes using dimension-level constraints and angular information. AdaBound~\cite{226} applies dynamic bounds to element-wise learning rates to mitigate the adverse effects of extreme step sizes. This strategy balances the rapid initial training of adaptive methods and the superior generalization of SGD~\cite{307}, supported by rigorous theoretical convergence proofs. Simultaneously, to address suboptimal direction alignment and oscillatory updates in high-dimensional scenarios, ACMo~\cite{189} operates as a first-moment optimizer that eschews second-moment estimates. It employs angle calibration to ensure that descent directions form acute angles with both current and historical gradients, which delivers convergence rates comparable to the methods of the Adam family. Similarly, HGM~\cite{207} introduces a hindsight mechanism to evaluate the cosine similarity between current gradients and accumulated momentum. It increases learning rates in coherent gradient regions and decreases them in oscillatory areas, retaining an efficiency similar to Adam~\cite{287}. Furthermore, AdaBFE~\cite{191} extends binary forward exploration by adapting learning rates per parameter dimension utilizing the information of the forward loss function. The advantage of these methods lies in their ability to correct suboptimal update directions, whereas the inherent limitation is the increased computational overhead for continuous angular or metric evaluation.\\
\textbf{Loss-sensitive and stability-aware scheduling.} To further enhance the robustness of optimization in complex non-convex topologies, subsequent paradigms integrate loss landscape information and rigorous stability controls. DecGD~\cite{238} decomposes gradients into surrogate gradients and vectors based on the loss function, adjusting learning rates according to loss information rather than squared gradients to achieve rapid convergence and better generalization. Addressing the critical challenge of maintaining optimization stability in scenarios where instability from large step sizes or stochastic noise causes deviation, LyAm~\cite{210} integrates the adaptive moments of Adam with stability mechanisms based on the Lyapunov function. This integration dynamically adjusts learning rates, enhancing robustness against noise while providing guarantees for non-convex optimization. Multistage SGDM~\cite{1} also utilizes a Lyapunov function to govern momentum deviation, enabling larger initial step sizes and smaller subsequent ones while matching the convergence rates of SGD. Additionally, the analysis of large step size SGD~\cite{2} models the dynamics through stochastic differential equations, revealing loss stabilization and implicit sparsity emerging from multiplicative noise. These methods provide the advantage of theoretical rigor and enhanced stability, but they suffer from the inherent limitation of relying on stringent theoretical assumptions that may not perfectly align with highly stochastic real-world data distributions.\\
\textbf{Scheduler-free adaptation.} The most recent milestone paradigm shift focuses on the complete elimination of manual learning rate scheduling through data-driven metrics and unified theoretical models. Schedule-Free~\cite{233} unifies scheduling with momentum interpolation. AutoDrop~\cite{234} utilizes angular velocity saturation without requiring extra tuning. Other parameter-free or automated variants include AdaS~\cite{9}, which utilizes knowledge gain per block, and SGD-G2~\cite{11}, which combines the Runge-Kutta method with adaptive SGD~\cite{307}. Furthermore, AutoSGD~\cite{15} evaluates neighboring rates, Step-Tuned SGD~\cite{125} derives the step size directly from gradients, and Amos~\cite{192} incorporates the scale of the model to reduce memory usage. Adam++~\cite{199} also functions as a parameter-free variant. These approaches are broadly categorized into theoretical groups~\cite{233},~\cite{11},~\cite{125},~\cite{199} and empirical groups~\cite{234},~\cite{9},~\cite{15},~\cite{192}. The primary advantage is the eradication of costly hyperparameter searches, while the inherent limitation is the potential loss of fine-grained control required for highly specialized architectures.

The evolutionary logic of learning rate scheduling demonstrates a transition from manual heuristics toward automated, theoretically grounded, and dimension-wise dynamic calibrations. The formal convergence analysis of these representative algorithms relies heavily on bounding the expected descent in non-convex settings, often utilizing Lyapunov functions or angle-preservation constraints to guarantee that the accumulation of gradient noise does not diverge. Despite these advancements, a critical gap remains in the existing literature: current fully automated scheduling methods frequently struggle to balance the rigorous theoretical guarantees of stability-aware methods with the low computational complexity demanded by extremely large-scale foundation models. Bridging the gap between the theoretical properties of scheduler-free mechanisms and the empirical noise characteristics of massive batch sizes constitutes the primary trajectory for future research.
\subsubsection{Enhancing Generalization Ability}
A fundamental challenge in modern deep learning is that traditional optimizers frequently converge to sharp minima, which significantly degrades generalization. To address this issue, numerous methods proactively seek flat minima by formulating a minimax optimization problem. The general mathematical framework, pioneered by the foundational SAM~\cite{131} algorithm, aims to minimize the maximum loss within a local neighborhood, formulated as $\min_{w} \max_{\|\epsilon\| \le \rho} J(w + \epsilon)$. This minimax objective can be realized within the unified framework~\cref{eq:unified_master} by defining the gradient estimator $\mathcal{E}$ to incorporate the worst-case perturbation: $\tilde{g}_t = \nabla J(\theta_t + \epsilon_t; \xi_t)$ with $\epsilon_t = \arg\max_{\|\epsilon\|\le \rho} J(\theta_t + \epsilon; \xi_t)$ approximated via one-step gradient ascent. Building upon this theoretical foundation, subsequent research advances optimization theory through core improvement dimensions, including step enhancement, gradient normalization, curvature guidance, momentum adaptation, noise utilization, and averaging strategies. These methods collectively construct a unified theoretical paradigm that balances the convergence rate and generalization bounds in complex non-convex landscapes.\\
\textbf{Fundamental framework of SAM.} The primary objective of the vanilla SAM~\cite{131} algorithm is to enhance generalization by concurrently optimizing the value of the loss and the sharpness of the loss. Because standard optimizers often overfit to sharp regions, the original framework formulates a minimax problem to improve robustness against label noise and to elevate benchmark performance. To mitigate the high computational overhead of the inner maximization step, subsequent variants focus on efficiency and resource utilization. ESAM~\cite{132} reduces the computational burden by utilizing stochastic weight perturbation and data selection based on sharpness. AsyncSAM~\cite{143} achieves near-zero additional overhead through asynchronous background perturbations utilizing prior gradients. Furthermore, AE-SAM~\cite{135} adaptively switches between SAM~\cite{131} and standard empirical risk minimization to decrease the update frequency of the perturbation while preserving the theoretical convergence rate. Methods such as SAMPa~\cite{136} and LightSAM~\cite{144} utilize adaptive optimizers to adjust the radius and rate independent of specific parameters. The primary advantage of this category is the direct optimization of generalization bounds, whereas the inherent limitation remains the fundamental trade-off between exact sharpness estimation and computational efficiency.\\
\textbf{Renormalized gradient norm adaptation.} To address the instability caused by unconstrained perturbation scales, some approaches refine the fundamental formulation of SAM through gradient norm renormalization. SSAM~\cite{141} introduces a lightweight renormalization step that rescales the gradient of the inner ascent step to strictly match the gradient norm of the descent step. This strategy prevents overly large perturbations without introducing additional hyperparameters or incurring significant computational costs, while preserving the foundational two-step structure of the optimization process. Supported by rigorous theoretical bounds, this method improves the training stability of the model. The prominent advantage of this paradigm is the stabilization of optimization trajectories in highly non-convex regions, whereas the inherent limitation is the potential under-exploration of the loss landscape when the descent gradient norm is exceptionally small.\\
\textbf{Multi-step Ascent Optimization.} Addressing the limitation of insufficient maximization in the vanilla formulation of SAM~\cite{131}, multi-step ascent methods incorporate multiple iterations to strengthen the robustness of the optimization process. Lookbehind-SAM~\cite{138} executes multiple ascent steps to compute a more precise approximation of the local maximum. The algorithm utilizes linear interpolation to aggregate gradients from these individual steps, which stabilizes the subsequent minimization phase. This approach improves robustness against noisy weights. The core advantage is the high precision of landscape exploration, but the inherent limitation is the linear increase in computational complexity scaling with the number of internal ascent steps.\\
\textbf{Curvature-guided Landscape Exploration.} Beyond first-order approximations, several methods extract curvature-related information to navigate complex loss landscapes. GAM~\cite{134} integrates first-order flatness to improve generalization performance. MIAdam~\cite{200} incorporates a multiple integral term into the standard Adam to dynamically filter sharp minima, which maintains the convergence speed while improving the generalization of the model. Furthermore, DEO~\cite{209} adapts the Dimer method to estimate the smallest eigenvector of the Hessian matrix via gradient differences, projecting updates orthogonally to the direction of minimum curvature to escape saddle points efficiently. SKA-SGD~\cite{240} projects gradients onto Chebyshev-basis Krylov subspaces using streaming Gauss-Seidel iterations, which eliminates the need for full Gram matrix computations and reduces the operational complexity. The main advantage of curvature-guided methods is their robust theoretical convergence in ill-conditioned problems, whereas the inherent limitation is the persistent difficulty of accurately estimating higher-order geometry in high-dimensional spaces.\\
\textbf{Momentum Landscape Adaptation.} To resolve the strict trade-off between the performance of optimization and computational overhead, momentum-based adaptations modify the traversal of the loss landscape. MSAM~\cite{137} perturbs parameters along accumulated momentum vectors rather than computing exact gradients for the ascent step. This utilizes momentum as an approximation for sharpness computations, which removes the necessity for extra forward and backward passes. SCSAdamW~\cite{39} integrates stochastic conjugate subgradients with techniques from~\cite{148}, utilizing adaptive sampling and theoretical sample complexity analysis to dynamically adjust training batch sizes. The advantage of these methods lies in achieving sub-linear or near-zero extra computational costs, while the inherent limitation is the deterioration of worst-case theoretical bounds due to the reliance on delayed historical approximations.\\
\textbf{Noise Injection Enhancement.} Parallel to momentum strategies, other methods leverage stochastic gradient noise to mitigate the computational burden of exact perturbations. F-SAM~\cite{139} filters full gradient components via the exponential moving average of historical stochastic gradients, utilizing the residual noise to mitigate suboptimal optimization trajectories. FGSAM~\cite{140} employs graph neural networks to generate the perturbation of SAM~\cite{131} and multi-layer perceptrons to execute the minimization efficiently. The advantage of noise injection is the intrinsic improvement of efficiency and implicit regularization, whereas the inherent limitation is the potential instability introduced during the late stages of convergence.\\
\textbf{Weight Averaging Strategies.} As a complementary paradigm to explicit perturbation, weight averaging methods operate on the macroscopic trajectory of the optimization process. Lookaround~\cite{128} introduces an iterative weight averaging strategy deployed continuously throughout the training phase to balance state diversity and local convergence. The formal derivation consists of two distinct phases: the algorithm first trains multiple parallel network instances using diverse data augmentations, and subsequently computes the arithmetic mean of the weights from these networks to form the updated central model. This paradigm aggregates diverse local optima to naturally approximate a flatter global minimum. The advantage of this approach is the consistent improvement of generalization without altering the inner optimization step, whereas the inherent limitation is the substantial memory footprint required to maintain and synchronize multiple network states.

The trajectory of generalization enhancement methods demonstrates a clear evolutionary logic. The foundational paradigm transitioned from the direct formulation of minimax optimization to the stabilization of gradient norms. Subsequently, the field evolved toward high-precision landscape exploration via multi-step and curvature-guided techniques. To bridge the gap between theoretical rigor and practical deployment, the paradigm shifted toward efficient approximations utilizing momentum, noise injection, and macroscopic weight averaging. Despite these advancements, a critical theoretical gap remains unresolved. Current literature lacks a unified mathematical framework capable of rigorously bounding the generalization error and convergence rate when dynamic, low-rank, or asynchronous approximations are deployed in highly non-convex, large-scale distributed training environments.
\subsubsection{Hybrid Methods}
Hybrid optimization methods overcome the inherent limitations of single-approach strategies by integrating complementary techniques to balance multiple optimization objectives, such as convergence speed, stability, generalization, and resource efficiency. The general mathematical framework of these methods can be abstracted as $\theta_{t+1} = \mathcal{P}_{\mathcal{X}}(\theta_t - \eta_t \mathcal{H}(\nabla J(\theta_t), m_t, v_t))$, where $\mathcal{H}$ denotes a unified hybrid operator merging momentum $m_t$, and variance $v_t$, while $\mathcal{P}_{\mathcal{X}}$ represents constraint projections or spatial smoothing operators. By selectively combining the strengths of different mechanisms, these methods establish a unified derivation form that systematically addresses diverse algorithmic bottlenecks.\\
\textbf{SGD-Adam hybrid.} The foundational step in hybrid design involves bridging the gap between stochastic gradient descent and adaptive methods. For example, AdaSGD~\cite{6} introduces a modified form of SGD~\cite{307} that adapts a global learning rate to isolate global adaptation from parameter-specific adjustments. This approach remains robust to hyperparameters and narrows the performance discrepancy between standard methods and adaptive algorithms. Similarly, AGD~\cite{232} utilizes a specific hyperparameter to govern the application of adaptive step sizes, which facilitates a seamless transition between SGD~\cite{307} and adaptive optimization. The advantage of this class is the unified treatment of convergence speed and generalization, whereas the inherent limitation is the reliance on precise hyperparameter tuning to control the transition boundary.\\
\textbf{Gradient smoothing hybrid.} To navigate highly non-convex loss landscapes and to prevent entrapment in suboptimal local minima, recent advancements emphasize smoothing mechanisms that transcend raw and localized gradient estimates. The method of AGS-GD~\cite{236} explicitly smooths the objective landscape by incorporating anisotropic Gaussian smoothing into traditional techniques. This replaces vanilla local gradients with non-local variants whose covariance matrix adaptively aligns with the topological properties of the function. Parallel to this spatial smoothing of the loss surface, the framework of NOVAK~\cite{309} addresses the identical fundamental challenge through temporal trajectory smoothing and variance rectification. Rather than altering the spatial gradient evaluation point, it implicitly smooths the optimization path via a memory-efficient lookahead mechanism. Collectively, these approaches illustrate a paradigm shift from pure point-estimate descent towards spatially or temporally smoothed dynamics. This shift profoundly enhances the ability of the optimizer to traverse complex and ill-conditioned structures of minima. The primary advantage is the robust avoidance of sharp minima, while the inherent limitation is the significant computational overhead introduced by covariance estimations or lookahead trajectories.\\
\textbf{Gradient filtering hybrid.} Beyond landscape smoothing, several methods enhance optimization efficiency and convergence across diverse settings by integrating gradient filtering techniques with hybrid update rules that are tailored to specific objectives. The algorithm of ADASS~\cite{5} adaptively selects training subsets using Lipschitz constants. Furthermore, FESS-GDA~\cite{108} applies smoothing techniques to federated minimax optimization, which achieves improved convergence rates. The method of VR-SGD~\cite{158} introduces a snapshot starting strategy with separate rules for smooth and non-smooth objectives to enhance algorithmic flexibility. These methods effectively boost optimization efficiency across diverse tasks while providing rigorous theoretical guarantees for various problem classes. The advantage of filtering is the reduction of gradient noise and computational cost, whereas the inherent limitation is the sensitivity to the predefined filtering thresholds and the potential loss of informative minority gradients.\\
\textbf{Projection gradient hybrid.} To overcome the limitations of vanilla optimizers in constrained spaces, recent advancements deeply couple projection mechanisms with adaptive strategies or momentum-based strategies. Rather than treating constraints as a naive post-processing step, methods such as Cayley SGD~\cite{8} and NAMO~\cite{328} enforce strict geometric structures directly within the update dynamics. While Cayley SGD~\cite{8} operates on the Stiefel manifold via invertible transforms, NAMO~\cite{328} isolates the orthogonalized momentum and uniquely stabilizes it against stochastic noise using a norm-based adaptive scalar. This philosophy of geometrically-aware adaptation is evident in other projection-based methods. Specifically, AdamP~\cite{229} and HVAdam~\cite{289} utilize directional projections to eliminate radial growth and iteratively approximate stable descent directions. Concurrently, methods including PadamP~\cite{202}, LDAdam~\cite{53}, AdaDiag~\cite{54}, and ADAGB2~\cite{174} fuse projection-aware updates with structured preconditioners, error feedback, or second-order tracking to render constrained optimization computationally viable. These methods transcend isolated projections and demonstrate that tightly fusing manifold constraints with variance adaptation is essential for achieving efficient convergence. The distinct advantage is the strict adherence to geometric constraints, but the inherent limitation is the extensive computational cost associated with manifold retractions or complex projection operations.\\
\textbf{Multi-objective hybrid.} Comprehensive hybrid methods address the inherent trade-offs in single-strategy approaches by integrating complementary techniques to balance multiple optimization goals simultaneously. Numerous methods merge adaptive step sizes with momentum calculations. For instance, FAdam~\cite{219} corrects the trajectory of Adam~\cite{287} via natural gradient connections, while Grams~\cite{250} decouples the direction of the gradient and the magnitude of the momentum. Furthermore, MADGRAD~\cite{171} uses momentum combined with dual averaging, and SET-Adam~\cite{197} adjusts the second momentum of Adam to mimic SGD. Similarly, CAdam~\cite{208} omits misaligned gradient-momentum updates, and ZetA~\cite{211} adds scaling factors to Adam~\cite{287}. Adaptive strategies are frequently paired with acceleration techniques. Accelerated GRAAL~\cite{179} combines the acceleration of Nesterov~\cite{324} with curvature-based step sizes. Moreover, GDA-AM~\cite{260} applies Anderson mixing to minimax problems, and Lookahead~\cite{225} employs fast and slow weights for stability. For efficiency-focused hybrids, BAdam~\cite{21} combines block coordinate descent with Adam~\cite{287} for LLMs, BADM~\cite{12} utilizes data splitting for parallelism, and ADAMBS~\cite{185} combines adaptive methods with bandit sampling for example prioritization. Specialized methods like AdaSAM~\cite{133} integrate flatness-aware techniques~\cite{131} with adaptive rates, and NIRMAL~\cite{248} incorporates five complementary strategies into a single framework. The core advantage is the high versatility across varied tasks, whereas the inherent limitation is the extreme complexity of hyperparameter interactions and the difficulty in isolating the contribution of individual components.

The evolution of hybrid optimization reflects a milestone paradigm shift from isolated heuristic designs to unified mathematical frameworks. The developmental trajectory progresses from elementary metric blending to sophisticated spatial smoothing, subsequently advancing toward geometrically constrained projections, and concluding with systemic multi-objective integrations. Despite these advancements, a critical gap remains in the current literature. The integration of multiple operators frequently leads to over-parameterized optimization systems where theoretical assumptions diverge significantly from practical implementations. Consequently, the primary challenge for future research is to develop autonomous hybrid frameworks that eliminate the reliance on manual hyperparameter tuning while maintaining strict theoretical convergence bounds across highly heterogeneous loss landscapes.
\subsubsection{Towards LLMs Training}
As the scale of models expands rapidly, conventional optimizers frequently encounter hardware memory bottlenecks and computational inefficiency. To address the challenges of training LLMs, recent methods redesign the optimization dynamics through four key dimensions: gradient transformation, dimensionality reduction, parameter management, and computational workflow. The general mathematical framework of these memory-efficient optimizers can be abstracted as $\theta_{t+1} = \theta_t - \eta_t \mathcal{M}_t(\mathcal{T}_t(\nabla_{i_t} J(\theta_t)))$, where $\nabla_{i_t}$ denotes block-wise or stochastic gradient evaluation, $\mathcal{T}_t$ represents a memory-efficient preconditioning or projection operator, and $\mathcal{M}_t$ signifies hardware-aware memory management strategies such as operation fusion.\\
\textbf{Gradient preconditioning mechanism.}
In contrast to traditional methods relying on heavy historical statistics, gradient preconditioning mechanisms focus on instantaneous spectral properties to reduce memory footprints. For example, SWAN~\cite{49} combines row-wise root mean square normalization and gradient whitening to precondition gradients. By leveraging a diagonal replacement heuristic, the algorithm of SWAN~\cite{49} reduces the computational complexity of whitening operations while maintaining the effective learning rate. This approach enhances standard stochastic gradient descent~\cite{307} through adaptive preconditioning, achieving a balance between optimization performance and memory efficiency without incurring the burden of historical momentum states. The advantage of this mechanism is the significant reduction in optimizer state memory, whereas the inherent limitation is the potential loss of long-term directional stability typically provided by moving averages.\\
\textbf{Gradient projection mechanism.}
Progressing from preconditioning to structural dimensionality reduction, projection mechanisms mitigate the computational overhead of complex matrix operations by constraining optimization dynamics within reduced subspaces. For instance, APOLLO~\cite{43} and the variant APOLLO-Mini~\cite{43} utilize stochastic projections to approximate gradient scaling within low-rank or rank-$1$ auxiliary subspaces. Transcending purely computational motivations, SSO~\cite{327} repurposes this projection paradigm to enforce strict geometric stability. Rather than computing the full spectrum, the method of SSO~\cite{327} isolates only the top singular components to project parameter updates onto the tangent space of a spectral manifold. Fundamentally, these methods share a core philosophy of leveraging low-rank spectral approximations to tame costly high-dimensional parameter updates. The primary advantage is the strict bound on computational complexity per step, while the inherent limitation is the approximation error introduced by discarding lower-rank gradient information, which may degrade fine-grained convergence.\\
\textbf{Block-wise computation.}
Beyond modifying the gradient vector, block-wise computation addresses memory constraints through the temporal segmentation of the parameter space. The framework of BAdam~\cite{21} integrates block coordinate descent into the adaptive moment estimation algorithm~\cite{287}, partitioning model parameters into manageable blocks and updating them sequentially rather than simultaneously. This mechanism effectively circumvents the memory bottleneck of full-parameter updates by limiting the active optimization state to a single block. Consequently, it offers a trade-off scheme to scale complex optimizers to LLMs under limited resources, thereby maintaining optimization effectiveness while ensuring video memory efficiency. The notable advantage is the ability to run sophisticated optimizers on severely memory-constrained hardware, but the inherent limitation is the substantial decrease in overall training speed due to sequential processing and potential coordinate misalignment.\\
\textbf{Real-time computation.}
The final evolution in this paradigm tackles the execution workflow directly through real-time computation strategies. Unlike traditional methods that strictly separate gradient accumulation from parameter updates, LOMO~\cite{17} adopts a fused computational strategy to minimize the overall memory footprint. By directly integrating gradient computation with parameter updates and employing mixed-precision training, the architecture of LOMO~\cite{17} eliminates the requirement to store full gradient states during the backward pass. This approach enables the efficient training of massive models on constrained hardware through immediate real-time processing, rigorously balancing computational stability with strict memory efficiency. The distinct advantage is the achievement of a theoretical minimum memory consumption for gradient-based training, whereas the inherent limitation is the inflexibility in applying complex gradient transformations or global gradient clipping before the update step.\\
The evolution of optimization for LLMs reflects a milestone paradigm shift from unconstrained optimization to hardware-aware algorithmic design. This developmental trajectory progresses logically: starting from the algorithmic simplification of preconditioning matrices, advancing to the mathematical reduction of update dimensions via low-rank projections, shifting to the spatial segmentation of parameter updates, and culminating in the hardware-level fusion of computational graphs. Despite these significant advancements, a critical gap remains in the current landscape. Existing memory-efficient methods typically sacrifice global information or temporal dynamics to fit within hardware constraints, leading to a fundamental trade-off between memory footprint and convergence speed. The core challenge for future research is to design optimization frameworks that can mathematically reconstruct global gradient dynamics from localized and low-memory approximations without incurring the overhead of full-state materialization.

In summary, the eight dimensions of FO optimization examined in this section collectively illustrate the remarkable evolution from simple gradient descent to sophisticated, task-aware optimization frameworks. Each dimension addresses a specific deficiency of the base algorithm: momentum facilitate escape from saddle points, adaptivity mitigates the burden of manual tuning, curvature awareness accelerates convergence in pathological regions, stability mechanisms ensure robust training under noise, scheduling optimizes the global progression of learning, generalization techniques seek flat minima, hybrid methods combine complementary strengths, and memory-efficient variants enable scaling to massive models. The common thread running through these diverse approaches is the pursuit of algorithms that can automatically adapt to the geometry of the loss landscape while maintaining computational efficiency and theoretical guarantees. Despite these advances, their reliance on linear approximations renders them inherently vulnerable to highly non-convex, ill-conditioned loss landscapes. Mathematically, FO methods optimize in Euclidean space, making them acutely sensitive to parameterization and prone to pathological zig-zagging in ravines with high condition numbers. Furthermore, the dimensional inconsistency between parameters and gradients forces FO methods to rely heavily on manually tuned, heuristic learning rates. These unresolved challenges naturally lead to the exploration of second-order methods, which explicitly incorporate curvature information to solve these problems, as discussed in the following section.
\begin{figure*}[t!] 
    \centering
    \adjincludegraphics[width=1.0\linewidth]{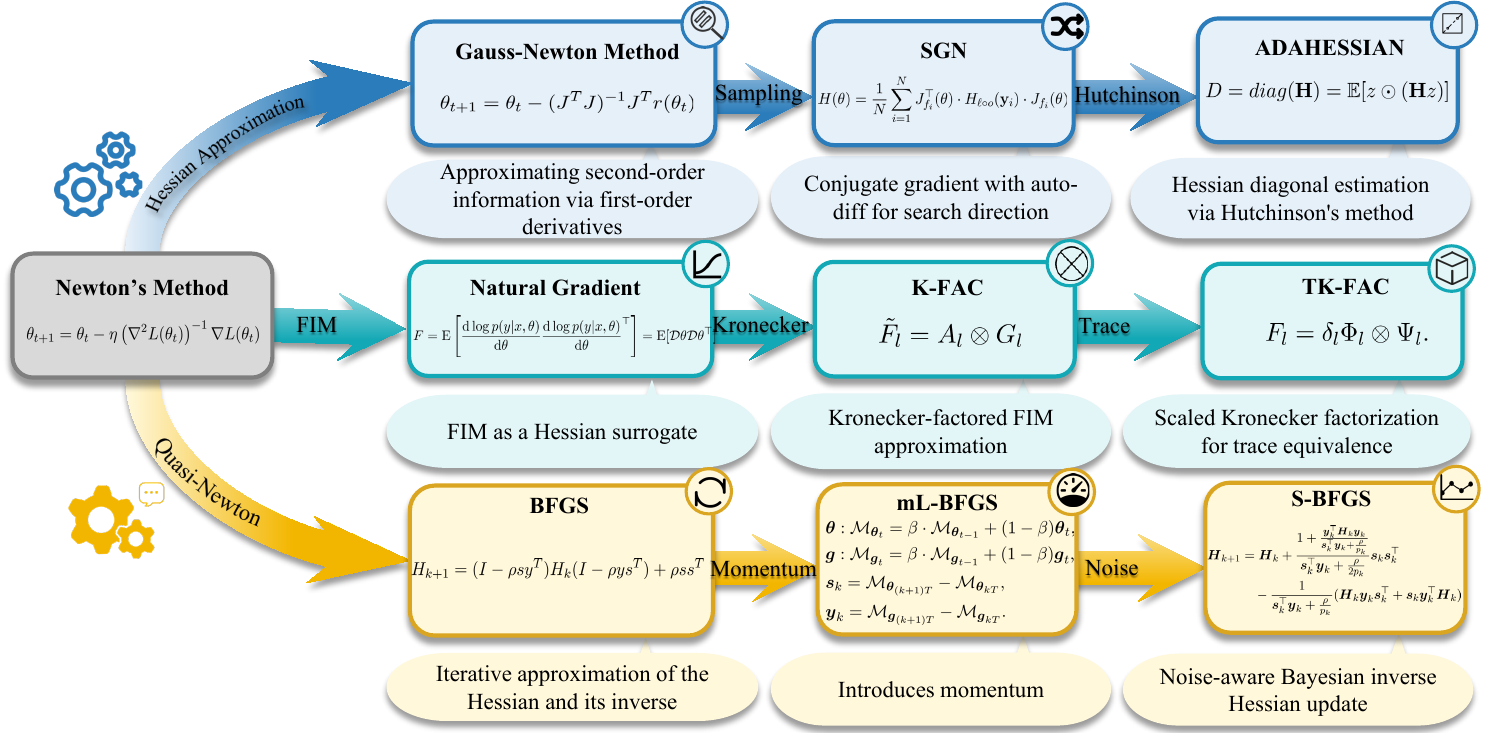}
   
    \caption{\textbf{Evolution and formulations of typical second-order methods.} We show the prominent algorithms from ~\cref{so_methods}, categorizing them by Hessian approximation (~\cite{281,257,258}), FIM application (~\cite{283,255,259}), and quasi-newton (~\cite{282,262,278}), and analyze their key transitional mechanisms through the lens of mathematical formulations.}
    
    \label{fig:evolution_so}
    \vspace{-1.0em} 
\end{figure*}
\subsection{Second-Order Algorithms}\label{so_methods}
These methods transcend FO fundamental limitations by incorporating local curvature information via the Hessian or FIM. Geometrically, applying the inverse curvature matrix acts as a preconditioner that performs an affine transformation, warping ill-conditioned ravines into isotropic basins where the negative gradient points directly toward the optimum. By transitioning the optimization trajectory from coordinate-dependent Euclidean distances to distribution-aware Riemannian manifolds (e.g., Natural Gradient Descent~\cite{283}), SO methods inherently provide scale-invariant updates and optimal intrinsic step sizes. This section systematically reviews the evolution of SO algorithms, focusing on how they balance the computational cost of curvature estimation with practical scalability. The discussion is organized into four core dimensions: Hessian Approximation and Estimation (~\cref{sec:second_1}), Fisher Information Matrix Applications (~\cref{sec:second_2}), Quasi-Newton Methods (~\cref{sec:second_3}), and Second-Order Moment Fusion (~\cref{sec:second_4}). These paradigms represent a progressive transition from exact but prohibitive full-matrix computations to efficient structural and stochastic approximations, ultimately aiming to achieve both theoretical robustness and empirical efficiency in large-scale deep learning. Furthermore, \cref{fig:evolution_so} maps the evolutionary trajectory of these algorithms, and \cref{tab:optim_method} details the merits and demerits of representative methods.
\subsubsection{Hessian Approximation \& Estimation} \label{sec:second_1}
First-order methods generally fail to exploit the geometric structure of the optimization landscape. While the classical Newton method~\cite{280} utilizes exact curvature to define the update direction, the prohibitive computational cost of full Hessian computation often limits practical application in large scale scenarios. The Gauss-Newton method~\cite{281} pioneered the indirect construction of second-order curvature information by utilizing the first-order derivatives to approximate the Hessian, which significantly reduces computational complexity. Building upon this foundation, subsequent research has established a unified formal derivation where the true Hessian is replaced by an approximated surrogate $\tilde{H}$. These methods employ intelligent structural approximation, stochastic sampling, and gradient estimation to extract rich curvature information, thereby accelerating convergence and enhancing optimization performance without the explicit computation of the full Hessian matrix. Theoretical convergence analysis of these methods typically guarantees sublinear or linear convergence rates under standard convexity or smoothness assumptions, provided that the approximated curvature matrix satisfies specific spectral bounds. However, a common limitation lies in the persistent boundary between approximation accuracy and computational overhead, where aggressive simplification often leads to degraded performance in highly non-convex landscapes.\\
\textbf{Block Hessian approximation.} To balance optimization performance and computational efficiency, certain approaches leverage structured block-wise curvature information, assuming independence among distinct parameter groups. The unified form typically partitions the Hessian into a block-diagonal matrix $ \tilde{H}_{block} $, where off-diagonal interactions are discarded. Athena~\cite{271} uses second-order matrix derivative information to guide block-wise quantization, grouping parameters by columns and rows to optimize the quantization process iteratively. Q-Newton~\cite{276} adopts hybrid scheduling, dynamically allocating the task of matrix inversion between quantum processors and classical processors based on real-time matrix properties, and utilizing a cost-aware scheduler to offload tasks when advantageous. The primary advantage of block approximation is the retention of rich local curvature information within parameter groups. Conversely, the inherent limitation is the complete loss of cross-block parameter dependencies, which may hinder convergence in highly coupled network architectures.\\
\textbf{Diagonal Hessian approximation.} Taking the structural simplification to its extreme, diagonal approximation addresses the critical trade-off between the curvature-aware convergence benefits of second-order optimization and the prohibitive cost of full matrix operations. The underlying mathematical derivation restricts the surrogate matrix to $ \tilde{H}_{diag} = \text{diag}(H) $, enabling efficient curvature-guided element-wise updates. AdaHessian~\cite{258} utilizes the method of Hutchinson for diagonal approximation. HesScale~\cite{270} improves the approximation quality with negligible overhead to scale steps in reinforcement learning. Sophia~\cite{264} applies lazy-updated diagonal estimates coupled with gradient clipping, matching the cost of Adam but achieving faster convergence. Fed-Sophia~\cite{265} implements the approach of Sophia~\cite{264} in federated learning through weighted averaging and clipping mechanisms. OptiQ~\cite{266} formulates optimization trajectories through ordinary differential equations, utilizing quiescence to enable adaptive large steps and applying partial Hessian inversion to enhance overall efficiency. Furthermore, CRNAS~\cite{177} combines cubic regularization and affine scaling for constrained problems, while SASSHA~\cite{272} integrates sharpness minimization with stable lazy updates to boost generalization. The advantage of these methods lies in their extreme computational efficiency and minimal memory footprint. However, the fundamental limitation is the complete ignorance of off-diagonal curvature, making it difficult to navigate highly ill-conditioned ravines where variable interactions are dominant.\\
\textbf{Stochastic Hessian sampling.} Transitioning from deterministic structural simplification to probabilistic estimation, stochastic sampling provides an essential mechanism for large-scale optimization tasks. This paradigm enables efficient curvature estimation by constructing a low-rank surrogate $\tilde{H}_{sample}$ through randomized data subset evaluation. SketchySGD \cite{263} uses randomized approximations of Nyström to estimate the Hessian from minibatches, featuring automated learning rate selection and infrequent preconditioner updates. SGN~\cite{257} employs the approximation of Gauss-Newton and conjugate gradient methods with automatic differentiation to determine search directions. Additionally, STDE~\cite{4} generalizes the concept of stochastic estimation to arbitrary derivative orders using automatic differentiation in Taylor mode. The core advantage of stochastic sampling is the ability to adapt to varying data scales while capturing global curvature trends. The primary limitation is the introduction of significant variance into the curvature estimation, which necessitates complex variance reduction techniques to ensure theoretical convergence.\\
\textbf{Gradient difference estimation.} Moving beyond explicit matrix construction, implicit methods estimate curvature through vector operations. SGDHess~\cite{261} represents an optimization algorithm based on stochastic gradient descent. The core formulation performs high-order estimation of gradient differences via Hessian-vector products $ \nabla^2 L v $, thereby correcting systematic biases in momentum without materializing the matrix. This method leverages Hessian-guided momentum corrections to boost convergence efficiency. The advantage is the circumvention of direct matrix storage, scaling linearly with the size of the model. The inherent limitation is the reliance on accurate finite difference approximations, which are highly sensitive to numerical instability and noise in stochastic gradients.

The evolutionary trajectory of SO optimization reflects a systematic paradigm shift from exact dense matrix computation to highly structured, sparse, or implicit approximations. The historical progression moved from full Hessian matrices to spatial simplifications involving block and diagonal forms, and subsequently evolved toward probabilistic and implicit estimations through stochastic sampling and gradient differences. A persistent gap in existing research is the lack of adaptive algorithms capable of dynamically transitioning between these paradigms during the training process. Current methods typically enforce a static approximation structure throughout optimization, which fails to accommodate the varying curvature characteristics across different training phases of deep neural networks.
\subsubsection{Fisher Information Matrix Applications} \label{sec:second_2}
The practical deployment of second-order optimization techniques, such as Natural Gradient Descent~\cite{283}, relies fundamentally on the efficient approximation of the Fisher Information Matrix. Within the general mathematical framework, the parameter update is governed by the inverse of the Fisher Information Matrix multiplied by the gradient of the objective function. However, the exact computation and inversion of this matrix incur prohibitive computational costs for large-scale models. Consequently, the core improvement dimensions in recent research focus on balancing computational tractability with the retention of critical geometric curvature through diagonalization, block Kronecker factorization, numerical constraints, and structural customization.\\
\textbf{Diagonal Fisher approximation.} The most fundamental paradigm shift toward scalability involves reducing the dense matrix to a diagonal structure, which represents the unified form of treating parameters as independent entities. This category achieves linear time complexity by ignoring cross-parameter correlations. Methods such as SOAA~\cite{267} reduce algorithmic complexity through direct diagonal approximations. Furthermore, OCAR~\cite{275} adapts this framework to continual learning environments by imposing KL divergence constraints, while RACS~\cite{220} introduces a structured approach to diagonal approximation. The primary advantage of this paradigm is extreme computational efficiency, whereas the inherent limitation is the severe loss of structural curvature information.\\
\textbf{Block-diagonal Kronecker approximation.} To address the limitations of pure diagonal methods, the evolutionary logic progresses to modeling intra-layer correlations. The unified derivation of this category approximates the Fisher Information Matrix as the Kronecker product of two smaller matrices, typically representing activation covariances and pre-activation gradient covariances. K-FAC~\cite{255} establishes the foundation for this approach, enabling efficient matrix inversion and accelerating convergence. Extending this paradigm, AdaFisher~\cite{273} serves as an adaptive optimizer that bridges diagonal and block-Kronecker structures, balancing the convergence benefits of second-order methods with the efficiency of first-order updates. The advantage of block-diagonal methods lies in the accurate capture of layer-wise curvature, though they inherently fail to model inter-layer dependencies.\\
\textbf{Trace-preserving Fisher approximation.} Advancing beyond standard Kronecker factorization, subsequent methods focus on numerical calibration. T-KFAC~\cite{259} decomposes blocks into Kronecker products constrained by trace-restricted coefficients. This derivation ensures that the sum of the diagonal elements matches the true FIM, thereby addressing the scaling inaccuracies present in the original K-FAC~\cite{255}. The advantage of trace preservation is enhanced approximation accuracy and spectral stability across diverse architectures, while the limitation is the additional computational overhead required for trace estimation.\\
\textbf{Curvature-aware approximation.} The most recent paradigm tailors the approximation to specific architectural properties. For instance, MAC \cite{277} utilizes mean activations to reduce the computational cost of estimating curvature. This method specifically applies Kronecker factorization to the attention layers of modern architectures and integrates attention scores into the preconditioning process. The advantage is highly optimized performance for specific models, whereas the inherent limitation is the lack of universal applicability to non-attention-based networks.

These approximation techniques ensure that the substitute matrix remains positive definite, which is a critical requirement for bounding the condition number and guaranteeing the stable descent of the objective function. Despite these theoretical guarantees, the comparison of method advantages and inherent limitations reveals a significant boundary in current methodologies. The core gap of existing work lies in the static nature of these approximations; there is an absence of a universally adaptive framework that can dynamically transition between diagonal, block, and trace-preserving structures based on the real-time curvature requirements of the loss landscape without demanding manual architectural derivations.
\subsubsection{Quasi-Newton Methods} \label{sec:second_3}
The core mathematical framework of quasi-Newton methods, quintessentially represented by the BFGS \cite{282} algorithm, is to dynamically approximate the curvature of loss functions through iterative updates. This methodology avoids the direct computation of expensive second-order derivatives. Formally, the approximation of the Hessian matrix, denoted as $B_t$, satisfies the secant condition $B_{t+1} s_t = y_t$, where $s_t$ represents the parameter difference and $y_t$ represents the gradient difference. To adapt this framework for deep learning, subsequent approaches systematically enhance the optimization process across three core improvement dimensions: resource constraints, algorithmic stability against noise, and system scalability. The progression of these methods logically evolves from reducing memory footprints to mitigating stochastic variance, and ultimately to distributing computations across massive architectures.\\
\textbf{Low-memory quasi-Newton.} The primary bottleneck of the standard framework is the quadratic memory requirement for storing the dense approximation matrix. Low-memory variants resolve this limitation by maintaining a sparse or implicit representation. Specifically, the approaches for deep neural network training introduce factorization strategies to maintain bounded eigenvalues in the approximations, as demonstrated by the K-BFGS~\cite{268} methods. These algorithms compute the search direction utilizing only a restricted recent history of gradients and parameters. This structural approximation significantly improves the scalability of the optimization process while preserving convergence stability, serving as the foundational step for subsequent deep learning adaptations.\\
\textbf{Stochastic quasi-Newton.} When applied to stochastic objectives, the traditional secant condition becomes unstable due to gradient variance. Therefore, stochastic adaptations integrate probabilistic modeling and variance reduction techniques to formulate a unified derivation. For instance, the S-BFGS~\cite{278} method utilizes Bayesian inference to assimilate noisy gradients while controlling the curvature updates. Furthermore, algorithms such as SpiderSQN~\cite{256} integrate specific variance reduction frameworks to achieve optimal complexity in non-convex stochastic settings. Another approach, FUSE-PV~\cite{241}, adapts the L-BFGS~\cite{288} formulation with mini-batch computations. These formulations mathematically unify second-order precision with stochastic efficiency to handle highly non-convex tasks.\\
\textbf{Distributed quasi-Newton.} To address the prohibitive computational loads of massive datasets, distributed paradigms partition the curvature estimation. The mL-BFGS~\cite{262} method employs block-wise approximations of the Hessian matrix to distribute memory demands across computation nodes. To accelerate convergence without relying on expensive variance reduction techniques, this method incorporates a momentum scheme into the L-BFGS~\cite{288} iterations to mitigate stochastic noise. The combination of block-wise distribution and momentum smoothing provides a robust unified formal derivation for scaling quasi-Newton optimizers to large distributed systems.\\
From a theoretical perspective, the convergence analysis of these methods establishes optimal sublinear rates for non-convex optimization, provided that the eigenvalue bounds of the approximated Hessian matrix are strictly maintained. However, the boundaries of these theoretical guarantees often assume bounded gradient variance, which does not universally hold in practical deep learning landscapes. The inherent limitation of the entire framework is the staleness of curvature information when gradients fluctuate rapidly across successive mini-batches.

The milestone paradigm shift in this domain moves from deterministic exact matrix constructions to stochastic structurally constrained approximations. For the low-memory paradigm, the primary advantage is the linear reduction in storage complexity, while the inherent limitation is the loss of historical curvature information. For the stochastic paradigm, the advantage lies in the mathematical robustness against batch noise, whereas the limitation is the reliance on complex hyperparameter tuning for variance reduction. For the distributed paradigm, the advantage is the parallel processing capability, but the inherent limitation is the communication overhead required to synchronize the block-wise updates. Consequently, the core gap in existing work is the lack of a generalized quasi-Newton optimizer that simultaneously achieves exact curvature matching, requires zero communication overhead, and maintains resilience against arbitrary stochastic noise in highly non-convex deep neural network architectures.
\subsubsection{Second-Order Moment Fusion} \label{sec:second_4}
To establish a general mathematical framework, second-order moment fusion methods integrate local curvature information into the momentum sequence, typically defining the update rule as $\theta_{t+1} = \theta_t - \eta P_t^{-1} m_t$, where $P_t$ is a preconditioner derived from the Hessian matrix. This formulation aligns with the unified perspective in~\cref{eq:unified_master} by interpreting $P_t^{-1}$ as the preconditioner $M_t^{-1}$ and $m_t$ as the output of the momentum update $\phi(\cdot)$. The core improvement dimension of these methods lies in resolving the critical trade-off between convergence speed and stability. Through a unified formal derivation, representative algorithms scale the accumulated historical gradients by an approximated curvature matrix, effectively adjusting the step size for each parameter based on the geometric properties of the loss landscape. Theoretical convergence analysis indicates that this fusion accelerates the traversal of flat regions and dampens oscillations in steep directions, thereby improving the overall convergence rate. The boundaries of this framework are primarily constrained by the accuracy of the curvature estimation and the susceptibility of second-order information to stochastic noise.\\
\textbf{Momentum-curvature fusion.} The evolution of this paradigm begins with the direct integration of curvature information into the gradient accumulation process. SGDHess~\cite{261} exemplifies this approach by combining the historical gradient accumulation of momentum with curvature estimations to mitigate optimization bias. This method refines the momentum vector using curvature data, which facilitates rapid convergence without the necessity for large batch sizes. This fusion paradigm enhances convergence efficiency and adaptability in specific optimization scenarios by leveraging the geometric awareness of second-order methods to guide the momentum trajectory.\\
\textbf{Noise-robust second-order momentum.} Building upon basic curvature integration, the evolutionary logic progresses to address the inherent noise sensitivity of full second-order methods in stochastic environments. Accurate Hessian estimation is frequently compromised by noisy gradients and the non-convex nature of deep learning objectives. To overcome these challenges, methods employ efficient diagonal approximations coupled with noise-reduction techniques. For instance, AdaHessian~\cite{258} utilizes spatial averaging to smooth spatial variations and incorporates Hessian momentum to suppress estimation noise. Similarly, Sophia~\cite{264} employs per-coordinate clipping to protect against inaccurate curvature estimates. Both algorithms prioritize efficient diagonal approximations and noise-mitigation strategies, successfully avoiding the computational burden of the full Hessian matrix while preserving the fundamental benefits of second-order optimization.

The trajectory of second-order moment fusion demonstrates a milestone shift from naive curvature integration to noise-resilient, computationally viable approximations. A comparative analysis highlights the specific trade-offs of each category. The primary advantage of momentum-curvature fusion is the significant acceleration of convergence in low-noise settings, whereas the inherent limitation is the vulnerability to inaccurate Hessian estimations caused by stochastic mini-batches. Conversely, noise-robust second-order momentum offers the advantage of stable optimization in highly stochastic environments, yet it introduces the inherent limitation of structural information loss due to the reliance on diagonal approximations. Ultimately, the core gap in existing research is the absence of an adaptive mechanism that can dynamically calibrate the degree of curvature integration based on real-time noise levels, ensuring optimal fusion of first-order momentum and second-order geometry across diverse training phases.

In summary, the four dimensions presented in this section collectively illustrate the progression of second-order optimization from dense and expensive computations to efficient, structure-aware approximations. Each paradigm offers a unique trade-off between curvature accuracy and computational feasibility, yet all share the common goal of accelerating convergence in ill-conditioned landscapes. While SO represents the mathematical pinnacle of gradient-based learning by exploiting rich geometric curvature, it is fundamentally handcuffed by strict physical and structural prerequisites: it demands complete white-box access, rigorous twice-differentiability, and prohibitive computational memory to materialize the curvature matrices. As modern deep learning scales towards billion-parameter LLMs and ventures into non-differentiable black-box environments, the stringent requirements of SO, and even FO methods, hit an insurmountable memory and applicability wall. The emergency of theses challenges naturally lead to the exploration of zeroth-order algorithms, as discussed in the following section.
\begin{figure*}[t!] 
    \centering
    \adjincludegraphics[width=1.0\linewidth]{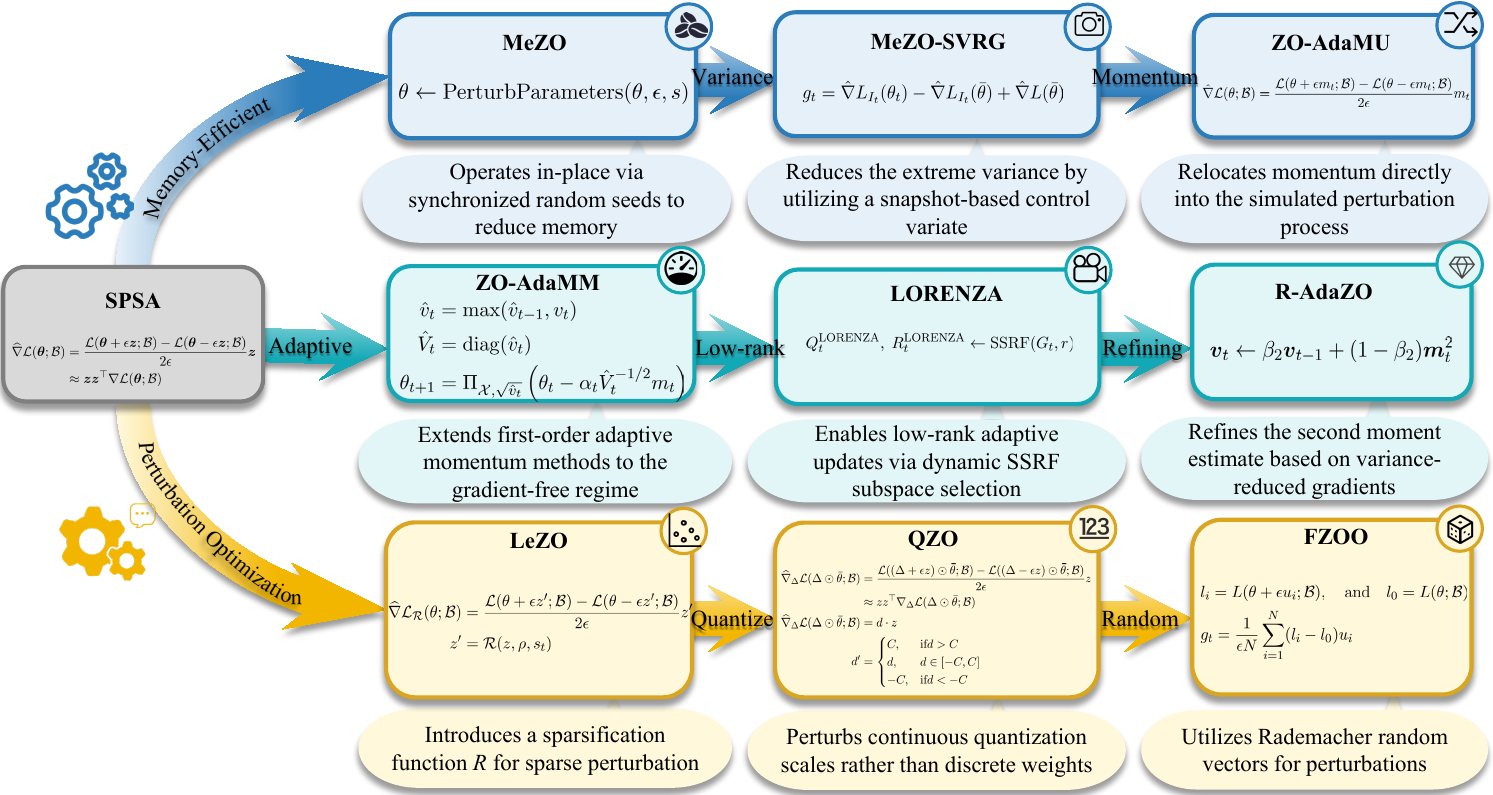}
   
    \caption{\textbf{Evolution and formulations of typical zeroth-order methods.} We show the prominent algorithms from ~\cref{zo_methods}, categorizing these algorithms by memory efficiency (~\cite{16,19,20}), adaptive strategies (~\cite{285,27,147}), and perturbation optimization (~\cite{50,36,37}), and analyze their key transitional mechanisms through the lens of mathematical formulations.}
    
    \label{fig:evolution_zo}
    \vspace{-1.0em} 
\end{figure*}
\subsection{Zeroth-Order Algorithms}\label{zo_methods}
These algorithms emerge not as a mathematical upgrade to SO, but as a radical paradigm shift that trades exact geometric precision for ultimate physical feasibility. By completely bypassing backpropagation, ZO estimates gradient directions solely through forward-pass evaluations of randomly perturbed parameters. This forward-only formulation obliterates the massive auxiliary memory overhead of activations, effectively decoupling model training from the catastrophic memory wall and enabling exact matching with inference memory footprints (e.g., MeZO~\cite{16}).This section systematically reviews advancements in ZO optimization, organizing them into six core dimensions. These dimensions represent a progressive effort to bridge the performance gap between ZO and gradient-based algorithms while preserving the unique advantages of gradient-free optimization. We further outline the evolutionary trajectory of these methods, which is illustrated in~\cref{fig:evolution_zo}, and we evaluate the merits and demerits of representative algorithms, which are detailed in~\cref{tab:optim_method}.
\subsubsection{Adaptive Methods}
To bridge the performance gap between zeroth-order optimization and first-order optimization, adaptive methods extend classical mechanisms from the first-order regime. These approaches formulate a unified paradigm that manipulates temporal sequences, parameter spaces, and geometric structures. The evolutionary logic of these methods progresses from mitigating temporal noise to restricting the search space, and ultimately to compressing the update representations.\\
\textbf{Momentum-based adaption.} The foundational step in this evolution involves the integration of momentum accumulation and adaptive learning rates to mitigate the inherent noise of random perturbations. The unified formal derivation of this category updates the first and second moments of the gradient estimates using historical exponential moving averages, thereby stabilizing the optimization trajectory. ZO-AdaMM~\cite{285} establishes this foundation by translating adaptive momentum into the zeroth-order regime. R-AdaZO~\cite{147} advances this paradigm by exploiting the variance reduction properties of first-moment estimates, which refines the second-moment sequence with noise-reduced signals to produce more reliable updates. The primary advantage of these momentum-based techniques is the significant reduction in estimation variance without requiring additional function evaluations. However, the inherent limitation is that the theoretical convergence rate remains strongly dependent on the ambient dimensionality of the problem, which restricts efficiency in ultra-large models.\\
\textbf{Projection-based adaption.} To address the dimensionality bottleneck of momentum methods, the evolutionary trajectory shifts toward optimizing the sampling space. Methods in this category boost the efficiency of zeroth-order optimization by confining the perturbation vectors to lower-dimensional subspaces or structured projections. The core approximates the gradient by accumulating the scaled function differences along a restricted set of anchor directions or random low-dimensional bases. DiZO~\cite{42} exemplifies this approach by adapting layer updates through anchor-based projections, representing the core mechanism of projection-based adaptation. ZO-SAH~\cite{279} extends this geometric approach by estimating the Hessian matrix through quadratic fitting within random two-dimensional subspaces. It utilizes periodic switching to reuse function evaluations while ensuring the positive definiteness of the approximation. The significant advantage of projection-based methods is the drastic reduction in the computational cost per iteration. The corresponding limitation is the potential loss of critical gradient information that lies outside the selected projection subspaces, which can lead to suboptimal local minima.\\
\textbf{Low-rank adaption.} The final stage of this evolutionary path synthesizes subspace projection with adaptive optimization to achieve both memory efficiency and enhanced generalization. LORENZA~\cite{27} embodies this synthesis by employing AdaZo-SAM, a framework that integrates the update rules of Adam~\cite{287} and SAM~\cite{131} using a single gradient approximation per iteration. This framework is combined with randomized singular value decomposition to dynamically generate adaptive subspace projections. By linking computational efficiency, sharpness awareness, and low-rank adaptivity, this method successfully balances operational cost and generalization performance. The advantage of low-rank adaptation is the ability to maintain a minimal memory footprint while actively seeking flat minima to improve model robustness. The inherent limitation is the computational overhead introduced by the frequent singular value decompositions required to maintain the adaptive subspaces.

In summary, the core design paradigm of adaptive zeroth-order algorithms demonstrates a clear transition from naive high-dimensional random sampling to history-aware and structurally constrained approximations. Despite these significant milestones, a critical gap remains in the existing literature. The current methodologies still struggle to achieve strict first-order comparable convergence rates in unstructured, massively parameterized spaces without incurring prohibitive sampling costs.
\subsubsection{Perturbation Optimization}
ZO optimization relies fundamentally on finite difference approximations to estimate gradients, a process that typically utilizes forward function evaluations based on perturbed model parameters. The general framework evaluates the objective function using random perturbation vectors, where the gradient approximation is derived from the difference between these evaluations. However, traditional methods employing isotropic Gaussian perturbations frequently suffer from high estimation variance, slow convergence, and significant computational overhead. To address these challenges, recent advancements focus on optimizing the perturbation mechanism itself. The core improvement dimensions encompass a progression from simplifying the underlying noise distribution to restricting the perturbation space, and ultimately to refining the temporal dynamics of the sampling process.\\
\textbf{Random sign perturbation.} The initial evolutionary step focuses on simplifying the fundamental sampling distribution to alleviate computational bottlenecks. Instead of utilizing dense continuous noise, FZOO~\cite{37} utilizes Rademacher random vectors for perturbations, replacing the Gaussian distributions utilized in conventional methods such as MeZO~\cite{16}. The unified formulation of this approach restricts the perturbation vector to discrete sign values to approximate gradients efficiently. The primary advantage of this method is the drastic reduction in the computational overhead required to generate perturbation vectors, which maintains convergence properties while minimizing the complexity of forward passes in large-scale learning scenarios. Conversely, the inherent limitation is that discrete sign perturbations may struggle to capture fine-grained directional information in highly ill-conditioned loss landscapes, potentially bounding the achievable precision of the model.\\
\textbf{Sparse perturbation.} Advancing from distribution simplification, the evolutionary trajectory moves toward restricting the spatial dimensions of the perturbation. Methods in this category address high computational and memory overheads by selectively perturbing sparse parameter subsets, which enables efficient fine-tuning while preserving task performance. The mathematical formulation applies two-point gradient estimation exclusively to a dynamically selected sparse mask. Sparse-MeZO~\cite{23} applies this strategy through per-layer magnitude thresholds to select parameter-level subsets. Similarly, LeZO~\cite{50} integrates the SPSA~\cite{284} framework with layer-wise sparsity to selectively perturb entire layers. Both methods successfully eliminate auxiliary memory costs. The distinct advantage of sparse perturbation is the ability to facilitate the adaptation of massive models under strict memory constraints. However, the inherent limitation is that hard thresholding or layer-wise selection may discard critical cross-layer gradient interactions, which can introduce bias into the optimization trajectory.\\
\textbf{Quantized perturbation.} Parallel to sparsity, this category restricts the perturbation space to accommodate low-precision architectures. QZO~\cite{36} addresses the requirement for the stable fine-tuning of quantized neural networks by perturbing only the continuous quantization scales rather than the discrete weights. The core advantage of this method is the provision of a memory-efficient and structurally stable mechanism tailored specifically for quantized environments. The inherent limitation is that restricting updates strictly to scaling factors severely limits the expressiveness of the gradient approximation, which may prove insufficient for complex tasks that require substantial weight transformations.\\
\textbf{Paired perturbation sampling.} The final evolutionary stage refines the temporal dynamics of the perturbation process to actively reduce estimation variance. ZO-AdaMU~\cite{20} replaces standard random vectors with an adaptive perturbation mechanism that blends historical momentum and current randomness. By constructing paired perturbations within the SPSA~\cite{284} framework, this method yields smoother and faster-converging gradient estimates. The significant advantage of this approach is the robust mitigation of estimation errors and the acceleration of convergence through history-aware sampling. The corresponding limitation is the necessity to store historical momentum states, which partially diminishes the memory-efficiency benefits that are central to standard zeroth-order techniques.

In all, perturbation optimization exhibits a clear evolutionary logic, transitioning from naive isotropic sampling to structurally constrained and history-aware perturbation generation. Despite these paradigm shifts, a critical gap remains in the existing literature. Current methods predominantly rely on heuristic spatial selections or static distribution replacements, and they lack the ability to dynamically adapt the perturbation geometry to the local landscape without incurring prohibitive computational costs associated with higher-order estimations.

\subsubsection{Zeroth-First Order Hybrid}
This framework aims to construct an estimator that minimizes both the memory footprint and the variance of the gradient. The core dimension of improvement lies in determining the optimal integration mechanism between the ZO gradient estimate and the exact ZO gradient. The evolution of these methods progresses from global blending to spatial decomposition, and finally to statistical correction, which enhances the comparative links among the different strategies.\\
\textbf{Weighted hybrid.} As an initial paradigm, Addax~\cite{52} implements a dynamic fusion strategy at each optimization step. It linearly combines the zeroth-order and first-order gradient estimates by modulating the contribution ratios of the estimates via a mixing weight parameter $\alpha$. The unified formal derivation can be expressed as a convex combination of the two directions. This approach enables continuous and fine-grained trade-offs, which allows the optimizer to strike an optimal balance between memory efficiency and gradient accuracy according to real-time requirements. While the advantage of this method is the simplicity of implementation, the inherent limitation is that it still requires the computation of first-order information at each step, which restricts the absolute reduction of memory.\\
\textbf{Layer-wise hybrid.} To achieve strict memory isolation and improve upon the continuous computation requirement of the weighted approach, ElasticZO~\cite{146} employs a spatial decomposition strategy instead of global gradient fusion. By confining the memory-efficient zeroth-order optimization to the heavy shallow layers and reserving the precise backpropagation for the final layers, this approach significantly lowers the memory overhead. It preserves the convergence accuracy via targeted first-order updates in the sensitive tail layers. The primary advantage is the substantial reduction in activation memory, but the intrinsic limitation is the vulnerability to the high variance of the zeroth-order estimates in the shallow layers, which can propagate through the network.\\
\textbf{Variance reduction hybrid.} To address the statistical flaws of spatial and global blending, VAMO~\cite{162} transcends simple linear blending by specifically utilizing first-order information for statistical correction. Addressing the inherent high variance of the zeroth-order estimates, this method applies a correction via a variance reduction term weighted by a configurable coefficient. The formal derivation relies on using precise first-order estimates at reference points to stabilize the noisy zeroth-order directions. By integrating these estimates, it effectively balances the efficacy of variance reduction against the computational overhead required to obtain the correction terms. This paradigm accelerates the theoretical convergence significantly, yet it suffers from the limitation of increased computational costs for maintaining and evaluating the reference points.

The trajectory demonstrates a clear shift from heuristic parameter mixing to rigorous variance reduction. However, a critical limitation remains across the existing literature. The core gap in the current research is the lack of an adaptive framework that can dynamically schedule the integration of first-order and zeroth-order information based on the local topography of the loss landscape, without relying on static structural heuristics or computationally expensive periodic reference points.

\subsubsection{Memory-Efficient Methods}
Some methods perform memory optimization from different dimensions, including storage management, structure exploitation, sparsification, quantization, and scheduling. By formulating the memory consumption of optimization as a constraint bounded by the memory required for forward inference, these methods significantly reduce auxiliary memory usage. They maintain the performance of the model through elaborately designed optimization strategies and avoid the degradation of performance caused by memory constraints. The core dimensions of improvement span from the algorithmic modification of estimators to structural subspace projections and the compression of data representation.\\
\textbf{Inference-level memory ZO.} Some methods address the memory bottleneck of fine-tuning LLMs by designing ZO optimizers that strictly match the memory footprint of inference. MeZO~\cite{16} is the foundational framework, which adapts ZO-SGD to operate in place under the $\mathcal{O}(1)$ auxiliary memory constraints of inference. Building upon this, ZO-AdaMU~\cite{20} relocates the computation of momentum to SPSA perturbations, and it uses an uncertainty schedule along with the second-order approximation of momentum to accelerate convergence. Furthermore, KerZOO~\cite{35} introduces a kernel-based ZO framework that improves the speed of convergence compared to baselines. ZO2~\cite{32} leverages CPU offloading to transfer inactive parameters to the system memory, which retains only a small number of modules required for current computations in the GPU. While these methods share the commonality of targeting the memory of inference, they enhance the ZO estimation $\hat{g}$ with distinct algorithmic strategies. The primary advantage of this category is the exact matching of optimization memory to inference memory, whereas the inherent limitation remains the slow rate of convergence in high-dimensional spaces due to the high variance of gradient estimation.\\
\textbf{Low-rank ZO finetuning.} By leveraging the structures of low-rank gradients, some methods address the prohibitive memory cost of vanilla ZO methods for the fine-tuning of large models. The unified mathematical formulation involves projecting the perturbations into a low-rank subspace $W = AB$, where the intrinsic rank is significantly smaller than the original dimensions, which reduces the effective dimensions of optimization. LOZO~\cite{24} maintains low-rank subspaces through the technique of lazy sampling. TeZO~\cite{38} captures the cross-dimensional low-rank structure using CPD, and it dynamically adjusts the ranks of layers for the optimal use of memory. SuZero~\cite{51} deeply couples layer-wise low-rank perturbations with ZO optimization, which leverages low-rank structures to reduce the variance of gradients and enable the efficient fine-tuning of LLMs. These methods provide memory-efficient and cost-effective solutions by balancing strong performance with reduced computational overhead. The main advantage of these methods is the theoretical reduction of gradient variance and parameter footprint, while the inherent limitation is the potential loss of expressive capacity when modeling highly complex target distributions.\\
\textbf{Sparse parameter.} Some methods address the high computational cost of traditional methods by restricting perturbations and updates to sparse subsets of parameters. Sparse-MeZO~\cite{23} applies two-point ZO estimation in the style of SPSA~\cite{284} on dynamically selected subsets through per-layer magnitude thresholds. LeZO~\cite{50} integrates SPSA~\cite{284} with layer-wise sparsity without introducing additional overhead. MaZO~\cite{31} leverages a metric of weight importance for multi-task learning to identify critical parameters, which effectively reduces the dimensionality. All these methods utilize dynamic sparse subsets to cut the costs of ZO optimization while preserving accuracy. The advantage of this paradigm is the highly targeted optimization with minimal active memory, but the limitation lies in the high sensitivity to the criteria of parameter selection, which can lead to suboptimal convergence if critical weights are ignored.\\
\textbf{Quantized ZO finetuning.} Some methods address the high memory overhead of full-precision gradient-based fine-tuning for quantized models, which enables efficient optimization in environments with constrained resources via ZO methods. The core mechanism involves operating directly on reduced-precision representations $Q(W)$ of weights and perturbations. QuZO~\cite{29} utilizes stochastic quantized perturbation for the fine-tuning of LLMs. QZO~\cite{36} perturbs the scales of quantization and applies directional derivative clipping to ensure training stability. ZOQO~\cite{145} adapts ZO-Sign SGD to quantized noise, and it uses sign updates along with the adjustment of learning rates to maintain the state of quantization. These methods share the avoidance of gradients based on ZO estimation, but they differ in the targets of perturbation and the strategies for stabilization. Collectively, they enable the memory-saving fine-tuning of quantized models. The outstanding advantage is the extreme compression of memory requirements, whereas the primary limitation is that quantization noise severely exacerbates the inherent variance of ZO estimators, which complicates the theoretical guarantees of convergence.\\
\textbf{Layer-wise ZO scheduling.} By decomposing the global optimization process into sequential local updates, some methods optimize the execution pipeline to reuse memory efficiently. DiZO~\cite{42} adapts the updates of layers using anchor-based projections, which embodies the core idea of layer-wise scheduling. This method enhances the efficiency and performance of ZO optimization through adaptive layer scheduling. The advantage of this approach is the substantial reduction of peak memory usage during computation, while the limitation is the increased time complexity and the delayed propagation of updates across the entire network.

The evolution of these methods demonstrate a clear shift from purely algorithmic memory-matching to deep structural exploitation, and finally to hardware-aware data representation and scheduling. The foundational paradigm focuses on algorithmically bypassing the memory overhead of backpropagation to reach the theoretical lower bound of inference memory. Subsequently, the field shifted toward structural paradigms, including low-rank projections and dynamic sparsity, to mathematically reduce the intrinsic dimensionality and variance of the ZO estimators. The most recent evolutionary stage integrates the compression of data precision and temporal-spatial scheduling to further push the boundaries of memory efficiency. Despite these advancements, a critical gap remains in the existing literature. The current works lack a unified theoretical framework that simultaneously integrates mixed-precision quantization, dynamic sparsity, and low-rank adaptations under strict mathematical guarantees of convergence. Future research must address this gap to unlock the full potential of ZO optimization for massive neural architectures in environments with extreme resource limitations.
\subsubsection{Variance Reduction}
To address the fundamental challenge of high variance in the gradient estimation of ZO optimization, existing research proposes solutions across dimensions including statistical properties, system architecture, and temporal benchmarks. The standard zeroth-order gradient estimator inherently suffers from a variance proportional to the problem dimensionality, which severely limits the convergence rate. A unified framework formulates the variance-reduced estimator by introducing an auxiliary control variable to offset the stochastic noise without altering the expectation of the gradient. Following the logical evolution of these mechanisms, the methodologies transition from rudimentary gradient smoothing to structured variance control, and ultimately to snapshot-based variance reduction.\\
\textbf{Gradient smoothing.} To alleviate the excessive gradient noise inherent in zeroth-order evaluation, methods based on statistical properties focus on refining the moment estimations. For instance, R-AdaZO \cite{147} accomplishes efficient gradient smoothing through the collaborative design of first-order moment variance reduction and second-order moment refinement. This technique improves the stability of optimization without increasing the memory overhead. Furthermore, the theoretical convergence analysis indicates that variance-aware moment refinement ensures smoother optimization trajectories and accelerates the convergence process. The advantage of this approach lies in the enhancement of stability with a low memory footprint, whereas the inherent limitation is that the reliance on historical moments might introduce estimation bias in highly non-convex landscapes.\\
\textbf{Structured variance control.} Progressing from scalar moment smoothing to structural design, FZOO \cite{37} establishes a comprehensive variance control system through the formulation of structured perturbations and parallelized engineering optimization. By constraining the perturbation space, this method systematically mitigates the variance issues present in conventional zeroth-order techniques. Consequently, this approach integrates structured variance reduction to boost the efficiency of optimization and provide strict guarantees for theoretical convergence. The primary advantage is the significant improvement in computational efficiency through parallel structures, while the limitation is the heavy dependency on specific system architectures and the lack of universal applicability across different hardware setups.\\
\textbf{Snapshot variance reduction.} Leveraging temporal benchmarks, recent methodologies employ snapshot-based control variates to further suppress the variance of estimators. By incorporating historical gradient information, these techniques provide a more rigorous theoretical guarantee for convergence speed and final performance. Specifically, VAMO \cite{162} substitutes full first-order checkpoint gradients with zeroth-order estimates while preserving unbiased variance correction. Similarly, VR-SZD \cite{249} integrates stochastic zeroth-order estimation with control variates to align the optimization trajectory. In addition, MeZO-SVRG \cite{19} constructs low-variance estimators by combining information from the full batch and the mini-batch. The advantage of the snapshot paradigm is the strict reduction of theoretical variance bounds, whereas the limitation is the requirement for additional memory resources to store the checkpoint data.

The advancement of these methods demonstrate a clear paradigm shift from isolated statistical smoothing to structural perturbation constraints, and finally to temporal control variates. While early methods primarily focus on the utilization of immediate historical moments, recent advancements emphasize the structural alignment of the perturbation space and the integration of snapshot checkpoints to achieve optimal convergence rates. However, a critical gap remains in the existing literature. Most algorithms face an unavoidable trade-off between the reduction of variance and the consumption of memory resources. Developing optimal variance reduction frameworks that strictly bound the memory overhead while maintaining linear convergence for extremely large-scale models remains a significant challenge for future research.
\subsubsection{Distributed Zero-Order Optimization}
Some methods address the optimization of a global objective function distributed across multiple network nodes, relying exclusively on function evaluations. The general mathematical framework formulates this as a consensus optimization problem, wherein local nodes estimate gradients utilizing zeroth-order oracles while communicating over a defined topology. The core improvement dimensions in this domain primarily revolve around overcoming the inaccessibility of global information through distributed perturbation sampling and enforcing data protection through privacy preservation mechanisms. The evolutionary logic of these paradigms transitions from establishing basic distributed consensus to ensuring secure communication topologies. Under a unified formulation, local updates rely on Gaussian-smoothed zeroth-order estimators, which are subsequently aggregated across the network topology to approximate the global gradient.\\
\textbf{Distributed perturbation sampling.} ZoPro~\cite{93} represents the fundamental paradigm of information estimation in distributed consensus optimization scenarios where global data or Hessian matrices are inaccessible. The algorithm augments the SoPro framework with Gaussian-smoothed zeroth-order estimators for gradient or Hessian blocks, thereby bridging distributed SO optimization with ZO techniques. The advantage of this method lies in the efficient estimation of high-order information without exact gradients. The inherent limitation is the increased communication overhead and the dependency on the variance of the random perturbation.\\
\textbf{Privacy-preserving zeroth-order.} Advancing the consensus framework, TOP-DP~\cite{58} introduces the privacy-preserving dimension. The method adopts a topology-aware noise reduction strategy that reuses neighbor noise estimates and computes reduced-variance Gaussian noise via additive decomposition to satisfy differential privacy constraints. The integration of differential privacy with zeroth-order optimization addresses the need for gradient-free learning while safeguarding sensitive data in distributed scenarios. The primary advantage is the robust theoretical guarantee of data privacy. The inherent limitation is the severe trade-off between the scale of the injected noise and the convergence rate of the optimization process. Theoretical convergence analysis of these paradigms generally demonstrates that both methods achieve sublinear convergence rates. The variance introduced by topology-aware noise and zeroth-order estimators requires strict theoretical upper bounds to ensure the stability of the global objective function.

These methods have undergone a transition from pure distributed information estimation to topology-aware privacy preservation. The commonality across these methods is the reliance on variance-reduced Gaussian perturbation for gradient approximation. The difference lies in whether the perturbation is utilized for search exploration or strictly for cryptographic privacy. Despite these advancements, the core gap in existing work is the lack of a unified theoretical framework that optimally balances the zeroth-order query complexity, the network communication efficiency, and the strict differential privacy bounds, particularly under highly heterogeneous data distributions.

{\sloppy
\textbf{Transition: from mathematical primitives to scenario-oriented optimization.} Up to this point,~\cref{fo_methods,so_methods,zo_methods} have delineated the fundamental mathematical primitives of optimization, FO, SO, ZO, categorized primarily by their reliance on exact, approximated, or derivative-free gradient information. While ZO optimization profoundly demonstrates how mathematical compromises (e.g., high-variance noise perturbation) can circumvent rigid physical barriers like the backpropagation memory wall, it concurrently reveals a broader truth: theoretical convergence alone is insufficient for modern real-world deployment.
As deep learning scales relentlessly towards massive LLMs and expands into decentralized, privacy-sensitive ecosystems, optimization ceases to be a purely algorithmic endeavor. It morphs into a highly constrained systems engineering challenge. The mathematical primitives must be re-architected to navigate four formidable systemic bottlenecks:
First, when extreme communication overheads cripple decentralized networks, how do we compress or synchronize gradients across disparate nodes? (\cref{distr_methods}). Second, given that ZO already utilizes stochastic perturbations, how can we mathematically formalize noise injection to provide rigorous data protection without destroying model utility? (\cref{privacy_methods}). Third, if we demand the precision of FO and SO algorithms but face the stringent hardware limits targeted by ZO, how can we compress the persistent optimizer states (e.g., momentum, variance) of billion-parameter models? (\cref{me_methods}). Finally, confronted with the compounding noise from decentralized staleness, privacy injections, and low-rank compressions, how do we transcend human heuristics to automate and robustify optimizer design? (\cref{to_methods}).
In the subsequent sections, we shift our paradigm from unconstrained mathematical theory to scenario-driven optimization frameworks, exploring how state-of-the-art algorithms co-evolve with extreme environmental constraints.\par}

\subsection{Distributed Optimization}\label{distr_methods}
The general mathematical framework of distributed optimization aims to minimize a global objective function formulated as $F(x) = \frac{1}{n} \sum_{i=1}^{n} f_i(x)$ across multiple computing nodes. This global objective can be instantiated in the unified perspective~\cref{eq:unified_master} by taking the loss function $f$ as $F$, and capturing the distributed nature through the scenario transformation $\mathcal{T}_{\text{scenario}}$, which handles gradient aggregation, compression, or quantization across nodes during each iteration. The fundamental challenge within this framework lies in the communication bottleneck, which is addressed through two core dimensions: communication compression techniques and structural update strategies. The former mitigates communication overhead via projection operators such as low-rank approximation, quantization~\cite{84}~\cite{100}, and sparsification~\cite{80}, frequently employing adaptive levels and error compensation to ensure theoretical convergence and training stability. The unified formal derivation of these methodologies typically replaces $g_t$ with a compressed mapping $\mathcal{C}(g_t)$, ensuring that the variance of the compression error remains strictly bounded to maintain optimal convergence rates. The structural update strategies encompass local updates, decentralized communication protocols, and federated learning paradigms. Local update strategies, including Local SGD~\cite{87}~\cite{82}, control variates, and local momentum, balance the computational costs against communication latency while addressing data heterogeneity through adaptive step sizes. Decentralized approaches eliminate central node bottlenecks via optimized peer-to-peer topologies and asynchronous coordination. Federated learning targets systemic heterogeneity through client sampling, momentum fusion, and parameter personalization. Furthermore, hybrid frameworks integrate these components to secure reliable convergence performance under resource-limited conditions, although the inherent boundaries of such integration often manifest as a fundamental trade-off between statistical efficiency and hardware utilization rates.\\
\subsubsection{Gradient Compression \& Quantization}
To systematically address communication bottlenecks in distributed machine learning, related methods reduce the volume of gradient transmission via a unified compression framework $\mathcal{C}(\cdot)$. The evolution of these techniques follows a logical progression from static scalar reduction to dynamic, error-compensated, and adaptive mechanisms.\\
\textbf{Quantization compression.} Aimed at addressing the critical communication bottleneck directly at the representation level, this paradigm encodes gradient signals into lower-precision numerical formats, thereby significantly reducing the volume of data transfer and enabling the efficient scaling of large models across diverse compute clusters. The SIGNSGD algorithm~\cite{164} compresses gradients via direct sign transmission, whereas signProx~\cite{75} extends this mechanism with proximal operators for one-bit nonconvex optimization scenarios. The approach of 0/1 Adam~\cite{92} adaptively freezes variance states to facilitate pure one-bit compression. The method of 1-bit Adam~\cite{84} adopts a sophisticated two-stage strategy, in which quantization compression and the specific convergence characteristics of Adam are deeply coupled, stipulating that the one-bit quantization of momentum is performed solely after the preconditioner variance stabilizes. Furthermore, LQ-SGD~\cite{100} integrates log-quantization techniques into PowerSGD to minimize the loss of optimization accuracy. DEED-GD~\cite{77} utilizes double encoding and error diminishing within an inexact gradient framework, rendering it applicable to diverse distributed environments. TAH-QUANT~\cite{98} introduces tile-wise quantization for localized precision control, thus improving the efficiency of activation compression. These methods elegantly balance the reduction of communication payloads with rigorous theoretical convergence for distributed nonconvex computational tasks.\\
\textbf{Sparsification compression.} Moving beyond uniform precision reduction, sparsification reduces communication overhead by selectively transmitting solely the most critical gradient components while preserving established convergence boundaries. Within this distinct category, the rTop-k method~\cite{80} captures inherent gradient sparsity via a parameterized statistical model. SPARQ-SGD~\cite{90} employs event-triggered communication architectures, whereby compressed differences are transmitted exclusively when the absolute changes of parameters surpass a predefined threshold. These methods facilitate highly scalable distributed training by balancing the extreme reduction of communication with robust convergence guarantees.\\
\textbf{Low-rank gradient compression.} To further exploit structural redundancies in optimization trajectories, certain methods compress high-dimensional gradients into low-rank subspaces, which is critical for scaling to massive worker counts and ultra-high-dimensional models while preserving the rate of convergence and final accuracy. PowerSGD~\cite{74} leverages iterative power methods for the rapid low-rank approximation of gradients. The framework of DOME~\cite{73} extends this matrix methodology to decentralized, privacy-preserving settings via correlation-aware sketching and secure mathematical aggregation. SketchedAMSGrad~\cite{89} adapts Adam-type preconditioning algorithms with matrix sketching, thereby exponentially reducing the cost of communication. These techniques construct a distinct structural paradigm that accelerates distributed learning by balancing low-rank approximation formulations with optimal convergence trajectories.\\
\textbf{Compression error compensation.} To address the inherent information loss and convergence limitations of the aforementioned static compression operators, error compensation mechanisms are introduced to systematically track and mitigate the subsequent accumulation of compression-induced variance. APMSqueeze~\cite{83} synthesizes the preconditioning mechanisms of Adam with error-compensated gradient compression, achieving a significant and stable reduction in communication payloads. ADEF~\cite{97} integrates contractive compression operators directly with mathematical error feedback loops. These approaches enhance the efficiency of distributed training by preserving strict theoretical bounds on the internal compensation memory, which is thoroughly supported by comprehensive theoretical convergence analysis and empirical performance gains.\\
\textbf{Adaptive compression level.} Advancing toward fully dynamic optimization protocols, adaptive methods resolve the fundamental problem that fixed compression hyperparameters frequently fail to accommodate dynamic network bandwidth conditions or evolving gradient properties, which inevitably leads to suboptimal algorithmic convergence. AdaCGD~\cite{91} mathematically extends the three-point compressor framework to robustly support bidirectional dynamic compression. DeCo-SGD~\cite{101} utilizes the Nested Virtual Sequence theoretical tool to comprehensively analyze the temporal interactions between synchronization staleness and data compression, dynamically adjusting both operational elements based on real-time network conditions. LAGS-SGD~\cite{78} synthesizes layer-wise adaptive sparsification protocols seamlessly with synchronous gradient descent algorithms. These methods dynamically continuously optimize the operator $\mathcal{C}(\cdot)$ to balance communication efficiency and iteration convergence, substantially enhancing the mathematical robustness of distributed optimization across widely varying hardware topologies and network conditions.

Furthermore, the paradigm shifts in gradient compression demonstrate a clear evolutionary logic, transitioning comprehensively from naive payload truncation to topology-aware, error-corrected, and fully dynamically adaptive compression matrices. Regarding the contrast of methodological advantages and inherent limitations, quantization offers unprecedented compression ratios but frequently suffers from strict precision bottlenecks within highly complex optimization landscapes; sparsification maintains precise magnitude accuracy but systematically incurs significant hardware overhead for index encoding; low-rank projections effectively capture global mathematical structures but inherently demand high computational costs for continuous matrix decomposition. The core gap of existing analytical work resides in the theoretical void concerning the optimal, mathematically continuous scheduling of hybrid compression operators under non-independent and identically distributed data settings, which emphatically highlights the necessity for unified algorithmic frameworks that dynamically balance statistical sampling efficiency, local computational overhead, and global network bandwidth constraints.

\subsubsection{Local Update Strategies}
These algorithms represent a critical adaptation of optimization paradigms for distributed learning environments. The general framework of these methods transforms synchronous gradient descent by allowing each distributed node to perform a sequence of local gradient steps prior to global aggregation. This structural modification establishes the core improvement dimensions: minimizing communication overhead while mitigating the adverse effects of data heterogeneity. The evolutionary logic of these strategies follows a trajectory from basic step delays to sophisticated variance correction and dynamic scheduling, addressing the fundamental trade-off between local computation efficiency and global convergence stability.\\
\textbf{Local SGD.} The foundational paradigm initiates with this approach, which addresses communication bottlenecks by executing multiple iterations locally before synchronization. The SQuARM-SGD algorithm~\cite{87} transmits compressed model differences exclusively when a defined threshold is exceeded, achieving consensus through the averaging of weighted graphs. The SLOWMO framework~\cite{82} integrates periodic momentum updates via nested loops, systematically bridging the sequence of local steps with global synchronization and momentum updates. The method advantage of this foundational approach is the significant reduction in communication frequency and the acceleration of local convergence. The inherent limitation, however, is its vulnerability to client drift under non-independent and identically distributed data regimes, as uncorrected local trajectories progressively diverge from the global optimum.\\
\textbf{Local momentum updates.} Building upon the standard local iterations, this evolutionary step introduces temporal smoothing to the unified form derivation of the update rule. The SQuARM-SGD method~\cite{87} extends into this domain by combining Nesterov~\cite{324} with multiple local steps. By incorporating historical gradients, methods such as FAdamGC~\cite{119} position gradient correction ahead of the momentum computation to explicitly resolve the discrepancy of client models. The method advantage is the enhancement of distributed learning scalability through stabilized optimization paths. The inherent limitation is the increased local memory footprint required to maintain momentum buffers across all distributed nodes.\\
\textbf{Variance-reduced local updates.} To formally address the limitations of client drift identified in earlier methods, the trajectory progressed toward variance reduction techniques, which modify the local update equations through the incorporation of correction terms. The BVR-L-SGD method~\cite{85} employs a variance-reduced gradient estimator, comparable to the SARAH algorithm, to jointly minimize the stochastic variance of local iterations and the bias of the global gradient. Additionally, approaches like SCAFFOLD~\cite{103} utilize control variates to approximate and rectify the update directions between the server and the clients. Theoretical convergence analysis demonstrates that these strategies can recover the optimal convergence rates of centralized algorithms despite arbitrary data distributions. The method advantage is the rigorous theoretical elimination of optimization bias caused by data heterogeneity. Conversely, the inherent limitation involves the substantial amplification of communication and memory overhead, as the framework mandates the synchronization of auxiliary control variables that are dimensionally equivalent to the model parameters.\\
\textbf{Local-global hybrid updates.} Recognizing the communication constraints of variance-reduced methods, hybrid strategies decompose the optimization process across structural dimensions. These methods balance communication efficiency and convergence consistency by accumulating partial parameters locally while synchronizing key components globally. The DeCo-SGD framework~\cite{101} dynamically coordinates the delay of local steps with the ratio of global compression. Similarly, HybridSGD~\cite{96} merges local multi-step updates along the row dimension with batch synchronization along the column dimension within a two-dimensional grid topology. Ringleader ASGD~\cite{344} leverages a structured two-phase update mechanism to bound stale gradients. The method advantage of this paradigm is the effective circumvention of the full-parameter synchronization bottleneck, significantly alleviating bias accumulation from purely local steps. The inherent limitation is the increased implementation complexity, which restricts its applicability across highly asymmetric network topologies.\\
\textbf{Adaptive local steps.} The final dimension of algorithmic evolution transitions from fixed synchronization intervals to dynamic temporal scheduling, adapting the local steps to the real-time state of the optimization process. The FedCET algorithm~\cite{118} utilizes the adaptive weighting of learning rates to constrain objective drift and reduce communication costs. The AbsSADMM approach~\cite{246} modulates the batch size based on the historical trajectory of the optimization path. Furthermore, DP-PASGD~\cite{104} integrates differential privacy constraints with periodic averaging to maintain a balance among model accuracy, computational efficiency, and data privacy. The method advantage of adaptive strategies is their superior flexibility and resource utilization in highly heterogeneous distributed scenarios. The inherent limitation is the reliance on heuristic thresholding mechanisms, which presents significant challenges for establishing tight theoretical bounds on worst-case convergence guarantees.

The milestone shift in these algorithms is the transition from heuristic communication reduction to theoretically grounded, dynamic drift correction. While existing methods successfully adapt optimization principles to distributed settings, a significant gap remains in the literature. Specifically, current theoretical convergence analyses heavily depend on bounded gradient assumptions, and the field lacks a unified framework capable of achieving exact consensus under simultaneous extremes of statistical heterogeneity and system resource asymmetry, without incurring prohibitive costs in auxiliary variable communication or hyperparameter optimization.
\subsubsection{Decentralized Communication}
This strategy serve as the foundational mechanism in distributed learning to optimize peer-to-peer interaction patterns and global consensus mechanisms. The general framework formulates the global objective as the minimization of the aggregated local loss functions across all nodes, where the update rule relies on a mixing matrix to govern information exchange. The core improvement dimensions of recent research focus on optimizing the communication graph, relaxing synchronization constraints, enhancing network reliability, and integrating rigorous confidentiality guarantees.\\
\textbf{Neighbor communication topology.} Regarding the optimization of network structures, algorithms adjust the interaction graph to balance the trade-off between local gradient updates and global consensus. Methods such as LD-SGD~\cite{88} introduce alternating schemes between local updates and decentralized first-order steps, establishing an analytical framework for non-convex and non-independent and identically distributed environments. Furthermore, DAT-SGD~\cite{94} extends the parallelism thresholds through the anytime stochastic gradient descent framework. By utilizing averaged query points, this approach mitigates local inconsistencies and reduces the consensus distance, thereby improving the theoretical bounds of parallelism.\\
\textbf{Distributed consensus optimization.} To achieve exact alignment, certain approaches prioritize the reduction of communication overhead while maintaining precise consensus. For instance, LT-ADMM~\cite{95} employs local training to decrease communication frequency, applying standard stochastic gradient descent~\cite{307} for local parameter updates. In addition, DLAS-R-FTC~\cite{102} implements automated stepsize selection through finite-time coordination. These methods eliminate the heterogeneity of the network and optimize the consensus mechanism by adapting to the underlying graph structure.\\
\textbf{Asynchronous decentralization.} To overcome the limitations of synchronous protocols, asynchronous methods relax the strict global clock. The A(DP)$^2$SGD~\cite{86} circumvents the efficiency bottleneck caused by waiting for stragglers. Through flexible topology adaptation and asynchronous update strategies, this method balances the extension of parallelism with local consistency, thereby accelerating the convergence of decentralized optimization.\\
\textbf{Communication fault robustness.} Progressing toward resilient systems, robustness strategies address the unreliability of realistic networks. The DES-LOC method~\cite{99} decouples the synchronization intervals of the model parameters and the optimizer states. This decoupling significantly reduces communication costs by transmitting the optimizer state less frequently while preserving the convergence rate. Such an approach provides a scalable and fault-tolerant solution for distributed training under high communication overhead.\\
\textbf{Privacy-preserving decentralization.} Finally, privacy-aware frameworks address the vulnerability of gradient sharing. To prevent the leakage of sensitive information, Interleaved-ShuffleG~\cite{70} merges private and public data. This framework minimizes the empirical excess risk through surrogate objectives and adaptive noise injection. Similarly, FedLAP-DP~\cite{115} utilizes synthetic samples to approximate and aggregate local loss landscapes into a unified global representation. This technique integrates record-level differential privacy without incurring supplementary computational costs.

These methods exhibit a shift from rigid synchronous topologies to asynchronous, robust, and privacy-aware protocols. This evolution reflects the adaptation of traditional optimization methods to complex distributed scenarios. Synchronous topology methods offer the advantage of straightforward theoretical guarantees but suffer from the inherent limitation of straggler bottlenecks. Consensus-driven approaches improve resource utilization efficiency; nevertheless, they are limited by the complexity of managing delayed gradients and second-order approximations. Asynchronous methods maximize hardware utilization at the cost of introducing gradient staleness. Robustness frameworks secure the training process against node failures but introduce significant algorithmic complexity. Privacy-preserving protocols prevent data leaks, yet they inject inherent optimization noise. The core gap in existing research lies in the absence of a unified theoretical framework capable of simultaneously achieving optimal communication complexity, strict privacy guarantees, and exact consensus without deteriorating the convergence properties of optimization methods in extremely heterogeneous environments.
\subsubsection{Federated Learning Optimization}
Federated learning optimization addresses the challenge of collaborative training across dispersed clients. The general mathematical framework formulates this as a distributed empirical risk minimization problem, typically expressed as $\min_{\theta} \sum_{k=1}^{K} p_k F_k(\theta)$, where $p_k$ represents the weight of the client and $F_k(\theta)$ denotes the local objective function. This global objective can be embedded into the unified perspective~\cref{eq:unified_master} by setting the loss function $f$ as the weighted sum $\sum_k p_k F_k(\theta)$, and by incorporating the client-level interactions (e.g., local updates, aggregation, privacy mechanisms) into the scenario transformation $\mathcal{T}_{\text{scenario}}$, which operates on per-client gradient estimates $\tilde{g}_t^{(k)}$ to produce the aggregated gradient $\hat{g}_t$. To tackle specific challenges arising from statistical and system heterogeneity, recent methodologies enhance this framework through three core improvement dimensions: first-order momentum acceleration, second-order curvature approximation, and zeroth-order or structural adaptations including client sampling and personalization. These techniques aim to harmonize global model convergence with personalized adaptation while mitigating communication bottlenecks and the phenomenon of client drift.\\
\textbf{First-order methods and unified momentum derivation.} The evolution of federated optimization initially focused on first-order methods, establishing a unified formulation where the global update incorporates historical gradient information, denoted as $m_t = \beta m_{t-1} + (1-\beta) g_t$, to stabilize the trajectory. By leveraging advanced momentum fusion, methods in this category alleviate heterogeneity issues along multiple dimensions. Specifically, FEDAC~\cite{79} avoids the heterogeneity bias resulting from local momentum accumulation through prepositioned gradient correction and the dynamic coordination between control variables and momentum updates. Following a similar principle, FedLion~\cite{110} initializes the local momentum using the global momentum at clients, subsequently uploading the momentum to the server for aggregation after multi-step local updates. The utilization of symbolized gradients reduces communication overhead, while the global momentum constrains the local heterogeneity drift. To enforce alignment between the local and global update directions, FedMuon~\cite{293} incorporates momentum aggregation. Furthermore, AdaFedAdam~\cite{107} fuses the local momentum via deterministic weighting based on local updates, wherein the global momentum adaptively adjusts per aggregated determinism to balance local heterogeneous information and global consensus. FADAS~\cite{111} dynamically adjusts the weight of the global momentum through a delay-adaptive learning rate. From a structural perspective, FedRepOpt~\cite{113} enables local momentum accumulation to align with the globally optimal structure learned by the server. Concurrently, FLeNS~\cite{112} combines Nesterov~\cite{324} acceleration with Hessian sketching to bridge the gap toward second-order techniques.\\
\textbf{Federated second-order optimization.} Transitioning beyond first-order paradigms, federated second-order optimization methods introduce curvature information to address the slow convergence, which remains a prominent limitation that first-order methods often fail to resolve effectively. The representative algorithmic formulation of these methods approximates the inverse Hessian matrix or the FIM to scale the gradient updates. For instance, FAGH~\cite{114} utilizes an approximated global Hessian matrix, circumventing the necessity of full Hessian storage to accelerate convergence. To enhance personalization, pFedSOP~\cite{123} constructs a regularized FIM via personalized updates based on the Gompertz function. Additionally, BFEL~\cite{120} integrates FedCurv, which employs the FIM for the preservation of client-specific knowledge, into a blockchain framework for trusted healthcare applications. These methods systematically improve both convergence speed and personalization capabilities through the effective utilization of second-order information.\\
\textbf{Zeroth-order and advanced structural adaptations.} Beyond gradient-based modifications, alternative adaptations focus on structural optimization, including client sampling, probabilistic personalization, and phase synchronization. Client sampling optimization mitigates communication overhead and update bias by refining the selection process. FedSTaS~\cite{117} integrates gradient-based stratification with data sampling, while FAdamGC~\cite{119} achieves sampling optimization via selective tracking, pivoting on the decoupling of training client sampling from the sampling of control variable updates. Furthermore, FedOne~\cite{124} activates a single client per round in black-box prompt learning scenarios. In the context of personalized federated optimization, FedIvon~\cite{116} approximates local posteriors with Gaussian distributions for personalized modeling based on a Bayesian framework. Targeting block-cyclic data, MM-PSGD and MC-PSGD~\cite{109} construct block-specific predictors via complementary training strategies. Addressing the aggregation phase, Kuramoto-FedAvg~\cite{122} reframes the aggregation process as a synchronization problem using the Kuramoto model, dynamically adjusting the update weight of each client based on phase alignment with the global update to reduce client drift.

The core design paradigm has shifted from simplistic gradient averaging to sophisticated, topology-aware, and curvature-informed information fusion. This evolution reflects a logical progression from directly mitigating gradient variance to adapting to local loss landscapes, and finally to optimizing the broader system architecture. The advantage of first-order methods lies in algorithmic simplicity, but they lack geometric awareness. Second-order methods capture geometric structures effectively, but the utility of these methods is constrained by matrix inversion costs. Structural adaptations offer maximal flexibility, though they risk instability in asynchronous environments. Despite these milestones, a critical gap remains in the literature: existing methods lack a unified theoretical framework capable of simultaneously bounding the generalization error and communication complexity under dynamic, non-stationary heterogeneity, necessitating future research into adaptive, assumption-free optimization algorithms.
\subsubsection{Communication Scheduling\&Threshold}
Communication scheduling and threshold techniques are mechanisms employed to orchestrate the timing and priority of data transmission. These methods regulate information exchange based on update significance and network conditions to alleviate bandwidth congestion and mitigate the impact of latency. Under a unified mathematical framework, the communication process at iteration $t$ can be abstracted as an operator $\mathcal{C}(\Delta_t, \mathcal{S}_t$), where $\Delta_t$ denotes the parameter or gradient update, and $\mathcal{S}_t$ represents the environmental state, such as network bandwidth or latency. Furthermore, these scheduling paradigms serve as essential adaptations for various optimization methods across first-order, second-order, and zeroth-order dimensions, tailoring their specific update characteristics to distinct communication scenarios. The core improvement dimensions of these techniques focus on designing the operator $\mathcal{C}$ to adaptively filter, reorder, or delay the transmission of $\Delta_t$ to optimize the trade-off between communication overhead and model convergence.\\
\textbf{Significance threshold communication.} The first paradigm focuses on the spatial or temporal filtering of updates. Diverging from fixed-interval synchronization, approaches such as SPARQ-SGD~\cite{90} adopt an event-driven mechanism. The communication operator is formulated as an indicator function $\mathbb{I}(||\Delta_t|| > \tau)$, where communication is triggered only when the magnitude of parameter updates exceeds a critical threshold $\tau$ after $H$ local steps. By transmitting compressed differences solely upon significant changes in the model, this approach prioritizes the reduction of communication frequency over continuous synchronization. This paradigm is distinctively suitable for environments constrained by bandwidth, though it inherently risks the loss of subtle gradient information.\\
\textbf{Priority communication.} The second paradigm shifts the focus from volume reduction to structural optimization. While threshold methods reduce communication volume by discarding data, DLCP~\cite{76} optimizes the transmission order of $\Delta_t$. Under the unified framework, the operator $\mathcal{C}$ acts as a permutation matrix that reorders the transmission sequence based on a priority score assigned to different gradient layers. This method implements a fine-grained, packet-level prioritization strategy. It ensures that critical gradient information takes precedence, thereby maximizing network utilization and transmission efficiency without discarding data. From a theoretical perspective, priority scheduling preserves the unbiased nature of the full gradient, which maintains tighter convergence bounds compared to threshold-based pruning. However, the computational overhead required to calculate priority scores introduces additional latency constraints.\\
\textbf{Delay-aware scheduling.} The third paradigm addresses temporal dynamics and environmental stochasticity. Unlike the static configurations of previous methods, DeCo-SGD~\cite{101} introduces adaptivity to the environment through delay-aware scheduling. In this context, the operator $\mathcal{C}$ dynamically modulates both compression ratios and staleness tolerance in response to fluctuating network conditions $\mathcal{S}_t$. This approach addresses the convergence instability caused by external network variance, ensuring robustness in scenarios where rigid threshold or priority schemes fail. The theoretical boundary of this method lies in the bounded delay assumption, where the convergence rate degrades smoothly with respect to the maximum communication delay rather than diverging abruptly.

The evolutionary logic demonstrates a clear transition from passive filtering to active environmental adaptation. The paradigm shifts from data-agnostic synchronization to the data-aware thresholding of SPARQ-SGD~\cite{90}, further evolving into the structure-aware scheduling of DLCP~\cite{76}, and culminating in the environment-aware dynamic modulation of DeCo-SGD~\cite{101}. Regarding the comparative analysis of advantages and inherent limitations, threshold methods offer maximum bandwidth savings but suffer from precision degradation. Priority schemes maintain mathematical precision but incur significant computational overhead for sorting. Delay-aware scheduling provides optimal robustness but requires complex system-level profiling. The core gap in existing research is the absence of a unified co-design framework that jointly optimizes the threshold parameter, priority order, and delay tolerance under a rigorous convergence guarantee, particularly in heterogeneous and highly dynamic network topologies.
\subsubsection{Distributed Hybrid Optimization}
The domain of distributed hybrid optimization encompasses integrated frameworks that fuse diverse algorithmic components, including quantization, curvature estimation, and privacy protocols. The core improvement dimensions of these approaches span from basic communication reduction to advanced curvature adaptation and privacy preservation.\\
\textbf{Compression\&local updates.} By combining parameter compression and local computations, certain methods address the critical communication bottleneck in distributed optimization. The unified formal derivation of these algorithms typically involves applying a mathematical compression operator to accumulated local gradients or model differences before aggregation. SQuARM-SGD~\cite{87} integrates multi-step local updates and arbitrary compression strategies for peer-to-peer networks. Qsparse-local-SGD~\cite{81} unifies quantization, sparsification, and local computations with error compensation, which supports synchronous and asynchronous modes for diverse distributed settings. The primary advantage of this paradigm is the significant reduction in bandwidth requirements, whereas the inherent limitation is the potential degradation of convergence accuracy due to the accumulation of compression variance.\\
\textbf{Quantization\&low-rank.} Following the evolutionary logic of extreme compression, LQ-SGD~\cite{100} incorporates logarithmic quantization into the PowerSGD framework to further reduce the costs of communication while maintaining the accuracy of the model. This method combines orthogonal compression techniques to balance the efficiency of communication and the performance of optimization in distributed settings. The advantage of this approach lies in the synergistic effect of structural and precision reduction, while the limitation is the increased computational overhead required for the decomposition of low-rank matrices during each communication round.\\
\textbf{Second-order\&distributed.} To transcend the limitations of first-order methods, recent advancements focus on merging curvature-aware second-order optimization with distributed systems. This integration addresses the high cost of full second-order methods and the slow convergence of first-order approaches in ill-conditioned landscapes. Fed-Sophia~\cite{265} utilizes a lightweight estimation of the Hessian diagonal, which allows the algorithm to incorporate curvature information without computing the full Hessian matrix, thus ensuring the efficiency of both communication and computation. The advantage is the accelerated convergence in complex topographies, while the limitation remains the sensitivity to the estimation variance of the diagonal preconditioner under heterogeneous data distributions.\\
\textbf{Privacy\&distributed.} Addressing the orthogonal requirement of data security, optimization frameworks must integrate rigorous protection mechanisms. TOP-DP~\cite{58} rigorously verifies theoretical privacy guarantees through the framework of differential privacy, which is suitable for visual tasks such as image classification. The advantage is the provable bound on information leakage, whereas the limitation is the inherent trade-off between the scale of noise injection and the final utility of the model.

The evolution of distributed hybrid optimization exhibits a paradigm shift from heuristic single dimensional compression to theoretically guaranteed, multi dimensional joint designs. Initial methods primarily focused on reducing communication via simple gradient quantization, which gradually evolved into sophisticated frameworks that simultaneously manage variance reduction, curvature approximation, and privacy noise. Despite these advancements, a significant gap remains in the existing work. Specifically, there is an absence of a unified theoretical framework that can simultaneously bound the convergence errors introduced by extreme compression, the inaccurate estimation of second-order curvature, and the strict constraints of differential privacy, particularly under conditions of severe data heterogeneity.
\subsection{Privacy-Preserving Optimization}\label{privacy_methods}
The framework of privacy-preserving optimization can generally be divided into fundamental algorithmic frameworks and trade-off management strategies. The primary distinction between these aspects relies on whether the objective focuses on the mechanism of privatization or the mitigation of adverse side effects on convergence. Within the fundamental frameworks, differential privacy optimization serves as the core paradigm. This approach typically revolves around modifying vanilla optimizers to incorporate rigorous privacy guarantees while addressing computational inefficiencies. By formulating the privatized update rule as $\theta_{t+1} = \theta_t - \eta_t \big[ \frac{1}{B} \big( \sum_{i \in B_t} \Pi_C(\nabla J(x_i, \theta_t)) + \mathcal{N}(0, \sigma^2 C^2 I) \big) \big]$, where $\Pi_C$ represents the per-example gradient clipping operator and $\sigma$ denotes the noise multiplier, this foundational paradigm can be seamlessly encapsulated into our unified formulation~\cref{eq:unified_master}, from which various methods are extended across different optimization orders. Furthermore, specific attention is directed toward gradient noise injection, which strategically perturbs computations to enforce privacy protocols. These techniques vary from adaptive intensity adjustment and scalar injection to noise-robust designs and geometric perturbations that align noise with the direction of the gradient. To manage the side effects, trade-off management strategies aim to reconcile data protection with model fidelity, predominantly relying on adaptive mechanisms to mitigate the impact of noise. Several studies focus on dynamic noise scheduling or post-processing optimization to refine gradients after noise injection. A closely related technique is privacy-aware gradient clipping, which regulates the magnitude of gradients to prevent instability. These methods encompass global clipping strategies, adaptive clipping based on gradient statistics, and curvature-aware clipping to guide updates in non-convex landscapes. Federated privacy enhancement extends these concepts to collaborative environments, where approaches focus on neutralizing noise amplification during aggregation through protection protocols and structural adaptations. The protection protocols involve federated gradient shuffling to mix private and public data. The structural adaptations target federated noise aggregation, employing techniques such as SVD reparameterization or SAM-based optimizers to ensure robustness against aggregated noise. Additionally, low-rank adaptations are utilized to eliminate truncation bias in heterogeneous settings.\\
\subsubsection{Differential Privacy Optimization}
Differential privacy optimization encompasses specialized algorithmic frameworks implemented within privacy-preserving training to resolve the tension between rigorous data protection and model fidelity. These techniques function to mitigate the adverse effects of noise injection and gradient bias through adaptive mechanisms and error correction, thereby maximizing utility and convergence stability without compromising privacy guarantees.\\
\textbf{DP-SGD variants.} Addressing the key limitations of the vanilla DP-SGD~\cite{59}, including high noise impact, biased optimizer estimates, and computational inefficiencies, requires targeted modifications across optimization orders. For first-order methods, TOP-DP~\cite{58} reduces noise variance through topology-aware reuse and time decay, whereas DPIS~\cite{57} utilizes importance sampling and adaptive clipping to achieve lower variance and privacy cost. DP-$\lambda$CGD~\cite{343} further enhances first-order privacy training by introducing lightweight iterative noise correlation. To address second-order momentum adaptations, DP-MicroAdam~\cite{295} utilizes top-$k$ sparsification to filter out updates with low signal-to-noise ratios for bias reduction, and it employs an error feedback accumulation formula to recover lost signals. Furthermore, Corrected DP-Adam~\cite{60} corrects the bias in the second moment estimation under differential privacy, while DP-AdamW~\cite{292} applies decoupling to make the regularization strength independent of instantaneous gradient noise levels. DP-aware AdaLN-Zero~\cite{342} mitigates conditioning-induced heavy-tailed gradients that trigger excessive clipping in conditional diffusion modelsTransitioning to zeroth-order optimization, SPARTA~\cite{69} implements structured perturbations to achieve efficiency under privacy constraints. Additionally, DP-SGD-JL~\cite{56} leverages Johnson-Lindenstrauss projections to reduce the computation overhead of gradient norms, DOPPLER~\cite{67} applies frequency-domain filtering to boost the signal-to-noise ratio, and RaCO-DP~\cite{72} adapts gradient ascent for rate-constrained tasks with theoretical convergence guarantees.\\
\textbf{Joint differential privacy.} To provide a theoretical convergence analysis, studies focus on deriving bounds for population loss. The method $A_{\text{NSGD}}$~\cite{61} derives theoretical bounds on population loss in stochastic convex optimization, and empirical studies on multi-class classification effectively validate the validity of these insights. It provides specialized solutions to balance differential privacy and optimization performance in joint learning scenarios.\\
\textbf{Dynamic noise scheduling.} Moving beyond static parameters, dynamic scheduling algorithms adjust noise injection based on the training context or data properties to address the rigid limitations of static noise levels. TOP-DP~\cite{58} employs topology-aware noise reuse and time-aware decay to reduce noise progressively over the training process. Similarly, DC-SGD~\cite{68} dynamically sets clipping thresholds utilizing differential privacy histograms of gradient norms. These approaches balance privacy guarantees and model utility through context-aware adjustments.\\
\textbf{Privacy-utility balance.} The critical trade-off between differential privacy guarantees and model utility necessitates advanced balancing strategies, as excessive noise degrades performance while insufficient noise compromises privacy. In distributed settings, TOP-DP~\cite{58} uses topology-aware noise reuse to maintain utility. Alternatively, Interleaved-ShuffleG~\cite{70} combines private and public data within a hybrid framework to balance differential privacy and utility through tailored data usage.

The evolutionary trajectory of privacy-preserving optimization demonstrates a fundamental shift from static, uniform noise injection to adaptive, structure-aware perturbations. The milestone paradigm transition involves adapting the three major optimization orders to differential privacy constraints. First-order modifications primarily focus on adaptive clipping and variance reduction. SO adaptations attempt to correct the biased moment estimations caused by noise injection. ZO methods explore structured perturbations for derivative-free privacy. The primary advantage of these methodologies is the provision of rigorous mathematical guarantees for data protection. However, the inherent limitation is the unavoidable degradation of convergence rates and model utility, particularly in highly non-convex landscapes. The core gap in existing research remains the lack of a unified theoretical framework capable of optimally bounding the privacy-utility trade-off across diverse optimization orders without relying on extensive hyperparameter tuning or prior knowledge of the data distribution.
\subsubsection{Gradient Noise Injection}
Gradient noise injection enforces differential privacy in model training via the update $\theta_{t+1} = \theta_t - \eta (\text{Clip}(g_t) + \mathcal{N}(0, \sigma^2 I))$, perturbing bounded gradients. Recent methods calibrate noise using adaptive and geometric mechanisms and enhance optimization resilience via second order and zeroth order approximations to preserve theoretical convergence and empirical performance.\\
\textbf{Adaptive noise intensity.} First order optimization conventionally applies uniform vector perturbation~\cite{59}. This method dynamically calibrates this; DC-SGD~\cite{68} adjusts clipping thresholds via gradient norm statistical distributions to optimize computation privacy tradeoffs.\\
\textbf{Geometric noise injection.} GeoDP-SGD~\cite{65} structurally decomposes perturbations,  which separates gradient direction and magnitude, offering granular control over isotropic noise to improve structural efficiency.\\
\textbf{Noise robust optimization.} Addressing privacy induced variance, DP-Adam~\cite{60} corrects second moment estimation bias when standard Adam~\cite{287} processes noisy gradients. For landscape smoothing, DP-FedSAM~\cite{106} integrates SAM~\cite{131} to locate flatter loss regions, naturally resisting gradient perturbations and stabilizing high noise convergence.\\
\textbf{Scalar noise injection.} In communication constrained vertical federated learning. DPZV~\cite{105} replaces high dimensional vector perturbations with scalar noise. Discarding explicit FO directional derivatives, it theoretically matches their convergence rates, balancing strict privacy with low communication overhead to achieve distributed memory efficiency.

The design paradigm evolves from isotropic perturbation to geometry aware scaling, landscape regularization, and memory efficiency. Adaptive methods offer real time calibration but increase per step computational complexity. Robust optimizers stabilize convergence but demand rigorous hyperparameter tuning preventing underfitting. ZO methods minimize communication but assume strict convexity for convergence bounds. A unified framework guaranteeing tight theoretical convergence, minimal memory footprint, and geometry aware calibration across heterogeneous nonconvex landscapes remains absent.

\subsubsection{Privacy-Utility Tradeoff}
To mitigate the utility loss caused by DP, existing studies have optimized from two dimensions: gradient clipping and noise handling. DC-SGD~\cite{68} focuses on dynamic adjustment during the gradient clipping phase. This framework adaptively sets the clipping threshold using a differentially private histogram of gradient norms, reducing the burden of manual hyperparameter tuning while improving optimization efficiency. In contrast, DP-LSSGD~\cite{55} devotes itself to signal recovery in the post-processing stage. By introducing a real-time denoising mechanism based on Laplacian smoothing, it smooths the gradients injected with Gaussian noise. These two methods effectively enhance model utility while maintaining strict DP guarantees, by optimizing the input clipping threshold and output noisy gradients, respectively.

\subsubsection{Federated Privacy Enhancement}
Federated privacy enhancement uses specialized protocols in collaborative learning ecosystems to fortify data confidentiality during aggregation. The mathematical framework defines the server update as $\theta_{t+1} = \theta_t - \eta \sum_{i=1}^K w_i \mathcal{M}(\Delta \theta_t^{(i)})$, with $\mathcal{M}(\cdot)$ being a randomized privacy mechanism for local updates. This update rule can be instantiated in the unified perspective~\cref{eq:unified_master} by interpreting the per-client local update $\Delta \theta_t^{(i)}$ as the gradient estimate $\tilde{g}_t^{(i)} = \mathcal{E}_i(f, \theta_t, \xi_t^{(i)})$ from client $i$, embedding the privacy mechanism and aggregation into the scenario transformation $\mathcal{T}_{\text{scenario}}$ such that $\hat{g}_t = \sum_i w_i \mathcal{M}(\tilde{g}_t^{(i)})$. Core improvements neutralize noise amplification and prevent information leakage via architectural reparameterization and algorithmic resilience. Theoretical convergence analysis shows that controlling distributed noise variance enables algorithms to approach non-private baseline convergence rates.\\
\textbf{Architecture collaboration.} To mitigate utility loss under strict differential privacy constraints, this dimension optimizes gradient interaction modes. The method in~\cite{70} introduces Interleaved-Shuffle, reformulating collaborative architecture and transmission mechanisms. Utilizing encrypted gradient transmission, it distributes noise calibration across clients and the server. This structural adaptation bounds aggregation function sensitivity, reducing privacy leakage risks without altering fundamental first-order optimization dynamics.\\
\textbf{Optimizer-level robustness.} Progressing from transmission modifications to intrinsic algorithmic stability, DP-FedSAM~\cite{106} enhances algorithmic tolerance to aggregated noise. Addressing noise amplification during server aggregation, it integrates the client-side SAM optimizer. Rather than modifying transmission structures, this approach identifies flat minima within local loss landscapes. Exploring a broader solution space grants the objective function landscape-aware robustness, improving resistance to gradient perturbations while maintaining theoretical convergence guarantees.

The design paradigm evolves from structural transmission modifications to landscape-aware algorithmic resilience. Architecture collaboration provides strict privacy guarantees via transmission encryption but exhibits high communication overhead. Optimizer-level robustness mitigates local noise amplification and improves generalization, yet requires substantial local computational resources. A unified framework simultaneously minimizing communication overhead, optimizing local landscape flatness, and adapting to heterogeneous data distributions under strict privacy constraints remains absent.

\subsubsection{Privacy-Aware Gradient Clipping}
Privacy-aware gradient clipping bounds algorithmic sensitivity in DP and stochastic optimization. The mathematical framework updates gradients via $\theta_{t+1} = \theta_t - \eta \hat{g}_t$, with the clipped gradient $\hat{g}_t = \min(1, \frac{C}{\|\tilde{g}_t\|}) \tilde{g}_t$ restricting update magnitude through a threshold $C$. Core improvements transition from static global constraints to dynamic statistical adaptations and curvature-informed reparameterization.\\
\textbf{Global gradient clipping.} Some methods enforces a uniform optimization constraint. Addressing DP in stochastic optimization, AClipped-dpSGD~\cite{64} uses a one-time global clipping strategy for convex problems. For similarity-based objectives, Logit-DP~\cite{66} clips logit gradients before loss computation. Although targeting distinct scenarios, both rely mathematically on predefined global thresholds to restrict sensitivity, reduce noise, and establish fundamental theoretical convergence bounds.\\
\textbf{Adaptive clipping.} Somw methods introduce dynamic first-order modulation. By continuously estimating gradient sequence statistics, these methods calibrate clipping thresholds in real-time. For example, DC-SGD~\cite{68} constructs DP histograms of gradient norms to dynamically approximate the underlying distribution. Conversely, Stable-SPAM~\cite{28} adopts an EMA of the maximum gradient for entry-wise clipping. These data-driven techniques adapt to varying gradient scales, outperforming static clipping in privacy efficiency and empirical stability.\\
\textbf{Curvature-aware clipping.} Advancing to second-order landscape utilization, some methods address naive scalar clipping limitations in highly non-convex environments. Traditional methods uniformly scale all parameter dimensions, ignoring local landscape geometry and yielding suboptimal updates. Resolving this, Sophia~\cite{264} leverages curvature estimates to precondition the clipping operation. Incorporating second-order insights makes the clipping threshold structurally aware of the parameter space, fundamentally improving convergence stability and update efficiency in complex tasks.

The trajectory advances from static global constraints to dynamic statistical calibration and curvature informed structural clipping, reflecting the sequential adaptation of FO, dynamic FO, and approximated SO optimization. Global clipping offers straightforward implementation but suffers suboptimal performance across heterogeneous training phases. Adaptive clipping improves the utility tradeoff via real-time calibration but incurs auxiliary computational overhead. Curvature-aware clipping maximizes landscape utilization to stabilize optimization but requires computationally expensive curvature approximations. The core gap remains the absence of a unified clipping operator seamlessly integrating ZO memory efficiency with SO curvature awareness under strict DP constraints.
\subsection{Memory-Efficient Optimization}\label{me_methods}
These methods bifurcates into structural design strategies (algorithmic reformulation) and numerical compression techniques (data representation scaling). Regarding structural design, Low-memory optimizer design minimizes storage overhead through mechanisms like statelessness~\cite{204}, approximate second-moment storage~\cite{154}, and state quantization or blocking. Stateless optimization methods further eliminate historical dependencies via real-time gradient statistics~\cite{153}, parameter-driven updates~\cite{98}, and curvature estimation~\cite{158}. Additionally, memory-efficient fine-tuning supports massive architectures using stateless backpropagation and selective updates. Regarding numerical compression, methods focus on low-rank methods and optimizer state compression. Low-rank methods compress statistics into lower-dimensional subspaces, encompassing general strategies like adaptive projections and specific low-rank gradient Storage techniques utilizing tensor decomposition. Optimizer state compression reduces the variable footprint through sparse compression~\cite{88}, low-rank approximation, and state sharing~\cite{49}.

\subsubsection{Low-Memory Optimizer Design}
It refers to algorithmic techniques engineered to minimize the storage overhead inherent in training processes through mechanisms such as state quantization, approximation, and structural partitioning. These methods alleviate critical memory bottlenecks, enabling scalable and high-throughput learning by effectively reconciling limited hardware resources with the demands of maintaining model performance.\\
\textbf{Stateless optimizers.} Representing the most aggressive approach to memory efficiency, AlphaGrad~\cite{204} addresses the overhead of stateful optimizers by completely eliminating historical gradient storage. By relying on a single steepness parameter for gradient transformation, it introduces a stateless optimization mechanism that serves as a baseline for reducing memory footprints without maintaining a history window.\\
\textbf{Approximate second-moment storage.} Moving beyond the binary choice of stateful versus stateless, Adafactor~\cite{154} seeks to retain the benefits of adaptivity while mitigating costs. Unlike~\cite{204}'s removal of state, Adafactor~\cite{154} reduces the high memory overhead of second-moment storage through matrix decomposition. This approach bridges the gap between efficiency and performance by managing optimizer states through approximation rather than elimination.\\
\textbf{Quantized optimizer states.} While~\cite{154} achieves compression through decomposition, 4-bit Shampoo~\cite{172} pursues state reduction via quantization. It preserves the structure of preconditioner eigenvector matrices but quantizes them to 4-bits. FlashOptim~\cite{341} further advances memory-efficient optimization by combining ULP-based weight splitting and tailored companding functions for 8-bit quantization of momentum and variance states. This method demonstrates that high-performance state management can be maintained even with reduced numerical precision, offering an alternative compression strategy to factorization for large-scale tasks.\\
\textbf{Blocked optimizer states.} Shifting focus from numerical compression (as seen in quantization or factorization) to structural efficiency, Adam-mini~\cite{45} partitions parameters into Hessian-structure-based blocks. By assigning a single learning rate per block, it matches AdamW’s performance with significantly fewer states. This structural approach addresses memory and communication bottlenecks in large-scale training by optimizing parameter grouping rather than just compressing individual state values.\\
\textbf{Target projection strategy.} Diverging fundamentally from the previous methods that focus on managing internal optimizer states, tpSGD~\cite{231} redefines the training paradigm itself. By employing random label projections to generate local targets, it enables layer-by-layer training. This effectively eliminates the need for global backward passes, reducing memory overhead by altering the gradient propagation flow rather than the optimizer's storage mechanism.

\subsubsection{Low-Rank Algorithms}
These methods denote algorithmic strategies that compress gradients or optimizer statistics into lower-dimensional subspaces using adaptive projections and randomized approximations. These techniques substantially mitigate memory constraints and communication overheads, effectively harmonizing the trade-off between resource efficiency and approximation fidelity to support large-scale model training.\\
\textbf{Low-rank projection.} Focusing on direct dimensionality reduction, AdaRankGrad~\cite{26} mitigates the memory footprint of Adam by applying low-rank projections to the gradients. Unlike full-rank updates, it employs a randomized SVD scheme to efficiently approximate the gradient subspace. This integrates adaptive rank reduction with Adam’s learning rates, establishing a baseline for reducing memory via efficient matrix factorization.\\
\textbf{Dynamic rank adjustment.} While~\cite{26} applies projections to gradients, Adapprox~\cite{163} specifically targets the memory-intensive second moment of the optimizer. Furthermore, it distinguishes itself from static low-rank methods by introducing an adaptive rank selection mechanism. This allows the algorithm to dynamically adjust the rank of the approximation during training, striking a more flexible balance between optimization accuracy and memory overhead for large-scale models.\\
\textbf{Low-Rank\&quantization.} Moving beyond matrix factorization, hybrid approaches combine low-rank structures with data precision reduction. QuZO~\cite{29} complements low-rank adaptation with stochastic quantized perturbation to specifically reduce gradient estimation bias. Similarly, LQ-SGD~\cite{100} extends the low-rank framework of~\cite{74} by incorporating logarithmic quantization. These methods demonstrate that combining low-rank constraints with quantization can simultaneously lower communication costs and memory usage without compromising model performance.

\subsubsection{Optimizer State Compression}
It refers to a suite of methodologies that drastically reduce the memory footprint of optimization variables by employing sparsity, low-rank approximations, and state-sharing mechanisms. These strategies mitigate storage constraints, securing a viable compromise between minimized resource consumption and the preservation of training stability and convergence performance.\\
\textbf{Sparse state compression.} By sparsifying state, some methods reduce memory usage while preserving performance and stability for efficient training. SPAM~\cite{46} integrates spike-aware clipping into Adam, and uses sparse momentum to cut memory while maintaining performance. MICROADAM~\cite{47} compresses gradients before optimizer state input, retaining convergence guarantees competitive. These methods balance memory efficiency with training performance and stability via sparse states.\\
\textbf{Low-rank state approximation.} Adapprox~\cite{163} employs randomized low-rank matrix approximation for Adam's second moment, introduces adaptive rank selection to balance accuracy and memory. It addresses the high memory and computational overhead of full-rank second-moment estimation in Adam.\\
\textbf{State sharing.} By sharing state information across parameters, some methods address excessive memory overhead in adaptive optimizers. Adafactor~\cite{154} proposes that for matrix parameters, instead of storing individual states for each element, it enables parameters in the same row or column to share states. AdaAct~\cite{237} shares neuron-wise rates for same input features. State sharing methods boost optimization efficiency, stability, and adaptability via cross-parameter state reuse.

\subsubsection{Low-Rank Gradient Storage}
It alleviates the substantial memory footprint associated with high-dimensional optimization by approximating full-matrix gradients through compressed, lower-dimensional representations. These approaches utilize techniques such as subspace projection, dynamic rank adaptation, and tensor decomposition to drastically curtail state storage requirements while preserving essential information fidelity for efficient and stable model training.\\
\textbf{Gradient low-rank projection.} While both methods aim to reduce storage by projecting high-dimensional gradients onto low-dimensional subspaces, they diverge in their mathematical foundations. SUMO~\cite{34} relies on matrix decomposition, utilizing truncated SVD of per-layer gradients to form adaptive projection subspaces. In contrast, GWT~\cite{41} approaches compression through signal processing, applying wavelet transforms to compress Adam~\cite{287} states. This distinction highlights two viable paths—algebraic decomposition versus spectral transformation—for achieving significant memory reduction in optimizer states.\\
\textbf{Dynamic gradient rank.} Moving beyond static projection, these methods introduce adaptivity into the rank selection process. Adapprox~\cite{163} directly targets memory efficiency by explicitly adjusting the storage rank of the second-moment matrix via an adaptive selection mechanism. HELENE~\cite{25}, conversely, achieves a similar effect implicitly; rather than manipulating rank parameters directly, it employs layer-wise adaptive clipping and annealing to prioritize gradient information. Both strategies ultimately enhance large model optimization, but distinguish themselves by whether the dynamic ranking is an explicit architectural choice or an emergent property of gradient regularization.\\
\textbf{Gradient reconstruction.} Distinct from the projection or ranking of existing gradients, TeZO~\cite{38} addresses memory through structured estimation within a zeroth-order framework. It employs canonical polyadic decomposition (CPD) to reconstruct gradients by capturing low-rank temporal and cross-model structures. Unlike methods that compress a calculated gradient, TeZO~\cite{38} reduces training costs by dynamically adjusting the estimator's rank based on layer characteristics, optimizing memory usage through tensor decomposition rather than direct state compression.

\subsubsection{Stateless Optimization}
These methods eliminate the dependency on historical data accumulation by deriving update directives directly from instantaneous computations and intrinsic parameter attributes. These strategies leverage immediate feedback mechanisms, ranging from real-time gradient statistics to lightweight curvature estimates, to drastically minimize memory overhead and computational latency, thereby ensuring efficient convergence and adaptability without the burden of maintaining persistent optimizer states.\\
\textbf{Real-time gradient statistics.} SWAN~\cite{49} integrates core preprocessing, efficiency tweaks, and variants for flexible trade-offs, all relying on instantaneous stats. It addresses the overhead and bias of exponential moving averages in adaptive optimizers by using instantaneous gradient statistics for stateless updates.\\
\textbf{Parameter characteristic-driven updates.} Some methods focus on adapting updates using parameter-specific properties. AlphaGrad~\cite{204} uses a single steepness parameter for memory-efficient scaling. SGD-SaI~\cite{13} adjusts initial learning rates through gradient signal-to-noise ratio. BFE~\cite{191} tunes learning rates through forward loss information; AdaBFE~\cite{191} extends it to per-parameter adaptation for high-dimensional spaces. These methods enhance optimization by parameter-specific adaptations, balancing efficiency, performance, and adaptability across tasks.\\
\textbf{Real-time curvature estimation.} Sophia~\cite{264} uses a lightweight diagonal Hessian estimate as a preconditioner, eliminating the dependency of traditional second-order optimizers on extensive storage via lightweight real-time curvature estimation. Such method efficiently use real-time curvature information to boost convergence without extra computational burden.

\subsubsection{Memory-Efficient for Large Models}
It circumvents the prohibitive resource demands associated with full-parameter training by innovating gradient computation and application protocols. These strategies deploy techniques ranging from stateless backpropagation and zeroth-order kernel estimation to sparse parameter selection, effectively minimizing memory footprints to enable the feasible optimization of massive architectures on constrained hardware while maintaining computational effectiveness.\\
\textbf{ZO fine-tuning.} KerZOO~\cite{35} is a kernel-function-based ZO framework using kernel smoothing to mitigate lower-order bias, enhancing convergence speed. This method boosts ZO fine-tuning efficiency for LLMs through kernel-based gradient estimation improvement.\\
\textbf{Staless fine-tuning.} LOMO~\cite{17} reduces memory usage by computing the gradient for each parameter, immediately updating the parameter, clearing the gradient afterward, and retaining the gradient of at most one parameter in memory. It enables staleless fine-tuning by fusing backpropagation with parameter update.\\
\textbf{Selective parameter fine-tuning.} By dynamically selecting subsets of parameters to perturb and update, some methods address the prohibitive computational and memory overhead of full-parameter fine-tuning for large models. Sparse-MeZO~\cite{23} dynamically selects a sparse parameter subset through layer-wise magnitude thresholds. LeZO~\cite{50} integrates SPSA with layer-wise sparsity, selectively updating layers to achieve efficient fine-tuning without additional memory. These methods enable cost-effective large model fine-tuning by dynamic parameter selection, balancing efficiency and performance.

\subsection{Tailored Optimization Approaches}\label{to_methods}
These methods are categorized into automated design paradigms and robustness enhancement mechanisms, distinguishing between rule discovery and engineered resilience. Auto-designed optimizers transcend human heuristics by automating rule formulation. These include programmatic search within structured spaces~\cite{287}~\cite{288}, neural controllers modeling physical transport, and symbolic derivation. Additionally, techniques leverage evolutionary strategies to escape local optima or meta-learning for dynamic, task-specific adaptation. Robust optimization mitigates training vulnerabilities through gradient resilience and systemic adaptability. Gradient resilience ensures stability via noise-robust gradients with adaptive clipping~\cite{93} and hardware noise countermeasures~\cite{204}. Systemic adaptability addresses broader constraints, encompassing distribution shift robustness for federated learning and structure-aware optimization for large scale feasibility, collectively ensuring reliable convergence under volatile conditions.

\subsubsection{Auto-Designed Optimizers}
These optimizers transcend the rigid boundaries of manually engineered heuristics by automating the discovery and formulation of update rules through algorithmic exploration. These systems harness diverse generative paradigms, ranging from symbolic derivation and evolutionary search to meta-learning and programmatic synthesis, to uncover novel optimization mechanisms that significantly enhance adaptability, generalization, and convergence capabilities beyond the reach of human intuition.\\
\textbf{Programmatic search optimizers.} Some methods address the limitation of hand engineered optimizers by automating rule discovery through structured program spaces, exploring beyond human heuristics to find efficient update mechanisms. The method in~\cite{166} replaces hand-engineered optimization rules through programmatic policy representation and automated search mechanisms, which carries significant heuristic value in the field of optimizer design. The method in~\cite{167} frames algorithm discovery as a program search in infinite sparse spaces, uses evolutionary techniques to explore efficiently, and introduces Lion~\cite{167}, an optimizer with sign momentum updates that reduce memory usage. The method in~\cite{165} employs a controller trained with RL to discover update rules such as PowerSign and AddSign. These methods share programmatic representation of rules but differ in search paradigms, uniting to automate optimizer discovery.\\
\textbf{Neural controller optimizers.} KO~\cite{205} models updates as stochastic collisions via Boltzmann transport equation to boost parameter diversity and generalization. It achieves precise regulation of parameter dispersion through hard and soft collision mechanisms, thus providing a physically inspired, theoretically provable, and engineering-efficient new paradigm for neural controller optimizers.\\
\textbf{Symbolic derivation optimizers.} AGS-GD~\cite{236} reconstructs the gradient update rule through the symbolic mathematical derivation of Gaussian smoothing, converting local gradients into non-local gradients and addressing the pain point of traditional optimizers getting trapped in suboptimal local minima at the theoretical root. This method addresses the pain point of traditional optimizers getting trapped in suboptimal local minima at the theoretical root by means of symbolic reasoning.\\
\textbf{Evolutionary algorithm optimizers.} Some methods address the local optima limitation of traditional gradient-based optimizers by integrating evolutionary strategies. GADAM~\cite{182} combines Adam~\cite{287} with genetic algorithms using crossover and mutation to escape local optima. BGADAM~\cite{187} extends this framework by adding a boosting strategy to generate diverse training sets. These optimizers merge evolutionary and gradient-based approaches to improve exploration and avoid local optima.\\
\textbf{Meta-learning optimizers.} They address limitations of fixed-update optimizers by learning specific task's update rules through meta-learning, enabling better adaptation to diverse objectives. WarpAdam~\cite{195} inserts a learnable distortion matrix into Adam’s pipeline. HyperAdam~\cite{222} uses an RNN~\cite{339} framework to adapt decay rates and weights, combining ensemble updates from varying Adam~\cite{287} configurations. MADA~\cite{22} unifies optimizers via hyper-gradient descent. These meta-learned optimizers dynamically tune update rules via meta-learning, enhancing task adaptability and outperforming fixed heuristics in diverse scenarios.

\subsubsection{Robust Optimization}
These methods mitigate the vulnerability of training procedures to systemic irregularities, ranging from stochastic gradient noise and hardware quantization errors to data distribution shifts. By incorporating resilience enhancing mechanisms such as adaptive clipping, drift correction, and structural decomposition, these strategies counteract destabilizing factors to ensure reliable convergence and consistent model performance amidst volatile or heterogeneous conditions.\\
\textbf{Noise-robust gradients.} Some methods focus on improving the noise-robust of optimizer through different techniques. AdaGC~\cite{30} uses exponential moving averages for per-parameter local clipping thresholds. AdaNorm~\cite{194} corrects norms by EMA of past gradients, resolving low and atypical gradient issues. These methods enhance robustness of the gradient, improving the stability of optimizer across tasks.\\
\textbf{Distribution shift robustness.} Some methods address client drift and local global objective misalignment under heterogeneous data distributions in federated learning, where vanilla methods suffer from performance degradation or high communication costs. FedCET~\cite{118} uses learning rates to weight client information. FAdamGC~\cite{119} integrates gradient correction into adaptive federated optimization. These methods enhance federated optimizers' robustness to distribution shift and reduce communication overhead through tailored strategies.\\
\textbf{Hardware noise robustness.} QuZO~\cite{29} reduces gradient bias via stochastic quantized perturbation. It enhances hardware noise robustness, enabling stable optimization in constrained setups.\\
\textbf{Structure-aware optimization.} BC-ADMM~\cite{244} decomposes large scale problems into parallelizable subproblems allowing larger steps and faster convergence, maintains solution feasibility throughout the process. Such structure-aware methods enhance optimization efficiency and feasibility for large-scale tasks by tailored decomposition and relaxation strategies.

\section{Experiments} \label{section:exp}
Although~\cref{section:method} details the evolution of optimizers for constrained environments, such as distributed or differentially private training, fair empirical comparisons are often hindered by hardware and engineering variations. We therefore present a controlled evaluation protocol designed to isolate fundamental algorithmic behavior from implementation details. Rather than benchmarking peak system throughput, we focus on evaluating the mathematical mechanisms underlying modern optimizers. We analyze hyperparameter sensitivity, convergence dynamics, and cross-architecture generalization to determine whether methods that perform well on vision tasks can transfer reliably to language modeling. The remainder of this section details our datasets, evaluation metrics, quantitative results, and qualitative analysis.
\subsection{Datasets}
To comprehensively evaluate the optimizers, we selected widely adopted deep learning benchmarks spanning image classification and language modeling.\\
\textbf{Vision tasks.} We used the standard ImageNet-1K~\cite{298} dataset, which comprises approximately 1.28M training and 50,000 validation images across 1,000 classes. We selected ResNet-50~\cite{297} and ViT-Small (ViT-S)~\cite{296} to represent CNN~\cite{337} and vision transformer architectures, respectively.\\
\textbf{Language tasks.} We trained a 60M-parameter Llama~\cite{330} model from scratch on the WikiText-103~\cite{329} dataset. We emphasize that this 60M configuration is not designed to replicate emergent behaviors at the scale of full-sized LLMs. Instead, it serves as a highly representative architectural proxy for modern autoregressive Transformers, retaining the core architectural features of Llama~\cite{330} while remaining computationally tractable. This setup enables us to create a rigorous testbed for assessing the cross-architecture transferability of optimizers, addressing the key question: Do optimizers heavily tuned for CNN~\cite{337} or ViT~\cite{296} perform catastrophically poorly when faced with the highly anisotropic loss landscapes characteristic of causal attention mechanisms?
\subsection{Metrics and Settings}
To rigorously evaluate the performance of the diverse optimizers, we apply a rich set of evaluation metrics tailored to the nature of each benchmark task.\\
\textbf{Hyperparameter setting.} For a fair comparison, we focus exclusively on optimizing one common hyperparameter: the learning rate. Starting with default values suggested in the original literature for each optimizer, we employ a grid search strategy that explores the vicinity of these baselines by scaling them with factors of $0.1$, $0.2$, $1.0$, $5.0$, and $10.0$. The optimal learning rates identified through this process are subsequently transferred to the ResNet~\cite{297} and Llama~\cite{330} models.\\
\textbf{Vision tasks.} To assess the performance of different optimizers on image classification, we use Top-1 accuracy as the primary metric. We utilize the DeiT~\cite{299} training setting for ViT-S~\cite{296}. For ResNet-50~\cite{297}, we evaluate models for 100 and 300 epochs, employing the A3 and A2~\cite{300} training settings, respectively. Specifically, we consider three regular training settings for ImageNet-1K~\cite{295} classification experiments:\\
\textbf{\textit{1) DeiT setting~\cite{298}:}} Designed for Transformer and modern CNN architectures like ViT-S~\cite{296}, this setting uses several advanced augmentations, including RandAugment~\cite{302}, Mixup~\cite{303}, CutMix~\cite{304}, and Random Erasing~\cite{305}.\\
\textbf{\textit{2) A2/A3 settings~\cite{300}:}} Designed for CNNs~\cite{337} to match the performance and convergence speeds of ViTs~\cite{296}, these settings reduce augmentation strengths and replace the cross-entropy (CE) loss with binary cross-entropy (BCE) loss compared to the DeiT setting.\\
Ingredients and hyperparameters used for image classification training settings are detailed in~\cref{table:training_setting}. We note that certain optimizers~\cite{307,32,149,154,196,306,223,225,171} and SGDP~\cite{229} exhibit unstable or failed convergence on the ResNet~\cite{297} architecture when trained with BCE loss; for these cases, we used CE loss instead.\\
\begin{table}[t!]
    \centering
    \caption{Ingredients and hyper-parameters used for image classification training settings. Taking ImageNet-1K as the template setups, the settings of A2/A3~\cite{299} take ResNet-50~\cite{297} as the examples, the DeiT~\cite{299} setting takes ViT-S~\cite{296} as the example. Gray regions should be modified for each optimizer.}
    \label{table:training_setting}
    \renewcommand{\arraystretch}{1.1} 
    \setlength\tabcolsep{10.0pt} 
    \resizebox{1.0\linewidth}{!}{
        \begin{tabular}{l | c | c | c}
        \toprule
        Procedure & DeiT & A2 & A3 \\
        Dataset & IN-1K & IN-1K & IN-1K \\
        \hline
        Epochs & 300 & 300 & 100 \\
        Batch size & 1024 & 1024 & 1024 \\
        
        \rowcolor{gray!15} Optimizer & AdamW & AdamW & AdamW \\
        \rowcolor{gray!15} Learning rate & $1 \times 10^{-3}$ & $1 \times 10^{-3}$ & $1 \times 10^{-3}$ \\
        \rowcolor{gray!15} Optimizer Momentum & 0.9, 0.999 & 0.9, 0.999 & 0.9, 0.999 \\
        \rowcolor{gray!15} Weight decay & 0.05 & 0.05 & 0.05 \\

        LR decay & Cosine & Cosine & Cosine \\
        Warmup epochs & 5 & 5 & 5 \\
        \hline
        Label smoothing $\epsilon$ & 0.1 & \textcolor{gray}{$\times$} & \textcolor{gray}{$\times$} \\
        Dropout & \textcolor{gray}{$\times$} & \textcolor{gray}{$\times$} & \textcolor{gray}{$\times$} \\
        Stochastic Depth & 0.1 & 0.05 & \textcolor{gray}{$\times$} \\
        Repeated Augmentation & \checkmark & \checkmark & \textcolor{gray}{$\times$} \\
        Gradient Clip. & \textcolor{gray}{$\times$} & \textcolor{gray}{$\times$} & \textcolor{gray}{$\times$} \\
        \hline
        Horizontal flip & \checkmark & \checkmark & \checkmark \\
        RandomResizedCrop & \checkmark & \checkmark & \checkmark \\
        Rand Augment & 9/0.5 & 7/0.5 & 6/0.5 \\
        Auto Augment & \textcolor{gray}{$\times$} & \textcolor{gray}{$\times$} & \textcolor{gray}{$\times$} \\
        Mixup $\alpha$ & 0.8 & 0.1 & 0.1 \\
        Cutmix $\alpha$ & 1.0 & 1.0 & 1.0 \\
        Erasing probability & 0.25 & \textcolor{gray}{$\times$} & \textcolor{gray}{$\times$} \\
        ColorJitter & \textcolor{gray}{$\times$} & \textcolor{gray}{$\times$} & \textcolor{gray}{$\times$} \\
        EMA & \textcolor{gray}{$\times$} & \textcolor{gray}{$\times$} & \textcolor{gray}{$\times$} \\
        \hline
        CE loss & \checkmark & \textcolor{gray}{$\times$} & \textcolor{gray}{$\times$} \\
        BCE loss & \textcolor{gray}{$\times$} & \checkmark & \checkmark \\
        \bottomrule
        \end{tabular}
    }
    \vspace{-1.0em}
\end{table}
\textbf{Language tasks.} We use perplexity (PPL) to evaluate the performance of different optimizers on language modeling tasks. Our training configuration aligns with that of~\cite{121}, utilizing a batch size of 256 and a sequence length of 512. However, due to computational constraints, we employ a 60M-parameter Llama-like architecture, deviating from the model size used in the reference. Furthermore, the total number of training tokens is determined following the Chinchilla scaling laws~\cite{71}. Specifically, we consider a standard training setting for the Llama~\cite{330} architecture (60M parameters) on the WikiText-103~\cite{329} dataset. This training scheme includes specific data configurations, optimization setups, and regularization tricks, using AdamW~\cite{148} as an example: with a peak learning rate of $1 \times 10^{-3}$, a weight decay of $0.5$, and an optimizer $\epsilon$ of $10^{-8}$. We use a cosine learning rate scheduler with a warmup phase of $2000$ steps. To ensure training stability and generalization, gradient clipping is applied with a threshold of $0.5$. The training is conducted on $8$ GPUs with a sequence length of $512$ and a per-device batch size of $32$ (resulting in a global batch size of $256$). Ingredients and hyperparameters used for this language modeling setting are strictly controlled to benchmark optimization performance.
\subsection{Analysis}
\begin{figure*}[t!] 
\centering
\includegraphics[width=1.0\linewidth]{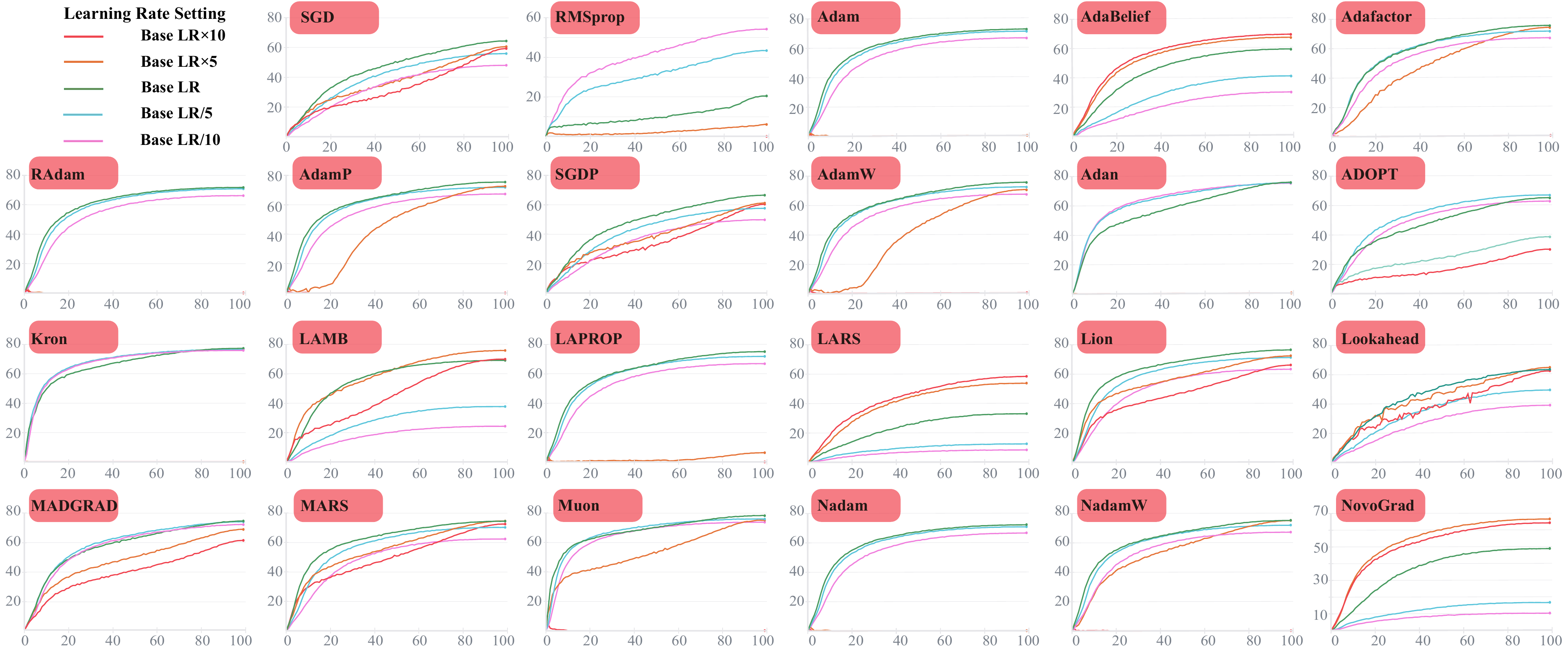}
    \vspace{-1.5em} 
    \caption{\textbf{Comparison of Top-1 accuracy and learning rate sensitivity analysis of various optimizers.} Each subfigure demonstrates the robustness and sensitivity of individual optimizers across a range of scaled learning rates (from 0.1× to 10× the base learning rate). Figures demonstrate that convergence stability and final Top-1 accuracy heavily depend on precise hyperparameter tuning.}
    \label{fig:figure1_1}
    \vspace{-1.5em} 
\end{figure*}

\begin{figure*}[t!] 
    \centering
    \includegraphics[width=1.0\linewidth]{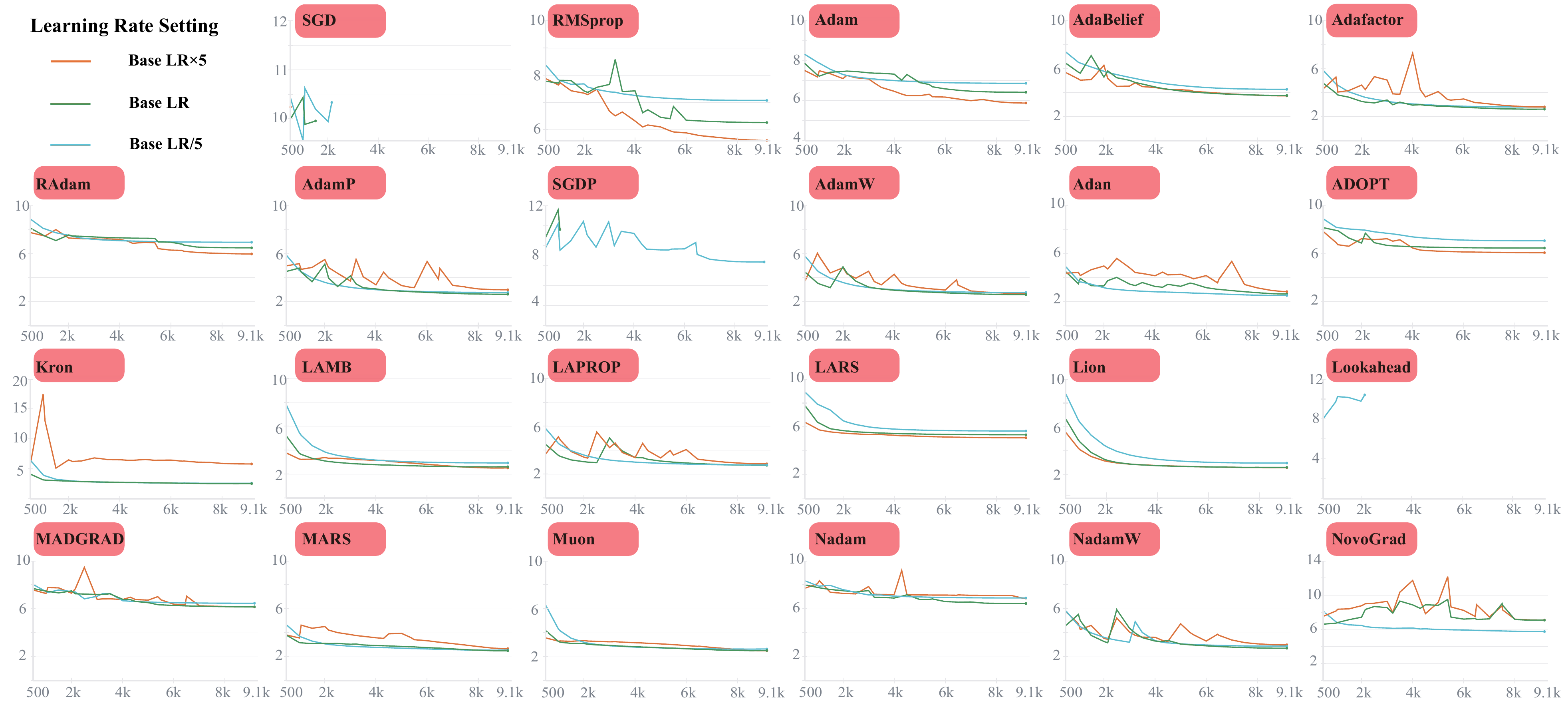} 
   
    \caption{\textbf{Optimizers' performance on Llama-60m.} Subfigures show validation loss curves of various optimizers across three learning rate settings. The robustness to learning rate variations differs significantly among methods, with some optimizers (e.g., Muon~\cite{173}, MARS~\cite{181}) maintaining stable convergence across all scales while others (e.g., SGD~\cite{307}, SGDP~\cite{229}) rapidly diverge. Missing data indicates gradient explosion (loss becomes NaN) at the initial step or the point of discontinuation.}
    
    \label{fig:figure1_2}
\end{figure*}

We systematically analyze the practical performance of various optimizers on diverse tasks and investigate their underlying mechanisms. These dynamics are visually corroborated by the training curves on the ViT-S~\cite{296}, ResNet-50~\cite{297}, and Llama-60m~\cite{330} models (as detailed in~\cref{fig:figure1_1,fig:figure1_2,fig:figure_vrcontrast}).
\subsubsection{Learning Rate Sensitivity on ViT-S}
To systematically assess learning rate sensitivity, we report the Top-1 accuracy across a broad spectrum of varying learning rates, explicitly highlighting the identified optimal values that yield peak performance (see~\cref{table:lr_sensitivity,fig:figure1_1}).\\
\begin{table}[t!]

    \centering

    \caption{Comparison of various optimizers' hyperparameter robustness and learning rate settings.}

    \label{table:lr_sensitivity}

    \renewcommand{\arraystretch}{1.2}

    \setlength\tabcolsep{2.0pt}

    \resizebox{1.\linewidth}{!}{

        \begin{tabular}{l p{1.8cm}<{\centering} p{1.8cm}<{\centering} p{1.8cm}<{\centering} p{1.8cm}<{\centering} p{1.8cm}<{\centering} c}

        \toprule

        \textbf{Optimizer} & 

        \textbf{0.1$\times$LR} & 

        \textbf{0.2$\times$LR} & 

        \textbf{1.0$\times$LR} & 

        \textbf{5.0$\times$LR} & 

        \textbf{10.0$\times$LR} & 

        \textbf{Best LR} \\

        \midrule

        \textbf{SGD}~\cite{307} & 

        48.056 & 

        55.918 & 

        \textbf{64.360} & 

        60.590 &

        59.262 & 

        0.1 \\

        \textbf{RMSprop}~\cite{325} & 

        \textbf{54.254} & 

        43.416 & 

        20.512 & 

        6.1140 & 

        - & 

        1e-4 \\

        \textbf{Adam}~\cite{287} & 

        66.234 & 

        70.772 & 

        \textbf{72.240} & 

        - & 

        - & 

        1e-3 \\

        \textbf{AdaBelief}~\cite{149} & 

        29.248 & 

        40.210 & 

        58.562 &  

        66.528 &

        \textbf{68.642} & 

        1e-2 \\

        \textbf{Adafactor}~\cite{154} & 

        66.294 & 

        70.834 & 

        \textbf{74.630} & 

        73.344 & 

        - & 

        1e-3 \\

        \textbf{RAdam}~\cite{160} & 

        66.220 & 

        70.826 & 

        \textbf{71.190} &  

        - &

        - & 

        1e-3 \\

        \textbf{AdamP}~\cite{229} & 

        67.336 & 

        71.856 & 

        \textbf{75.444} & 

        72.606 &

        - & 

        1e-3 \\

        \textbf{SGDP}~\cite{229} & 

        49.800 & 

        57.662 & 

        \textbf{66.536} & 

        61.182 & 

        60.416 & 

        0.1 \\

        \textbf{AdamW}~\cite{148} & 

        67.082 & 

        72.076 & 

        \textbf{75.196} & 

        70.136 & 

        0.2640 & 

        1e-3 \\

        \textbf{Adan}~\cite{168} & 

        74.562 & 

        74.956 & 

        \textbf{75.190} & 

        - & 

        - & 

        1.25e-2 \\

        \textbf{ADOPT}~\cite{196} & 

        62.408 & 

        \textbf{66.552} & 

        64.684 & 

        38.170 & 

        29.670 &

        2e-4 \\

        \textbf{Kron}~\cite{306} & 

        75.878 & 

        76.432 & 

        \textbf{77.304} & 

        - & 

        - & 

        1e-3 \\

        \textbf{LAMB}~\cite{227} & 

        24.346 & 

        37.750 & 

        69.100 & 

        \textbf{75.858} & 

        69.956 & 

        5e-3 \\

        \textbf{LAPROP}~\cite{156} & 

        66.834 & 

        71.834 & 

        \textbf{75.050} & 

        6.3640 & 

        - & 

        1e-3 \\

        \textbf{LARS}~\cite{223} & 

        8.3020 & 

        12.426 & 

        32.892 & 

        53.660 &

        \textbf{58.276} & 

        1 \\

        \textbf{Lion}~\cite{167} & 

        63.192 & 

        71.092 & 

        \textbf{76.334} & 

        72.254 & 

        65.988 & 

        1e-4 \\

        \textbf{Lookahead}~\cite{225} & 
        38.492 & 
        48.896 & 
        62.898 & 
        \textbf{64.320} &
        61.962 & 
        0.5 \\

        \textbf{MADGRAD}~\cite{171} & 
        72.218 & 
        74.150 & 
        \textbf{74.634} & 
        68.946 & 
        61.526 & 
        1e-3 \\

        \textbf{MARS}~\cite{181} & 
        62.414 & 
        70.300 & 
        74.474 & 
        \textbf{74.608} & 
        72.542 & 
        5e-3 \\

        \textbf{Muon}~\cite{173} & 
        73.844 & 
        75.944 & 
        \textbf{78.252} & 
        75.206 & 
        - & 
        2e-3 \\
        \textbf{Nadam}~\cite{155} & 
        66.594 & 
        70.742 & 
        \textbf{72.180} & 
        - & 
        - &
        1e-3 \\

        \textbf{NadamW}~\cite{326} & 
        67.332 & 
        71.872 & 
        \textbf{75.278} & 
        74.982 & 
        - &
        1e-3 \\
        \textbf{NovoGrad}~\cite{157} & 
        10.290 & 
        16.686 & 
        48.892 & 
        \textbf{66.514} & 
        64.258 &
        5e-3 \\
        \bottomrule
        \end{tabular}
    }
\end{table}
\textbf{High-performance robustness.}
Kron~\cite{306} and Adan~\cite{168} show minimal variance at lower scales ($0.1\times$ to $1.0\times$) but are prone to divergence at high learning rates (LR). This occurs because Adan~\cite{168} reformulates Nesterov~\cite{324} momentum and integrates adaptive gradient mechanisms, accelerating convergence via a targeted look-ahead momentum strategy. However, the coupling of momentum terms with adaptive gradients creates high update inertia; when the base LR is scaled up, large gradient magnitudes combined with strong momentum inertia cause drastic oscillations of parameter updates on the loss landscape, negating the stabilizing effect of the adaptive denominator. Muon~\cite{173} stands out as the most hyperparameter-insensitive, maintaining $>75\%$ accuracy from $0.2\times$ to $5.0\times$, indicating superior ease of tuning. It constrains parameter updates to a limited manifold through iterative orthogonalization of the accumulated momentum matrix, regardless of how the external learning rate is scaled. This intrinsic regularization makes it highly insensitive to hyperparameter variations.\\
\textbf{High lower-bound stability.}
Despite notable degradation outside optimal settings, Lion~\cite{167}, MADGRAD~\cite{171}, and MARS~\cite{181} maintain a "safe" baseline ($>60\%$ accuracy) across the full $0.1\times$ to $10\times$ range. The core of Lion~\cite{167} is its parameter update mechanism, which uses only the sign of the momentum-weighted gradient ($\text{sign}(c_t)$) and discards the exact magnitude. This means that regardless of whether gradients explode or vanish due to learning rate scaling, the sign update term remains bounded to $\{-1, 0, +1\}$. This ensures the core step size of parameter updates is determined solely by the LR and is identical across all dimensions for the sign component. This mechanism inherently protects against numerical instability caused by anomalous gradient magnitudes, resulting in a remarkably high training floor. MADGRAD~\cite{171} combines an averaging-based update mechanism with adaptive scaling properties, centered around a dual averaging framework. Instead of applying adaptive scaling to gradients step by step, it smoothly accumulates historical gradients with weights, followed by normalization via a cube-rooted adaptive denominator. This mechanism naturally buffers against extreme gradient shocks.\\
\textbf{Large learning rate sensitivity (Adam family).}
These methods are highly vulnerable to aggressive LRs. Scaling by $5\times$ or $10\times$ typically leads to non-convergence or sharp performance declines.\\
\textbf{Specific preferences.}
RMSprop~\cite{325} exhibits limited robustness and prefers conservative LRs. In contrast, AdaBelief~\cite{149}, LAMB~\cite{227}, LARS~\cite{223}, and NovoGrad~\cite{157} face convergence difficulties at small LRs, necessitating larger learning rates. The core of LARS~\cite{223} and LAMB~\cite{227} is a trust ratio: they normalize the update step size for each layer by computing the ratio of the parameter norm to the update vector norm, ensuring stable update magnitudes across layers. This mechanism has a critical weakness: if the global base learning rate is set excessively small, multiplying it by the trust ratio will excessively compress the actual parameter update step size. This prevents the model from escaping initial flat regions in the early training stage. Consequently, these optimizers require larger learning rates to drive normalized updates and produce sufficient parameter movement.\\
\textbf{SGD resilience.}
While SGD~\cite{307} lags in peak accuracy ($64.36\%$), it demonstrates exceptional robustness to large learning rates, remaining usable even at $10\times$ scaling where other optimizers fail. The update formula for adaptive optimizers like Adam~\cite{287} involves scaling the first moment of gradients by the square root of the second moment of gradients plus a small epsilon value, all multiplied by the negative learning rate. For parameters with extremely small gradients, the square root is also minuscule, and the division yields an enormous multiplier. When the base learning rate is further increased by $10\times$, the actual step size for these parameters becomes astronomically large, instantly corrupting the model weights. In contrast, the update rule for SGD~\cite{307} is straightforward: it simply multiplies the negative learning rate by the gradient directly. Even if the LR is scaled up $10\times$, the update step size remains linearly proportional to the gradient. While a large LR causes SGD~\cite{307} to oscillate violently around the optimal point, it never triggers numerical explosion due to division by a tiny value.
\begin{table}[t!]
    \centering
    \caption{Results on ViT-S~\cite{296}, ResNet-50~\cite{297}, and Llama-60m~\cite{330} models across various optimizers. We compare different learning rate multipliers on Llama-60m~\cite{330} and show that Muon~\cite{173} consistently achieves strong performance across all architectures.}
    \label{table:optim_generalization}
    
    \renewcommand{\arraystretch}{1.2}
    \setlength\tabcolsep{2.0pt}
    
    \resizebox{1.\linewidth}{!}{
        \begin{tabular}{l p{1.8cm}<{\centering} p{1.8cm}<{\centering} p{1.8cm}<{\raggedleft\arraybackslash} p{2.2cm}<{\raggedleft\arraybackslash} p{2.0cm}<{\raggedleft\arraybackslash}}
        \toprule
        \multirow{2}{*}{\textbf{Optimizer}} & 
        \textbf{ViT-S} & 
        \textbf{ResNet-50} & 
        \multicolumn{3}{c}{\textbf{Llama-60m}} \\
        
        \cmidrule(lr){4-6}
        & 
        Acc (\%) & 
        Acc (\%) & 
        \multicolumn{1}{c}{\textbf{Base LR}} & 
        \multicolumn{1}{c}{\textbf{5.0$\times$LR}} & 
        \multicolumn{1}{c}{\textbf{0.2$\times$LR}} \\
        
        \midrule
        
        \textbf{SGD}~\cite{307} & 
        64.360 & 
        75.926 & 
        - & 
        - & 
        - \\

        \textbf{RMSprop}~\cite{325} & 
        54.254 & 
        71.536 & 
        522.57 & 
        269.27 & 
        1178.34 \\

        \textbf{Adam}~\cite{287} & 
        72.240 & 
        74.480 & 
        614.93 & 
        357.95 & 
        964.46 \\

        \textbf{AdaBelief}~\cite{149} & 
        68.642 & 
        74.232 & 
        43.27 & 
        41.89 & 
        72.43 \\
        
        \textbf{Adafactor}~\cite{154} & 
        74.630 & 
        76.808 & 
        13.78 & 
        16.55 & 
        16.48 \\

        \textbf{RAdam}~\cite{160} & 
        71.190 & 
        74.612 & 
        667.83 & 
        399.59 & 
        1063.62 \\

        \textbf{AdamP}~\cite{229} & 
        75.444 & 
        76.818 & 
        13.79 & 
        20.01 & 
        15.97 \\

        \textbf{SGDP}~\cite{229} & 
        66.536 & 
        74.066 & 
        - & 
        - & 
        592.58 \\

        \textbf{AdamW}~\cite{148} & 
        75.196 & 
        76.314 & 
        13.42 & 
        14.70 & 
        15.91 \\

        \textbf{Adan}~\cite{168} & 
        75.190 & 
        76.770 & 
        14.11 & 
        17.14 & 
        12.41 \\

        \textbf{ADOPT}~\cite{196} & 
        66.552 & 
        74.584 & 
        655.08 & 
        44.22 &  
        1208.53 \\

        \textbf{Kron}~\cite{306} & 
        77.304 & 
        75.226 & 
        12.58 & 
        336.21 & 
        13.39 \\

        \textbf{LAMB}~\cite{227} & 
        75.858 & 
        77.590 & 
        13.56 & 
        12.28 & 
        18.72 \\

        \textbf{LAPROP}~\cite{156} & 
        75.050 & 
        76.146 & 
        15.32 & 
        17.37 & 
        15.86 \\ 

        \textbf{LARS}~\cite{223} & 
        58.276 & 
        76.196 & 
        210.44 & 
        289.03 & 
        164.25 \\

        \textbf{Lion}~\cite{167} & 
        76.334 & 
        75.582 & 
        13.80 & 
        19.72 & 
        13.52 \\

        \textbf{Lookahead}~\cite{225} & 
        64.320 & 
        76.872 & 
        - & 
        - & 
        - \\

        \textbf{MADGRAD}~\cite{171} & 
        74.634 & 
        76.940 & 
        473.88 & 
        473.79 & 
        642.97 \\

        \textbf{MARS}~\cite{181} & 
        74.608 & 
        77.294 & 
        12.12 & 
        14.46& 
        12.75 \\

        \textbf{Muon}~\cite{173} & 
        78.252 & 
        77.996 & 
        12.40 & 
        12.21 & 
        14.01 \\

        \textbf{Nadam}~\cite{155} & 
        72.180 & 
        74.732 & 
        627.00 & 
        988.49 & 
        999.46 \\

        \textbf{NadamW}~\cite{326} & 
        75.278 & 
        76.514 & 
        14.95 & 
        19.96 & 
        17.75 \\

        \textbf{NovoGrad}~\cite{157} & 
        66.514 & 
        71.694 & 
        1176.20 & 
        1192.70 & 
        309.63 \\
        
        \bottomrule
        \end{tabular}
    }
\end{table}%
\textbf{Efficacy of decoupled weight decay modifiers} The integration of decoupled weight decay consistently yields measurable performance improvements over foundational algorithms. At the 1.0$\times$ learning rate, AdamW~\cite{148} achieves 75.196, outperforming standard Adam~\cite{287} by a notable margin. A parallel improvement is evident when comparing NadamW~\cite{326}, which reaches 75.278, to its unmodified counterpart Nadam~\cite{155}, which achieves 72.18. Furthermore, AdamP~\cite{229} also demonstrates superior performance over standard Adam~\cite{287} by reaching 75.444, highlighting the general effectiveness of refined parameter update rules.
\subsubsection{Cross-Architecture Generalization}\label{generalization}
To evaluate cross-architecture generalization, we introduce a parameter transfer test and report the resulting Top-1 accuracy and PPL in~\cref{table:optim_generalization}. Specifically, the optimal learning rates identified for ViT-S~\cite{296} are directly applied to ResNet-50~\cite{297} and Llama-60M~\cite{330}. Under this setting, the core learning rate remains strictly fixed while auxiliary training strategies and weight decay are adapted to architecture-specific best practices. Furthermore, we conduct learning rate scaling experiments on Llama-60M~\cite{330} to comprehensively assess the hyperparameter robustness of different optimizers on language model architectures.\\
\textbf{Superior robustness.}
Muon~\cite{173} and MARS~\cite{181} achieve excellent results across ResNet~\cite{297} and Llama~\cite{330}. They are particularly effective in maintaining low and stable PPL (PPL $\approx 12$-$14$) on Llama~\cite{330}, even under extreme learning rate scaling. Muon~\cite{173}'s cross-architecture generalization stems from matrix-level orthogonalization in its algorithmic design. Unlike methods relying on scalar variance histories, it projects update matrices onto restricted manifolds. This intrinsic geometric regularization effectively flattens exaggerated step sizes in the highly anisotropic loss landscape of large language models. MARS~\cite{181}, on the other hand, introduces a scaling parameter to control the strength of variance reduction and subtracts the gradient of the previous parameters on the current batch from the current step's gradient. This gradient correction mathematically cancels the inherent stochastic noise introduced by diverse network architectures and complex data distributions.\\
\textbf{High transferability with constraints.}
While showing strong general performance, Adafactor~\cite{154}, AdamP~\cite{229}, AdamW \cite{148}, Adan~\cite{168}, LAMB~\cite{227}, LAPROP~\cite{156}, Lion~\cite{167}, NadamW~\cite{326}, and Kron~\cite{306} slightly trail Muon~\cite{173} and MARS~\cite{181}. Kron~\cite{306} serves as a prime example: despite achieving a top-tier PPL of $12.57$ on Llama at the base LR (demonstrating a high ceiling), its performance collapses at $5\times$ LR (PPL $336.2$). This reveals a significantly narrower hyperparameter "safety margin" compared to robust alternatives. One core mechanism enabling optimizers like AdamW~\cite{148} and NadamW~\cite{326} to remain competitive across diverse architectures is their successful decoupling of adaptive gradient updates from weight decay. When transitioning from shallow vision models to large language models with drastically different parameter scales, traditional regularization is easily distorted by the uneven accumulation of historical second moments. By placing the penalty term outside the scaling denominator, these methods ensure consistent and faithful regularization regardless of the absolute magnitudes of the parameters in each layer. LAMB~\cite{227} adopts layer-wise adaptation to rigorously constrain step sizes. This mathematically decouples the magnitude of stochastic gradients from the applied updates across networks of varying depths. Similarly, Lion uses sign-based updates to maintain uniform magnitudes for all parameter updates, rendering the optimizer insensitive to gradient spikes common in complex attention mechanisms. Methods such as Adan~\cite{168} and AdamP~\cite{229} significantly improve cross-architecture generalization by introducing sophisticated momentum estimation and trajectory correction.\\
\textbf{Architecture-specific instability.}
A large group of optimizers including RMSprop~\cite{325}, Adam~\cite{287}, MADGRAD~\cite{171}, and Adabelief~\cite{149} performs acceptably on vision backbones but struggles to converge on Llama~\cite{330}. The core mechanism behind this generalization failure lies in gradient outliers common in large language model training. Relying on highly biased scalar variance histories causes adaptive methods like Adam~\cite{287} to suffer from second-moment contamination by extreme gradients, leading the optimizer to incorrectly scale per-dimension step sizes during parameter updates and deviate from valid optimization trajectories in high-dimensional nonconvex spaces.\\
\textbf{Training collapse.} While SGD series optimizers are viable in vision architectures, they face catastrophic training collapse (as shown in~\cref{fig:figure1_2}) when applied to the Llama~\cite{330} architecture. The extreme depth and complex attention mechanisms of large language models result in massive disparities in gradient magnitudes and parameter scales across different layers. Lacking adaptive parameter-wise or layer-wise scaling mechanisms, pure first-order momentum methods fail to simultaneously satisfy the convergence requirements of all layers. Consequently, a uniform global learning rate triggers exploding gradients in certain layers while inducing vanishing gradients in others, ultimately leading to the complete loss of the model's representational capacity.
\subsubsection{Long-term Training Scalability}
\begin{figure*}[t!] 
    \centering
    \adjincludegraphics[width=1.0\linewidth]{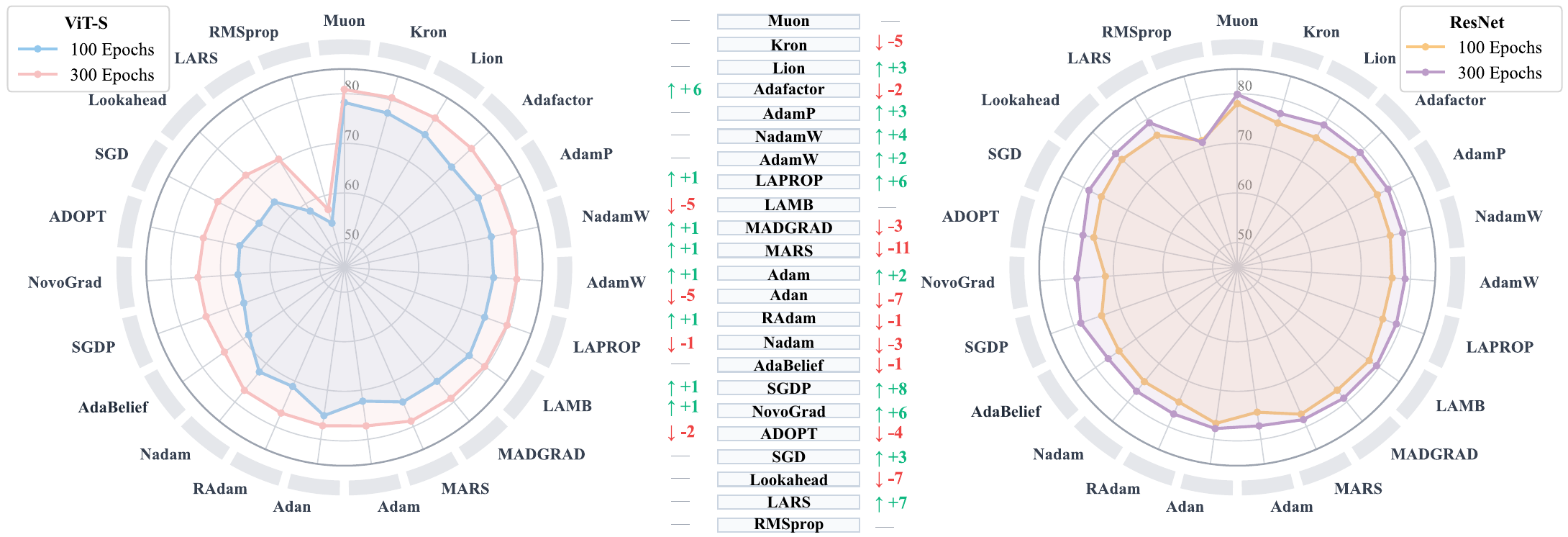}
   
    \caption{\textbf{Performance Comparison of Various Optimizers.} We illustrate the ranking shifts of various optimizers on ViT-S~\cite{296} and ResNet-50~\cite{297} after 100 and 300 training epochs.}
    
    \label{fig:lidar}
    \vspace{-1.0em} 
\end{figure*}

To thoroughly investigate the long-term scalability of these optimizers, we extend the 100-epoch training protocol to 300 epochs on both ViT-S~\cite{296} and ResNet-50~\cite{297} models. We detail the Top-1 accuracy at both training milestones and quantify the resulting performance improvements ($\Delta$) in~\cref{table:Long-term Training Scalability}, with the overall convergence trajectories visualized (\cref{fig:figure_vrcontrast}), and comprehensive optimizer performance profiles illustrated through radar charts (\cref{fig:lidar}).\\
\textbf{Architecture-specific scaling.}
The SGD series exhibits strong scalability, with SGD~\cite{307} gaining $9.41\%$ on ViT-S~\cite{296}. Mechanistically, these optimizers rely on first-order momentum without aggressive element-wise variance accumulation. This mathematical design prevents rapid step-size decay; while it leads to slower initial convergence, it enables the model to extensively explore the loss landscape extensively.\\
\textbf{Consistent scalability.} When training is extended to 300 epochs, advanced optimizers including Muon~\cite{173}, Lion~\cite{167}, Kron~\cite{306} and AdamW~\cite{148} exhibit consistent yet modest accuracy gains, ranging from $+1.896\%$ to $+4.690\%$, across various architectures. For instance, Muon~\cite{173} achieves a high baseline of 78.252 on ViT-S~\cite{296} at 100 epochs, with its 300-epoch gain capped at $+2.622\%$. This phenomenon occurs because these methods leverage sophisticated preconditioning, orthogonalization, or decoupled weight decay to rapidly traverse the optimization trajectory. By efficiently exploiting local curvature and structural priors, they converge toward high quality's optima within the initial 100 epochs. Consequently, further training iterations yield diminishing returns, as these optimizers have already extracted the vast majority of the network's representational capacity during the early stages.\\
\textbf{Performance regression.}
Conversely, RMSprop \cite{325} exhibits a detrimental response to prolonged training on ResNet-50~\cite{297}, experiencing a performance regression between 100 and 300 epochs. This degradation underscores the structural vulnerability of relying on an uncorrected EMA of squared gradients over protracted training cycles. Lacking the stabilizing effect of first-order momentum or decoupled regularization mechanisms, the continuous accumulation of late-stage gradient noise disproportionately distorts parameter-wise step sizes. Over the course of 300 epochs, this imbalanced scaling disrupts the fragile equilibrium within local minima, forcing the optimizer to escape optimal basins and leading to a decline in representational capacity in the late stage.\\
\subsubsection{Correlation of Optimizers}
\begin{figure*}[t!] 
    \centering
    \includegraphics[width=1.0\linewidth]{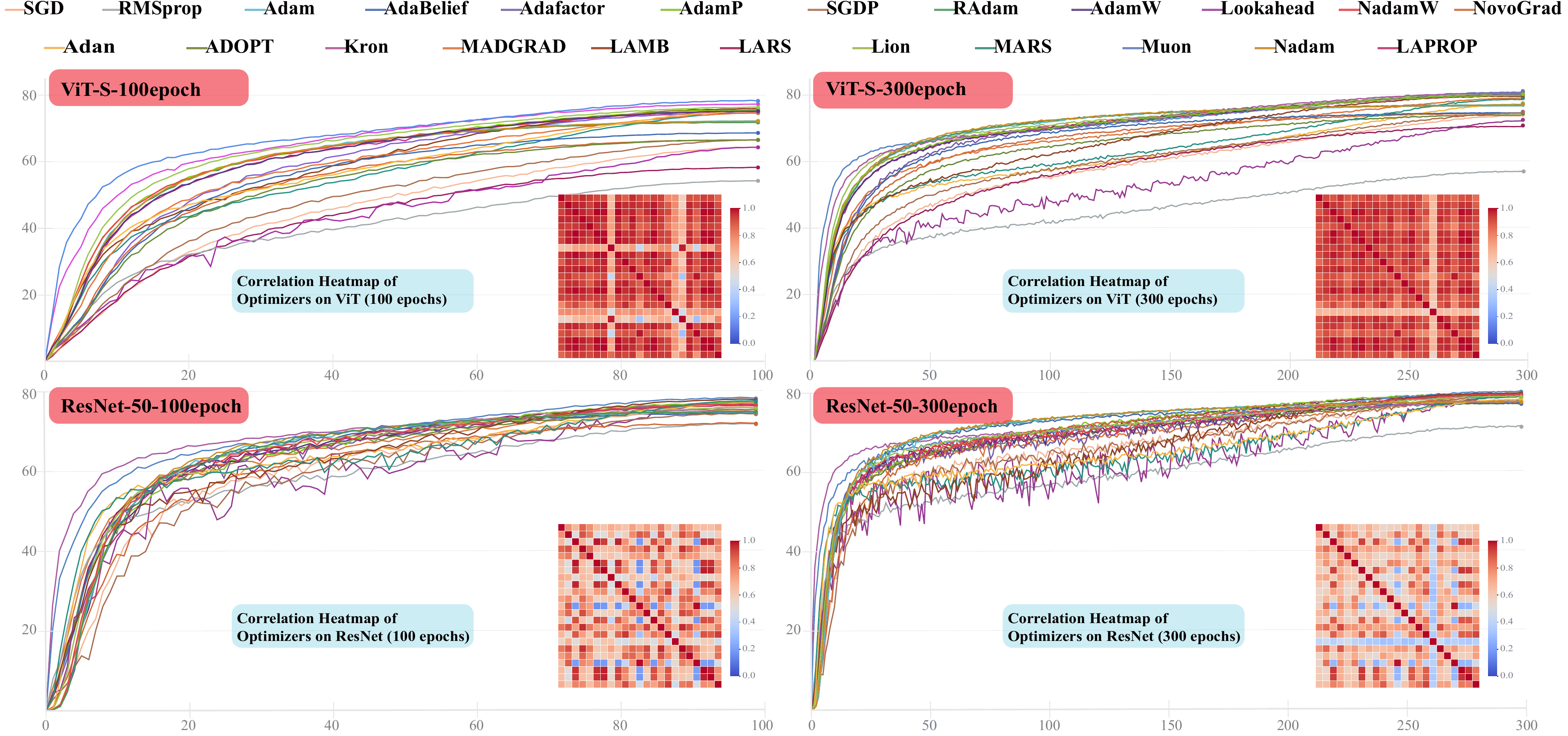} 
   
    \caption{\textbf{Top-1 accuracy and trajectory correlation of popular optimizers.} The main line plots display the Top-1 accuracy performance of 23 optimizers on ViT-S~\cite{296} and ResNet-50~\cite{297} models over 100 and 300 epochs. The inset heatmaps present the Pearson correlation matrices calculated via first-order differencing of the validation metrics, revealing the algorithmic categorization and trajectory divergence across different architectures and training horizons.}
    
    \label{fig:figure_vrcontrast}
\end{figure*}

We present a correlation analysis (\cref{fig:figure_vrcontrast}) of optimizer performance on ViT-S~\cite{296} and ResNet-50~\cite{297}. To accurately capture the dynamic behavior of various optimizers, we analyze the first-order differencing of the validation loss rather than the raw loss values. Raw loss trajectories often exhibit high collinearity due to the overarching trend of network convergence. In contrast, first-order differencing isolates the step-wise acceleration and deceleration of the loss function. This approach effectively highlights the temporal inflection points where an optimizer escapes suboptimal regions or enters rapid descent phases. By stripping away the cumulative scale of the loss, the differencing method reveals the intrinsic rhythm and pacing of the underlying update mechanisms.\\
\textbf{Architectural influence on convergence homogeneity}
The empirical correlation matrices reveal that model architecture exerts a profound influence on the optimization trajectory, occasionally superseding the optimizer's algorithmic design. In the ViT~\cite{296} experiments, the correlation matrix displays overwhelming homogeneity across all evaluated optimizers. This phenomenon indicates that the highly constrained training regime of ViT~\cite{296}, which heavily relies on prolonged linear learning rate warmup and stringent regularization, dictates the loss landscape traversal. Consequently, the scheduling strategy effectively overrides the idiosyncratic step-size adaptations of individual optimizers. Conversely, the ResNet-50~\cite{297} experiments exhibit significant algorithmic bifurcation. The inherent inductive bias and smoother loss landscape of CNNs grant optimizers the geometric freedom to explore distinct descent paths, resulting in highly variable inter-optimizer correlations.\\
\textbf{Algorithmic categorization and trajectory divergence}
It should be noted that in all presented correlation heatmaps, the optimizers are arranged from left to right along the horizontal axis and from top to bottom along the vertical axis, strictly following the sequence listed in~\cref{fig:figure_vrcontrast}. Within the less constrained ResNet-50~\cite{297} environment, these heatmaps reveal distinct optimizer taxonomy clusters based on their dynamic behavior.\\
\textbf{Adaptive family.} Methods such as Adam~\cite{287}, RAdam~\cite{160}, AdamP~\cite{229}, AdamW~\cite{148}, Nadam~\cite{155}, and NadamW~\cite{326} form a tightly correlated cluster. Their shared reliance on the exponential moving average of squared gradients yields synchronized adaptation to landscape steepness, causing their rapid descent phases to align temporally.\\
\textbf{Preconditioning and orthogonalization.}
Notably, Kron~\cite{306} and Muon~\cite{173} exhibit an exceptionally high intra-group correlation while remaining distinctly divergent from the adaptive scalar family. This empirical divergence perfectly mirrors their theoretical foundations. By leveraging structural gradient statistics for preconditioning or employing matrix orthogonalization techniques, these methods navigate the parameter space via structural matrix updates rather than simple diagonal scaling, resulting in fundamentally different loss traversal trajectories.\\
\textbf{Mechanistic outliers.} Optimizers with structural deviations, such as Lookahead~\cite{225}, MADGRAD~\cite{171}, and SGDP~\cite{229}, present consistently low correlations with standard methods. For instance, the dual-weight interpolation mechanism of Lookahead~\cite{225} introduces a temporal lag in the first-order difference sequence, shifting its inflection points relative to standard step-based methods.\\
\textbf{Temporal saturation and cross-epoch homogenization.}
Comparing short-term and long-term training regimes reveals a clear temporal smoothing effect. During the initial training phase, the exploratory divergence among optimizers is maximized, highlighting their unique mathematical characteristics. However, as training extends from 100 to 300 epochs, the correlation matrices demonstrate an overall increase in homogeneity. This shift occurs because the majority of optimizers eventually converge into a localized basin of attraction. In this exploitation phase, learning rates decay to minimal values, and the first-order differencing sequences are dominated by near-zero fluctuations. This asymptotic saturation mathematically forces the correlation coefficients to converge, illustrating that diverse algorithmic paths ultimately lead to similar geometrical destinations over extended temporal horizons.\\
   
    
\begin{table}[t!]
    \centering
    \caption{Results on ViT-S~\cite{296} and ResNet-50~\cite{297} models across various optimizers, where we compare the 100 epoch accuracy and 300 epoch accuracy and show that ViT-S~\cite{296} achieves larger performance improvements from extended training than ResNet-50~\cite{297}.}
    \label{table:Long-term Training Scalability}
    \fontsize{7.5pt}{9pt}\selectfont
    \setlength\tabcolsep{3pt}
    \resizebox{\linewidth}{!}{
        \begin{tabular}{c l c c c}
        \toprule
        \textbf{Model} & \textbf{Optimizer} & \makecell{\textbf{100 Epoch} \\ \textbf{Acc (\%)}} & \makecell{\textbf{300 Epoch} \\ \textbf{Acc (\%)}} & \makecell{\textbf{Improvement} \\ \textbf{($\Delta$)}} \\
        \midrule
        \multirow{23}{*}{\rotatebox[origin=c]{90}{\textbf{ViT-S}}} 
          & SGD~\cite{307} & 64.360 & 73.774 & \textbf{+ 9.414} \\
          & RMSprop~\cite{325} & 54.254 & 57.090 & \textbf{+ 2.836} \\
          & Adam~\cite{287} & 72.240 & 77.274 & \textbf{+ 5.034} \\
          & AdaBelief~\cite{149} & 68.642 & 74.640 & \textbf{+ 5.998} \\
          & Adafactor~\cite{154} & 74.630 & 80.108 & \textbf{+ 5.478} \\
          & RAdam~\cite{160} & 71.190 & 77.032 & \textbf{+ 5.842} \\
          & AdamP~\cite{229} & 75.444 & 79.916 & \textbf{+ 4.472} \\
          & SGDP~\cite{229} & 66.536 & 74.620 & \textbf{+ 8.084} \\
          & AdamW~\cite{148} & 75.196 & 79.886 & \textbf{+ 4.690} \\
          & Adan~\cite{168} & 75.190 & 77.238 & \textbf{+ 2.048} \\
          & ADOPT~\cite{196} & 66.552 & 74.058 & \textbf{+ 7.506} \\
          & Kron~\cite{306} & 77.304 & 80.510 & \textbf{+ 3.206} \\
          & LAMB~\cite{227} & 75.858 & 79.672 & \textbf{+ 3.814} \\
          & LAPROP~\cite{156} & 75.050 & 79.850 & \textbf{+ 4.800} \\
          & LARS~\cite{223} & 58.276 & 70.530 & \textbf{+ 12.254} \\
          & Lion~\cite{167} & 76.334 & 80.258 & \textbf{+ 3.924} \\
          & Lookahead~\cite{225} & 64.320 & 72.220 & \textbf{+ 7.900} \\
          & MADGRAD~\cite{171} & 74.634 & 79.080 & \textbf{+ 4.446} \\
          & MARS~\cite{181} & 74.608 & 78.800 & \textbf{+ 4.192} \\
          & Muon~\cite{173} & 78.252 & 80.874 & \textbf{+ 2.622} \\
          & Nadam~\cite{155} & 72.180 & 76.978 & \textbf{+ 4.798} \\
          & NadamW~\cite{326} & 75.278 & 79.912 & \textbf{+ 4.634} \\
          & NovoGrad~\cite{157} & 66.514 & 74.616 & \textbf{+ 8.102} \\
        \midrule
        \multirow{23}{*}{\rotatebox[origin=c]{90}{\textbf{ResNet-50}}} 
          & SGD~\cite{307} & 75.926 & 78.726 & \textbf{+ 2.800} \\
          & RMSprop~\cite{325} & 71.536 & 71.152 & \textbf{- 0.384} \\  
          & Adam~\cite{287} & 74.480 & 77.274 & \textbf{+ 2.794} \\
          & AdaBelief~\cite{149} & 74.232 & 76.914 & \textbf{+ 2.682} \\
          & Adafactor~\cite{154} & 76.808 & 78.940 & \textbf{+ 2.132} \\
          & RAdam~\cite{160} & 74.612 & 77.252 & \textbf{+ 2.640} \\
          & AdamP~\cite{229} & 76.818 & 79.252 & \textbf{+ 2.434} \\
          & SGDP~\cite{229} & 74.066 & 78.534 & \textbf{+ 4.468} \\
          & AdamW~\cite{148} & 76.314 & 78.946 & \textbf{+ 2.632} \\
          & Adan~\cite{168} & 76.770 & 77.818 & \textbf{+ 1.048} \\
          & ADOPT~\cite{196} & 74.584 & 76.818 & \textbf{+ 2.234} \\
          & Kron~\cite{306} & 75.226 & 77.216 & \textbf{+ 1.990} \\
          & LAMB~\cite{227} & 77.590 & 79.404 & \textbf{+ 1.814} \\
          & LAPROP~\cite{156} & 76.146 & 79.066 & \textbf{+ 2.920} \\
          & LARS~\cite{223} & 76.196 & 79.124 & \textbf{+ 2.928} \\
          & Lion~\cite{167} & 75.582 & 78.616 & \textbf{+ 3.034} \\
          & Lookahead~\cite{225} & 76.872 & 78.610 & \textbf{+ 1.738} \\
          & MADGRAD~\cite{171} & 76.940 & 79.028 & \textbf{+ 2.088} \\
          & MARS~\cite{181} & 77.294 & 78.478 & \textbf{+ 1.184} \\
          & Muon~\cite{173} & 77.996 & 79.892 & \textbf{+ 1.896} \\
          & Nadam~\cite{155} & 74.732 & 77.226 & \textbf{+ 2.494} \\
          & NadamW~\cite{326} & 76.514 & 79.076 & \textbf{+ 2.562} \\
          & NovoGrad~\cite{157} & 71.694 & 77.472 & \textbf{+ 5.778} \\
        \bottomrule
        \end{tabular}
    }
\end{table}%
\subsubsection{Comprehensive Optimizer Evaluation.}
A rigorous empirical comparison of more than twenty optimizers is detailed in~\cref{table:optim_comparison}, which employs a multifaceted star evaluation paradigm across five essential criteria to clearly demonstrate that an increased star count reflects enhanced algorithmic proficiency. Specifically, the star ratings indicate relative performance percentiles among the evaluated methods. The quantitative basis for these ratings is grounded in specific empirical evaluations: accuracy aggregates the comprehensive performance across three different architectures (see~\cref{table:optim_generalization}); generalization is demonstrated through cross-architecture generalization experiments (see~\cref{generalization}); and hyperparameter robustness is evaluated via learning rate scaling experiments (see~\cref{table:lr_sensitivity}). Additionally, convergence speed and memory efficiency are quantified by the iteration steps to reach a target loss and the peak GPU memory footprint, respectively. Overall, Muon~\cite{173} achieves the best comprehensive performance.
\newcommand{\sI}{\textcolor{blue}{\ding{72}}}               
\newcommand{\sII}{\textcolor{green!60!gray}{\ding{72}}}   
\newcommand{\sIII}{\textcolor{orange}{\ding{72}}}  
\newcommand{\sIV}{\textcolor{red!75!black}{\ding{72}}}         
\newcommand{\sV}{\textcolor{yellow!80!red}{\ding{72}}}

\begin{table}[t]
    \centering
    \caption{Comparison of various optimizers across five key dimensions. We evaluate convergence speed, accuracy, memory efficiency, generalization, and hyperparameter robustness, and show that Muon achieves the best overall performance.}
    \label{table:optim_comparison}
    \renewcommand{\arraystretch}{1.2} 
    \setlength\tabcolsep{2.0pt}
    \resizebox{1.\linewidth}{!}{
        \begin{tabular}{l p{1.8cm}<{\centering} p{3.5cm}<{\centering} p{2.5cm}<{\centering} p{3.5cm}<{\centering} p{3.0cm}<{\centering} p{5.0cm}<{\centering}}
        \toprule
        \multirow{2}{*}{\textbf{Optimizer}} & \textbf{Convergence} & \multirow{2}{*}{\textbf{Accuracy}} & \textbf{Memory} & \multirow{2}{*}{\textbf{Generalization}} & \textbf{Hyperparameter} \\
         & \textbf{Speed} & & \textbf{Efficiency} & & \textbf{Robustness} \\
        \hline
        SGD~\cite{307} & \sI & \sII\sII & \sV\sV\sV\sV\sV & \sII\sII & \sI \\ 
        RMSprop~\cite{325} & \sI & \sI & \sIV\sIV\sIV\sIV & \sII\sII & \sII\sII \\
        Adam~\cite{287} & \sIII\sIII\sIII & \sIII\sIII\sIII & \sIII\sIII\sIII & \sII\sII & \sIII\sIII\sIII \\
        AdaBelief~\cite{149} & \sIV\sIV\sIV\sIV & \sII\sII & \sIII\sIII\sIII & \sII\sII & \sII\sII \\
        Adafactor~\cite{154} & \sII\sII & \sIII\sIII\sIII & \sIV\sIV\sIV\sIV & \sIII\sIII\sIII & \sIII\sIII\sIII \\
        RAdam~\cite{160} & \sIV\sIV\sIV\sIV & \sIII\sIII\sIII & \sIII\sIII\sIII & \sII\sII & \sIII\sIII\sIII \\
        AdamP~\cite{229} & \sIII\sIII\sIII & \sIV\sIV\sIV\sIV & \sIII\sIII\sIII & \sIII\sIII\sIII & \sIII\sIII\sIII \\
        SGDP~\cite{229} & \sIII\sIII\sIII & \sII\sII & \sIV\sIV\sIV\sIV & \sII\sII & \sI \\
        AdamW~\cite{148} & \sIV\sIV\sIV\sIV & \sIV\sIV\sIV\sIV & \sIII\sIII\sIII & \sIV\sIV\sIV\sIV & \sIII\sIII\sIII \\
        Adan~\cite{168} & \sIV\sIV\sIV\sIV & \sIV\sIV\sIV\sIV & \sII\sII & \sV\sV\sV\sV\sV & \sV\sV\sV\sV\sV \\
        ADOPT~\cite{196} & \sIII\sIII\sIII & \sII\sII & \sIII\sIII\sIII & \sII\sII & \sIII\sIII\sIII \\
        Kron~\cite{306} & \sV\sV\sV\sV\sV & \sV\sV\sV\sV\sV & \sII\sII & \sIV\sIV\sIV\sIV & \sV\sV\sV\sV\sV \\
        LAMB~\cite{227} & \sIV\sIV\sIV\sIV & \sIV\sIV\sIV\sIV & \sIII\sIII\sIII & \sIV\sIV\sIV\sIV & \sII\sII \\
        LAPROP~\cite{156} & \sIII\sIII\sIII & \sIV\sIV\sIV\sIV & \sIII\sIII\sIII & \sIII\sIII\sIII & \sIII\sIII\sIII \\
        LARS~\cite{223} & \sI & \sI & \sIV\sIV\sIV\sIV & \sII\sII & \sII\sII \\
        Lion~\cite{167} & \sV\sV\sV\sV\sV & \sV\sV\sV\sV\sV & \sIV\sIV\sIV\sIV & \sIV\sIV\sIV\sIV & \sIV\sIV\sIV\sIV \\
        Lookahead~\cite{225} & \sI & \sII\sII & \sII\sII & \sII\sII & \sII\sII \\
        MADGRAD~\cite{171} & \sIII\sIII\sIII & \sIII\sIII\sIII & \sIII\sIII\sIII & \sII\sII & \sIV\sIV\sIV\sIV \\
        MARS~\cite{181} & \sIV\sIV\sIV\sIV & \sIII\sIII\sIII & \sIV\sIV\sIV\sIV & \sV\sV\sV\sV\sV & \sIV\sIV\sIV\sIV \\
        Muon~\cite{173} & \sV\sV\sV\sV\sV & \sV\sV\sV\sV\sV & \sIV\sIV\sIV\sIV & \sV\sV\sV\sV\sV & \sV\sV\sV\sV\sV \\
        Nadam~\cite{155} & \sIII\sIII\sIII & \sIII\sIII\sIII & \sIII\sIII\sIII & \sII\sII & \sIII\sIII\sIII \\
        NadamW~\cite{326} & \sIV\sIV\sIV\sIV & \sIV\sIV\sIV\sIV & \sIII\sIII\sIII & \sIV\sIV\sIV\sIV & \sIII\sIII\sIII \\
        NovoGrad~\cite{157} & \sII\sII & \sII\sII & \sIV\sIV\sIV\sIV & \sII\sII & \sII\sII \\

        \bottomrule
        \end{tabular}
    }
    \vspace{-1.0em}
\end{table}%
\begin{table*}[tp]
\centering
\caption{Comprehensive estimation of computational cost (GPU Hours). The table details the estimated single NVIDIA A100 GPU hours required for the entire empirical benchmark. The workload accounts for 23 optimizers, including rigorous grid searches across 5 learning rate scales and extended 300-epoch training horizons. The massive total computational footprint ($\sim$1073 GPU Hours) underscores the rigorousness of our evaluation and highlights the fundamental computational barrier in modern optimizer benchmarking.}
\label{tab:gpu_hours}
\renewcommand{\arraystretch}{1.2} 
\setlength\tabcolsep{3.0pt}
\resizebox{1.0\linewidth}{!}{
\begin{tabular}{lcccl}
\toprule[1.5pt]
\textbf{Benchmark Task} & \textbf{Hardware Type} & \makecell[c]{\textbf{Hyperparameter Search} \\ (5 LR Scales $\times$ 23 Optimizers)} & \makecell[c]{\textbf{Final Evaluation} \\ (Best LR $\times$ 23 Optimizers)} & \textbf{Total GPU Hours} \\
\midrule
\makecell[l]{\textbf{ResNet-50~\cite{297}} \\ (ImageNet-1K~\cite{295})} & NVIDIA A100 & - & $\sim$288 Hours\scriptsize{(100/300 epochs)} & $\sim$288 Hours \\
\hline
\makecell[l]{\textbf{ViT-S~\cite{296}} \\ (ImageNet-1K~\cite{295})}     & NVIDIA A100 & $\sim$544 Hours \scriptsize{(100 epochs)} & $\sim$213 Hours \scriptsize{(300 epochs)} & $\sim$757 Hours \\
\hline
\makecell[l]{\textbf{LlaMA-60M~\cite{330}} \\ (WikiText-103~\cite{329})}& NVIDIA A100 & $\sim$28 Hours \scriptsize{(Chinchilla scaling)} & - & $\sim$28 Hours \\
\midrule
\rowcolor{gray!10} \textbf{Total Aggregated Cost} & \multicolumn{3}{r}{\textbf{Equivalent to $\sim$45 Days on a single NVIDIA A100 GPU}} & \textbf{$\sim$1073 Hours} \\
\bottomrule
\end{tabular}
}
\end{table*}%
\subsubsection{Computational cost disclosure.} To guarantee the internal validity and reproducibility of our findings, we enforce a strict grid-search protocol for hyperparameters across all 23 evaluated optimizers. However, this rigorous methodology induces an exponential explosion in computational overhead. As detailed in~\cref{tab:gpu_hours}, the complete execution of our empirical framework, encompassing multi-scale learning rate sweeps and extended 300-epoch horizons, amasses an estimated footprint of over 1073 single NVIDIA A100 hours.

\section{Future Prospect} \label{section:fut}
We summarize the current challenges in the field of optimization. Based on challenges and the preceding experimental analysis, we explore the potential development directions for optimization algorithms.

\textbf{Challenges in current optimization frameworks.} Optimizing large models is fundamentally a highly constrained, multi objective trade-off. When current methods pursue specific metrics, such as convergence speed, memory efficiency, or distributed scaling, they often sacrifice generalization, introduce computational latency, or exacerbate system instability. Therefore, a core challenge for existing optimization algorithms is to break these interconnected bottlenecks under a unified design framework, effectively balancing stability, computational cost, and noise robustness.

\textbf{Generalization and stability margins.}
Current adaptive methods frequently trap models in sharp minima, resulting in worse generalization than standard SGD~\cite{307}. While preconditioned approaches use structural approximations to improve convergence, they exhibit a narrow stability margin where large learning rates cause training divergence~\cite{306}. Furthermore, stateless memory-efficient designs lose the historical accumulation of second moments, removing their ability to average extreme gradient shocks and risking training divergence in non-convex loss landscapes~\cite{154}.

\textbf{Memory and computational overheads.}
As models scale to billions of parameters, dense optimizer states create a primary memory bottleneck~\cite{331}. Structural approximations attempt to balance precision and memory overhead, but matrix inversions and structural overhead significantly decrease global efficiency~\cite{255}. Although certain methods theoretically require fewer iterations to converge, their per-step computational latency often negates these gains, limiting their application in large-scale settings.

\textbf{Noise amplification and estimation variance.}
In the anisotropic loss landscapes typical of deep attention networks, the stochastic noise from mini-batch training is heavily amplified during curvature estimation, reducing update stability~\cite{264}. Similarly, random perturbations~\cite{284} used to estimate directional gradients suffer from approximation variance that increases rapidly with parameter dimensionality. Privacy-preserving optimizers face a parallel utility tradeoff, as the required noise injection for differential privacy degrades gradient fidelity and destabilizes convergence~\cite{332}.

\textbf{Scalability and consensus tensions.}
Optimization trajectories remain highly sensitive to hyperparameters, requiring extensive manual tuning for successful scaling. In distributed settings, combining gradient compression with local updates increases client drift~\cite{81}, creating a persistent tension between communication efficiency and global consensus. Additionally, optimizing within low-rank subspaces to manage high-dimensional search spaces can prematurely constrain the trajectory and prevent access to global optima~\cite{333}.

\textbf{First-order optimization.}
To address these limitations, future development of first-order optimization could focus on three key directions: 
\textbf{\textit{1) Automated symbolic discovery:}} The methodological focus should shift from fragile heuristic tuning to the automated generation of architecture-specific optimizers, enabling models to inherently navigate complex loss landscapes without manual intervention. 
\textbf{\textit{2) Preconditioning and orthogonalization:}} Given that methods like Kron~\cite{306} and Muon~\cite{173} exhibit fundamentally different loss traversal trajectories compared to the adaptive scalar family, future research should focus on advancing structural matrix updates. By leveraging structural gradient statistics for preconditioning or employing matrix orthogonalization techniques, these approaches overcome the representational bottlenecks of simple diagonal scaling, opening up novel and highly efficient pathways for navigating the parameter space.
\textbf{\textit{3) Dynamic precision scaling:}} To alleviate the severe memory bottlenecks imposed by dense optimizer states, future research should integrate adaptive low-precision arithmetic that significantly reduces memory footprints without compromising convergence stability.

\textbf{Second-order optimization.}
To address these prohibitive bottlenecks, future development of second-order frameworks should emphasize structural innovation across three key areas: 
\textbf{\textit{1) Hardware-algorithm co-design:}} To overcome emerging memory bottlenecks, research should pivot toward optimizing global computational efficiency by integrating structure-aware adaptation and low-precision arithmetic to balance computational intensity with convergence stability. 
\textbf{\textit{2) Robust curvature estimation:}} Future designs should develop mathematically sound mechanisms to isolate extreme gradient outliers, expanding the dangerously narrow safety margin without inflating memory footprints. 
\textbf{\textit{3) Sparse preconditioning:}} To mitigate prohibitive per-step computational latency, future architectures should explore highly efficient sparse matrix inversion techniques that preserve the intrinsic advantages of second-order information while achieving global efficiency comparable to first-order baselines.

\textbf{Zeroth-order optimization.}
To bridge this substantial performance gap, the future development of gradient-free frameworks must focus on noise and space management. 
\textbf{\textit{1)}} Exact noise cancellation: Future frameworks must mathematically cancel the inherent stochastic noise introduced by diverse network architectures, potentially drawing inspiration from exact gradient correction mechanisms that subtract historical perturbation bias to stabilize the highly variable update steps. 
\textbf{\textit{2)}} Dimensionality robustness: Research should discover advanced subspace projection methods that safely constrain the search space without prematurely restricting access to high quality global solutions. \textbf{\textit{3)}} Adaptive perturbation scheduling: Rather than relying on static random perturbations, future strategies should dynamically adjust the variance and direction of blind estimates based on local landscape geometry, thereby preventing continuous disruption of the fragile equilibrium within local minima.

\textbf{Scenario-oriented frameworks.} To ensure scalable and trustworthy deployment in critical infrastructure, future development of these specialized frameworks must shift from empirical compromises to rigorous guarantees across three key dimensions: 
\textbf{\textit{1)}} Rigorous theoretical bounds: Distributed and privacy preserving frameworks must establish strict mathematical lower bounds for communication efficiency and privacy utility. 
\textbf{\textit{2)}} State fidelity preservation: Future memory efficient architectures must be designed to retain just enough historical context to smoothly average out extreme gradient shocks without hitting the rigid fidelity ceiling or exceeding memory constraints. 
\textbf{\textit{3)}} Heterogeneous consensus mechanisms: To resolve the persistent tension between communication efficiency and client drift, distributed setups must introduce topology aware gradient compression techniques that naturally accommodate the massive disparities in gradient magnitudes across diverse network architectures.

\textbf{Discussion.}
Future optimization frameworks will likely move beyond isolated algorithmic improvements by integrating FO, SO, and ZO algorithms. To address current memory bottlenecks and structural compromises, research must shift from minimizing iteration complexity to improving global wall-clock efficiency. This transition requires hardware-algorithm co-design. Specifically, structural preconditioning and matrix orthogonalization are critical. Integrating these techniques with dynamic low-precision arithmetic balances computational cost and convergence stability, allowing optimizers to overcome the representational limits of simple diagonal scaling. Methodologically, the focus should shift from heuristic tuning and static noise injection to automated symbolic discovery and exact noise cancellation. Combined with the geometric corrections of preconditioning and orthogonalization, these advancements enable architecture-specific optimizers that handle highly anisotropic loss landscapes and resist extreme gradient outliers. In distributed and privacy-preserving settings, frameworks must also advance beyond empirical trade-offs. Establishing theoretical lower bounds for communication and utility, along with topology-aware consensus mechanisms, will ensure the scalable and reliable deployment of these specialized architectures.
\section{Conclusion} \label{section:con}
This survey presents a comprehensive, systematic review of the latest advancements in deep learning optimization methods. We first establish a rigorous theoretical foundation by reviewing core background concepts in optimization theory, providing a unified framework for understanding the evolution of optimization algorithms. We then systematically categorize and analyze technical approaches across four main domains: first-order, second-order, and zeroth-order optimization algorithms, as well as specialized scenario-oriented optimization frameworks. Our analysis examines both the structural design principles and functional characteristics of each class of optimization methods. 
To provide practical, actionable insights, we conduct a rigorous, controlled empirical evaluation that benchmarks representative optimizers across diverse task domains and model architectures. Our experimental analysis reveals critical trade-offs between hyperparameter sensitivity, cross-architecture generalization capability, and long-term training scalability, offering valuable guidance for algorithm selection and hyperparameter tuning in real-world applications. 
Finally, we identify key open challenges in the field and outline promising future research directions to guide the development of next-generation high-efficiency, robust, and trustworthy optimization technologies. By synthesizing theoretical insights with extensive empirical evidence, this survey aims to serve as a comprehensive, authoritative resource for researchers and practitioners seeking to advance the state-of-the-art in deep learning optimization.


\bibliography{april_aigc}

\end{document}